\documentclass[review,fleqn,sort&compress]{elsarticle}
\usepackage{amssymb,amsbsy,amsthm,amsmath,amsfonts,amscd}
\usepackage{geometry}
\usepackage{bm} 
\usepackage{graphicx}
\usepackage{float}
\usepackage{multirow}
\usepackage{subfigure}
\usepackage{lineno}
\usepackage{hyperref}
\hypersetup{colorlinks, citecolor=black, filecolor=black, linkcolor=black, urlcolor=black}
\usepackage{booktabs}
\usepackage[ruled,linesnumbered]{algorithm2e}
\usepackage{appendix}
\usepackage{algpseudocode}
\usepackage{mathtools,nccmath}
\usepackage{adjustbox}
\usepackage{cleveref}
\usepackage{dutchcal}
\usepackage[utf8]{inputenc}
\usepackage[scr=rsfso,cal=euler,frak=euler,bb=ams]{mathalfa}
\usepackage{longtable}
\usepackage{comment}
\usepackage{color}
\usepackage[normalem]{ulem}
\newcommand{\stkout}[1]{\ifmmode\text{\sout{\ensuremath{#1}}}\else\sout{#1}\fi}

\graphicspath{{fig/}}
\theoremstyle{definition}

\usepackage{color}
\usepackage{soul}

\begin{document}

\begin{frontmatter}
\title{Error convergence and engineering-guided hyperparameter search of PINNs: towards optimized I-FENN performance}
\author{Panos Pantidis\corref{cor1}}
\author{Habiba Eldababy}
\author{Christopher Miguel Tagle}
\author{Mostafa E. Mobasher\corref{cor2}}
\cortext[cor1]{Corresponding author. \emph{E-mail address:} \texttt{pp2624@nyu.edu} (Panos Pantidis)}
\cortext[cor2]{Corresponding author. \emph{E-mail address:} \texttt{mostafa.mobasher@nyu.edu} (Mostafa Mobasher)}
\address{Civil and Urban Engineering Department, New York University Abu Dhabi, Abu Dhabi, P.O. Box 129188, UAE}

\begin{highlights}

\item The PINN numerical convergence is assessed while accounting for statistical learning error analysis 

\item The training and the global error converge in complexity and sample limits through Adam and L-BFGS

\item A novel set of holistic performance metrics for the PINN accuracy, cost, training efficiency and impact on I-FENN is established

\item A higher complexity PINN leads to more efficient training and I-FENN performance than a larger training dataset

\item The optimum PINN architecture for the non-local gradient PDE has a close-to-square shape

\end{highlights}

\begin{abstract}

In our recently proposed Integrated Finite Element Neural Network (I-FENN) framework \cite{pantidis2022integrated} we showcased how PINNs can be deployed on a finite element-level basis to swiftly approximate a state variable of interest, and we applied it in the context of non-local gradient-enhanced damage mechanics. In this paper, we enhance the rigour and performance of I-FENN by focusing on two crucial aspects of its PINN component: a) the error convergence analysis and b) the hyperparameter-performance relationship. Guided by the available theoretical formulations in the field, we introduce a systematic numerical approach based on a novel set of holistic performance metrics to answer both objectives. In the first objective, we explore in detail the convergence of the PINN training error and the global error against the network size and the training sample size. We demonstrate a consistent converging behavior of the two error types for any investigated combination of network complexity, dataset size and choice of hyperparameters, which empirically proves the conformance of the PINN setup and implementation to the available convergence theories. In the second objective, we establish an a-priori knowledge of the hyperparameters which favor higher predictive accuracy, lower computational effort, and the least chances of arriving at trivial solutions. The analysis leads to several outcomes that contribute to the better performance of I-FENN, and fills a long-standing gap in the PINN literature with regards to the numerical convergence of the network errors while accounting for commonly used optimizers (Adam and L-BFGS). The proposed analysis method can be directly extended to other ML applications in science and engineering. The code and data utilized in the analysis are posted publicly to aid the reproduction and extension of this research.

\end{abstract}

\begin{keyword}
\texttt Physics Informed Neural Networks \sep I-FENN \sep convergence \sep  hyperparameter optimization \sep numerical methods \sep non-local damage 
\end{keyword}

\end{frontmatter}

\section{Introduction}
\label{Sec:Introduction}

\subsection{Literature review}
 
The recent re-emergence of Physics-Informed Neural Networks (PINNs) \cite{lagaris1998artificial, raissi2019physics} has signified a pivotal shift both in the extent and in the way that Machine Learning algorithms are utilized in the fields of science and engineering \cite{mao2020physics,mathews2021uncovering,kovacs2022conditional,henkes2022pinnsmicromech,haghighat2021physics,rao2020physics}, thanks to their ability to infuse physical knowledge inside the optimization scheme. Despite the remarkable success of PINNs in several benchmark cases \cite{cuomo2022scientific}, studies have reported that PINNs may fail to learn more complex physical phenomena, and understanding their limitations is a subject of a currently growing body of literature \cite{krishnapriyan2021characterizing, wang2021understanding, rohrhofer2022understanding, rohrhofer2021pareto}. Moving beyond the fundamental training pathologies of PINNs, there are several more aspects which need to be thoroughly investigated before PINNs can mature into more rigorous numerical tools, and below we mention two key ones: the first is their theoretical consistency from a convergence viewpoint, and the second is the optimal selection of hyperparameters. In view of our recently proposed Integrated Finite Element Neural Network (I-FENN) framework, and informed by the available theoretical formulations in the field, the present paper aims to address both.

Before we present the relevant literature review, it is crucial to expand on the reasoning why it is imperative to examine their performance from this lens. In the development of robust numerical methods, such as the Finite Element Method (FEM), Finite Difference (FD), and Boundary Element Method (BEM), researchers spend extensive efforts in establishing the healthy norms and best practices of using these methods. For example, in linear and non-linear FEM, many studies have established convergence criteria \cite{ciarlet1973maximum,babuska1982rates}, approximation functions and integration laws \cite{malkus1978mixed,fried1974numerical}, as well as other efforts that lead to enhanced and robust numerical performance \cite{jansen2000generalized, moresi2003lagrangian}. On the other hand, efforts on PINNs and their variations \cite{lu2021physics,kharazmi2019variational,yang2021b} are mostly focused on modeling new phenomena or improving the network ability to discover or represent the underlying mathematical features \cite{lu2021learning,wang2021learning}. At the same time, some fundamental issues remain unresolved; for example there is no guarantee that the PINN optimization algorithm will reach a global minimum \cite{shin2020convergence}, the relationship between the network topology and performance is still unclear \cite{cuomo2022scientific}, as well as other challenges \cite{wang2022and,rohrhofer2022understanding} which still impede the robustness of PINNs. Therefore, there is a true need for PINN-focused studies that address convergence and numerical performance.

The studies which are concerned with proving the convergence of PINNs are relatively sparse. Karniadakis and collaborators were among the first to provide theoretical justification of the PINN generalization error convergence in the training sample limit \cite{shin2020convergence,shin2020error}. Jiao et al. \cite{jiao2022rate} established upper bounds on the number of training samples as well as depth and width of ReLU$^{3}$ PINNs, while Mishra and Molinaro \cite{mishra2022estimates} established a framework to derive upper bound estimates of the PINN generalization error. The error minimization capability of PINNs was established mathematically in these proofs. However, their derivations do not account for the common optimizers, e.g. Adam and L-BFGS, which may raise challenges associated with the non-convex nature of the optimization landscape \cite{shin2020error,shin2020convergence,cuomo2022scientific}. One of the first theoretical studies that accounted for the optimizer's role within the training process is the Neural Tangent Kernel (NTK) approach \cite{wang2022and}, but its application is still limited to single-layer networks trained with gradient descent. Therefore, there is a true need for further investigations concerned with the convergence of PINNs, especially while accounting for the optimizer's effects within the common MLP-PINN setup.

Considering the hyperparameter landscape, there is only a handful of available studies which have attempted to explore the \textit{optimal} network parameters for PINNs. Wang et al \cite{wang2022auto} applied a Neural Architecture Search (NAS) with automated hyperparameter optimization (HPO), while the effect of hyperparameters choice was examined more manually in the work of Markidis \cite{markidis2021old}. Escapil et al \cite{escapil2022hyper} applied an HPO tuning process via Gaussian processes-based Bayesian optimization. A major drawback of these methods is the additional computational cost. Network training is an expensive optimization problem, and if one needs to establish a higher level optimization, parameter search, or other expensive analysis for every new PINN training this may end up defeating the purpose of using PINNs for computational physics and mechanics applications in the first place. Ideally, one should have some \textit{a priori} understanding of expected trends in terms of the selected architecture and the factors of predictive accuracy, computational effort, or chances of complete training failure. Therefore, there is a true need to establish well-engineered optimal guidelines for the PINN training in view of these factors. 


\subsection{Scope and Outline}
\label{Scope_and_Outline}

PINNs have thus far been mainly used as a standalone computational tool for PDEs in the field of Scientific Machine Learning (SciML). Bringing PINNs into the engineering world and utilizing them for practical applications requires substantial efforts, especially when we consider that many problems in the engineering practice are multi-scale and non-linear, and therefore very challenging from a computational standpoint. Consequently, we advocate the synergistic use of PINNs with other methods of well-established robustness for engineering applications, such as FEM. As a result, we presented I-FENN \cite{pantidis2022integrated}, a framework where PINNs are deployed on a finite element-level basis to swiftly approximate a state variable of interest; hence, it alleviates the need for the expensive coupled and mixed FEM setups. The goal of I-FENN is to accelerate the solution of nonlinear solid mechanics and multi-physics problems, and in \cite{pantidis2022integrated} it was applied in the context of non-local gradient damage \cite{peerlings1996gradient}. The integrated PINN targets the non-local strain gradient PDE, which is a second-order linear elliptic differential equation, and the PINN prediction is used within the iterative non-linear FEM setup for the calculation of damage. However, there are still several open questions which hinge upon the aforementioned gaps on PINNs, related to the network convergence and performance. These questions are presented in detail below:  

\begin{itemize}

\item {\bf{Research Objective 1: Error Convergence}} 

Does the error of the proposed PINN setup conform with the established convergence trends in the network size and training sample limit? 


\item \textit{\bf{Research Objective 2:} PINN engineering-guided Hyperparameter Search (HPS)}

What is the optimal PINN hyperparameter choice which exhibits maximum predictive accuracy, minimum computational effort, and least chances of trivial solutions?


\end{itemize} 

The first objective examines the sanity of our PINN from an error convergence perspective, in order to solidify the ground upon which our mechanics-inspired PINN is established. Answering this question is not trivial by any means. Instead, any newly proposed numerical method or variation needs to abide by this requirement, and PINNs should also follow this approach especially since their actual performance hinges upon the underlying optimizer. From this perspective, we investigate the response of both the training error and the global error with respect to the sizes of the network and the training dataset. The second objective is to identify engineering-guided optimal hyperparameters, and for this purpose we explore thoroughly the relationship between the network topology and the factors of predictive accuracy, computational effort and chances of complete training failure. All of the above factors constitute crucial criteria when selecting one PINN network over another, and we denote that such a decision needs to be made for any physics-based problem where PINNs are utilized as the PDE solution approximator. To answer both research questions, we launch a strategically designed analysis that is motivated and informed from the available mathematical theories on PINN convergence and performance. We explore the landscape shaped by all the major hyperparameters and we validate our findings by $closing \ the \ circle$ and integrating several from the obtained PINNs into I-FENN to cross-examine their performance.

The main novel contributions of this work are summarized as follows:
\begin{itemize}

    \item We establish a concise and systematic approach to analyze the PINN performance based on a novel combination of comprehensive metrics, which are informed from the available theoretical formulations and can be adapted to any PINN-based or similar framework.

    \item We provide sufficient numerical evidence of the PINN setup convergence against its complexity and the size of the training dataset.

    \item We establish a-piori knowledge of the hyperparameters which favor higher predictive accuracy, lower computational effort and least chances of arriving at trivial solutions for the non-local strain gradient PDE.

    \item We improve the rigor and robustness of the I-FENN framework by establishing its convergence, and we enhance its performance by thoughtfully selecting appropriate hyperparameters.
    
\end{itemize}

The paper is structured as follows. Section \ref{Sec:Theoretical_Overview} provides introductory remarks on the MLP-PINN setup and discusses existing convergence theories of PINNs. Section \ref{Sec:Methodology} introduces our methodology approach, while Sections \ref{Sec:Section_PINN_Error_Convergence} and \ref{Sec:PINN_Engineering_Guided_HPS} present our results with regards to the network error convergence and hyperparameter search respectively. We then validate these observations using different models in Section \ref{Sec:Section_Validation}, and we finally provide a summary of our conclusions in Section \ref{Sec:Discussion_Summary}.  

\section{Theoretical Overview}
\label{Sec:Theoretical_Overview}

\subsection{Deep Neural Networks}
\label{Sec:MLPs}

\begin{figure}[H]
	\centering
	\includegraphics[scale=0.5]{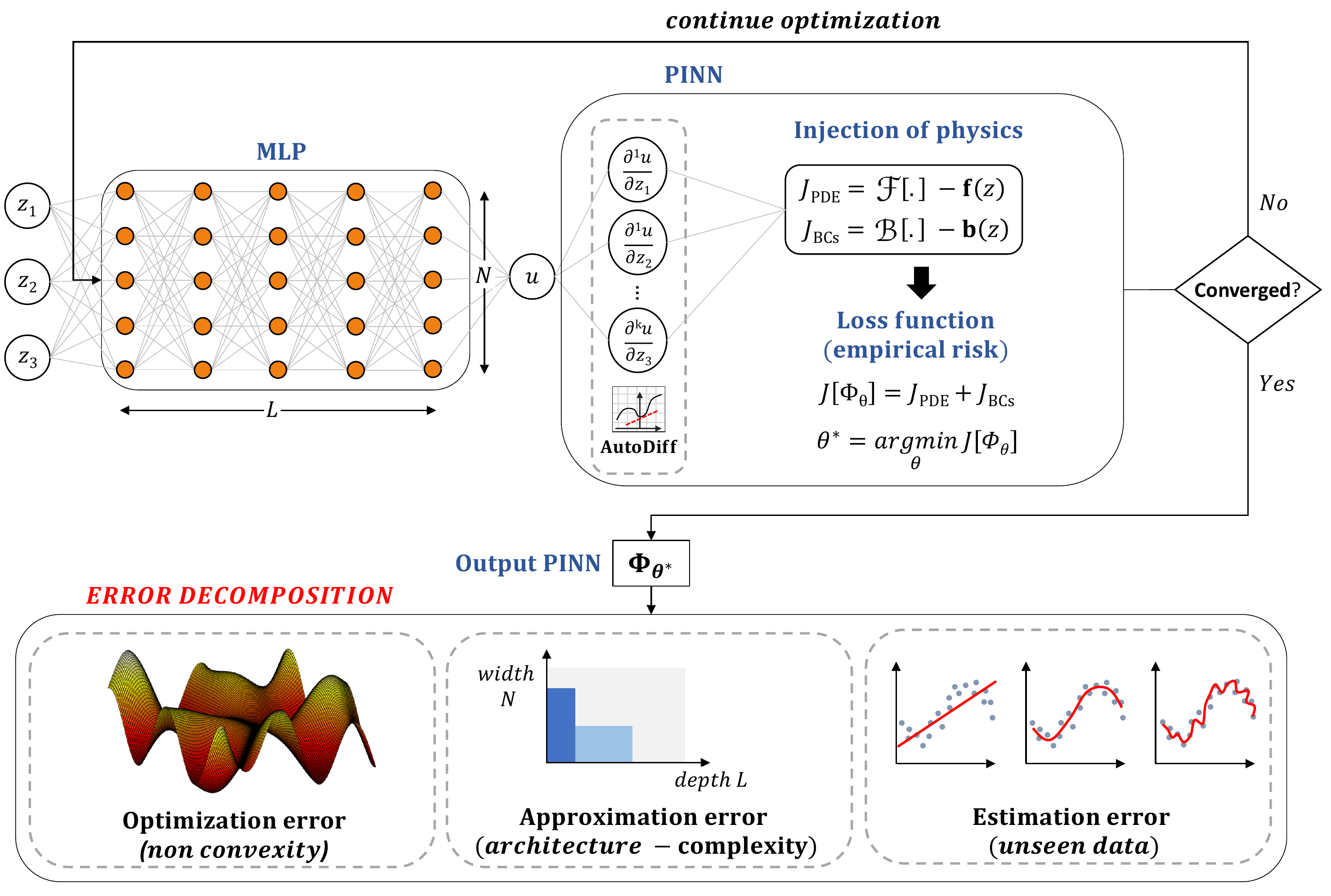}
	\caption{Schematic overview of the PINN setup and the global error decomposition}
	\label{Figure_PINNs_and_error_decomposition}
\end{figure}

Let us represent a feed-forward neural network - also termed multi-layer perceptron (MLP) - with a single output variable using the following notation. Let $L$ be the number of hidden layers,  $l = 1,...,L$ be the index of a hidden layer, $N_{0}, N_{l} \in \mathbb{N}$ and $1$ be the dimensions of the input, hidden and output layers, $z \in \mathbb{R}^{N_{0}}$ represent the input variables, $\alpha: \mathbb{R} \rightarrow \mathbb{R}$ be a non-linear \textit{activation} function, and $C_{l}$ be an affine-linear mapping $C_{l}: \mathbb{R}^{N_{l-1}} \rightarrow \mathbb{R}^{N_{l}}$ of the following form:

\begin{equation}
\begin{split}
     C_{l}(z_{l}) = W^{(l)} z_{l} + b^{(l)}
\end{split}
\label{MLP_eq1}
\end{equation}

\noindent where $W^{(l)}$ and $b^{(l)}$ are the weights matrix and bias vector of the $l$-th layer respectively. These constitute the adjustable \textit{parameters} $\theta^{(l)}$ of this layer. By denoting with the vector $\theta$ the total number of parameters across all layers, we can define the associated neural network $\Phi_{\theta}$: \cite{mishra2022estimates}:

\begin{equation}
\begin{split}
     \Phi_{\theta}(z) =  C_{L+1} \; \circ \; \alpha \; \circ \; C_{L} \; ... \; \alpha \; \circ \;  C_{2} \; \circ \; \alpha \; \circ \; C_{1}(z)
\end{split}
\label{MLP_eq2}
\end{equation}

\noindent where $\circ$ denotes composition of functions. From Equation \ref{MLP_eq2} it follows that information in MLPs is processed and transmitted in a forward and sequential fashion, through a recursive series of a) affine-linear computations across the successively stacked neurons and b) non-linear activation functions within the neurons \cite{mishra2022estimates,cuomo2022scientific}. A schematic overview of a sample MLP structure with $N, L = 5$ is shown in Figure \ref{Figure_PINNs_and_error_decomposition}. 

In the standard paradigm of supervised learning \cite{kutyniok2022mathematics,kotsiantis2007supervised}, one aims to approximate a known function $f(z)$ by minimizing the error between the network prediction $\Phi_{\theta}(z)$ and the target solution. Assuming a given sample of $M$ training points $(z_{i}, f(z_{i}))_{i = 1}^{M}$, the \textit{training error} $J$ is typically expressed as:

\begin{equation}
\begin{split}
    J(\Phi_{\theta}) = \frac{1}{M} \sum_{i=1}^{M} (\Phi_{\theta}(z_{i}) - f(z_{i}))^{2}
\end{split}
\label{MLP_eq3}
\end{equation}

\noindent and the goal is to find the optimum parameter vector $\theta^{*}$ over the admissible set of parameters $\Theta$, i.e.:

\begin{equation}
\begin{split}
    \theta^{*} = \arg \min_{\theta \in \Theta} (J(\Phi_{\theta}))
\end{split}
\label{MLP_eq4}
\end{equation}

Typical minimization techniques for this type of problem \cite{markidis2021old} are stochastic gradient descent methods \cite{robbins1951stochastic}, their more advanced variations such as Adam \cite{kingma2014adam} or AdaGrad \cite{duchi2011adaptive}, or higher-order methods such as the quasi-Newton L-BFGS algorithm \cite{liu1989limited}.

\subsection{Physics-Informed Neural Networks}
\label{Sec:PINNs}

PINNs is a machine learning technique which utilizes neural networks as the surrogate approximation for the solution of partial differential equations (PDEs). PINNs seek to find a neural network that satisfies PDEs of the following generic form:

\begin{equation}
\begin{split}
    {\mathcal{F}} ({\Phi}(z); \gamma) = {\bf{f}}(z) \; \; \; \; \;  z \; \; on \; \; \Omega 
\end{split}
\label{MLP_eq5}
\end{equation}

\begin{equation}
\begin{split}
    {\mathcal{B}} ({\Phi}(z)) = {\bf{b}}(z) \; \; \; \; \;  z \; \; on \; \; \Gamma
\end{split}
\label{MLP_eq6}
\end{equation}

\noindent where ${\mathcal{F}}$ is a non-linear differential operator, ${\mathcal{B}}$ represents the boundary/initial condition operator, $\bf{f}$ and $\bf{b}$ are point-wise known evaluations in the domain $\Omega$ and its boundary $\Gamma$ respectively, and $\gamma$ are physics-related parameters \cite{cuomo2022scientific}. PINNs are mainly used in the data-sparse regime, where in the absence of known data pairs they inject the physical laws described in Equations \ref{MLP_eq5} and \ref{MLP_eq6} into their objective function. Their approach is to minimize the PDE residual at selected collocation points inside and at the boundary of the domain, $m_{c}$ and $m_{b}$ respectively. The training error $J$ in this case typically reads as \cite{cuomo2022scientific}:

\begin{equation}
\begin{split}
    J = \underbrace{\frac{1}{m_{c}} \sum_{i=1}^{m_{c}} (\mathcal{F} (\Phi_{\theta}(z_{i})) - {\bf{f}}(z_{i}) )^{2}}_{J_{PDE}} + \underbrace{\frac{1}{m_{b}} \sum_{i=1}^{m_{b}} (\mathcal{B}(\Phi_{\theta}(z_{i})) - {\bf{b}}(z_{i}) )^{2}}_{J_{BCs}} 
\end{split}
\label{MLP_eq7}
\end{equation}


A schematic figure of the PINN setup is shown in Figure \ref{Figure_PINNs_and_error_decomposition}. We denote that the differential relationship between the output and input variables - which is necessary in order to encode the PDE residual - can be readily computed in most ML libraries using automatic differentiation \cite{rumelhart1986learning, baydin2018automatic}. 


\subsection{Convergence theories}
\label{Sec:Convergence_analysis}

\subsubsection{Relation to statistical learning error analysis}
\label{Sec:LearningErrorAnalysis}

Understanding how to better approximate the solution of a physical problem with a highly parameterized setup, such as the MLP/PINN formulation, poses several challenges. In order to establish a robust understanding of this aspect and resort to the most suitable tools and metrics, it is instructive to approach the process from a statistical learning error angle. Throughout our discussion we follow the notation of Shin et al. \cite{shin2020convergence}, where the MLP/PINN global error $\mathcal{E}$ is decomposed to the \textit{approximation} $\mathcal{E}_A$, \textit{estimation} $\mathcal{E}_E$ and \textit{optimization} $\mathcal{E}_O$ components, and we utilize the mathematical representations by Kutyniok \cite{kutyniok2022mathematics} and Cuomo et al. \cite{cuomo2022scientific} for each term.

The approximation error $\mathcal{E}_{A}$, which is the most well-studied of the three, refers to the network expressivity and its ability to approximate the target field of interest. A general expression of $\mathcal{E}_{A}$ is given as:

\begin{equation}
\begin{split}
     \mathcal{E}_{A} = \inf_{\theta \in \Theta} R(\Phi_{\theta})
\end{split}
\label{MLP_eq8}
\end{equation}

\noindent which is the greatest lower bound (infimum) of a network's risk over the parameter vector $\theta$. The most well-established result on this thread is the Universal Approximation Theorem (UAT), which states that given a suitable number of neurons, neural networks with even a single layer can approximate any continuous function on a compact domain with an arbitrary degree of accuracy \cite{hornik1989multilayer, chen1995universal, pinkus1999approximation, mhaskar2016deep, yarotsky2017error}. The essential message of the UAT is that as the neural network increases in size, its approximation capability will also increase \cite{yarotsky2018optimal,shen2019deep,de2021approximation}. In practice, the magnitude of this error is dictated by the network architecture and \textit{complexity}, a term which refers to the dimensionality component of the network \cite{schumann2010applications,cuomo2022scientific}. There are several measures of this feature \cite{raghu2017expressive, kutyniok2022mathematics}, such as the number of parameters \cite{shin2020error, shin2020convergence, yarotsky2017error} or the number of neurons, where the latter is dictated by the width and depth \cite{berner2021modern}. Here we note that characterizing the network complexity based on the width and depth has a twofold advantage compared to the number of parameters: a) it offers practical advice on the network design, and b) it is univocal, implying that given the depth and width, it is straightforward to compute the number of trainable parameters. On the contrary, there are many networks of different sizes and architectures that share the same number of parameters \cite{shen2019deep}. 

The estimation error $\mathcal{E}_{E}$ emerges when the trained network is evaluated against unseen data:
    
\begin{equation}
\begin{split}
     \mathcal{E}_{E} = \sup_{\theta \in \Theta} | R(\Phi_{\theta}) - \widehat{R}(\Phi_{\theta})| 
\end{split}
\label{MLP_eq10}
\end{equation}

\noindent which measures the distance between the empirical risk and the actual risk \cite{kutyniok2022mathematics}. Essentially, the estimation error provides a measure of how well the network performs in the test sample limit \cite{shin2020convergence, mishra2022estimates}. It is dominated by the quality of the training data and the definition of the objective function, and together with the approximation error they constitute the \textit{generalization} error $\mathcal{E}_{G}$ \cite{mishra2022estimates, shin2020convergence}.

Finally, the optimization error $\mathcal{E}_{O}$ is tied to the underlying optimization scheme, and it is given as:

\begin{equation}
\begin{split}
     \mathcal{E}_{O} = \widehat{R}(\Phi_{\theta^{*}}) - \inf_{\theta \in \Theta} \widehat{R}(\Phi_{\theta})
\end{split}
\label{MLP_eq9}
\end{equation}

\noindent which measures the accuracy with which the learnt neural network $\Phi_{\theta^{*}}$ minimizes the empirical risk \cite{kutyniok2022mathematics}.
Optimization is mostly carried out by gradient-descent or quasi-Newton based techniques, as they have empirically shown that they can converge to a suitable local minimum \cite{robbins1951stochastic,duchi2011adaptive,kingma2014adam,liu1989limited,shin2020convergence,kutyniok2022mathematics}. However, since the cost function is highly non-convex, these methods are not guaranteed to find a global minimizer. This is one of the widely open problems in the literature, and fundamentally understanding the workings and limitations of the training process is a subject of ongoing research \cite{berner2021modern}.

\subsubsection{Convergence estimates for PINNs}

Even though PINNs have already received extensive attention across a wide range of fields in computational science and engineering, it has not been until recently that theoretical justification of their convergence has been attempted. The majority of the theoretical work on PINNs has focused on providing bounds of the generalization error $\mathcal{E}_{G}$ for various classes of PDEs \cite{shin2020convergence,shin2020error,mishra2022estimates,de2021approximation,jiao2022rate,mishra2021physics}. One of the first studies strictly related to the consistency of the generalization error in the training sample limit was Shin et al. \cite{shin2020convergence}. In that study, the authors considered linear elliptic and parabolic PDEs and proved the following bound (Theorem 3.1 in \cite{shin2020convergence}, notation adapted for consistency):

\begin{equation}
\begin{split}
    \mathcal{E}_{G} \leq C_{m} \mathcal{E}_{T} + C' \left( m_{c}^{-\frac{a}{N_{0}}} + m_{b}^{-\frac{a}{N_{0}-1}} \right)
\end{split}
\label{Conv_eq3}
\end{equation}

\noindent where $0 < a \leq 1$, $C_{m}, \ C'$ are constants. We note that $\mathcal{E}_{T}$ is the training error ($J = \mathcal{E}_{T}$), and we use the former notation for consistency with the statistical error analysis approach. By postulating Eqn. \ref{Conv_eq3}, the authors illustrated how the generalization error is tied to the training error, and the theoretical and numerical results of this work demonstrated the error convergence against the increasing number of training data. The authors then extended their work in \cite{shin2020error} and established error estimates for loss functionals of residual minimization in continuous strong form, discrete strong form and continuous weak form (see Theorems 3.5, 4.7 and 5.8 respectively in \cite{shin2020error}). This body of work was restricted to linear PDEs. This limitation was relaxed in the work of Mishra and Molinaro \cite{mishra2022estimates} where the authors proposed a more abstract estimate of the generalization error for various classes of nonlinear PDEs (Eqn. 2.22 in \cite{mishra2022estimates}):

\begin{equation}
\begin{split}
     \mathcal{E}_{G} \leq C_{pde} \mathcal{E}_{T} + C_{pde} C_{quad}^{\frac{1}{p}} M^{-\frac{a}{p}} 
\end{split}
\label{MLP_eq13}
\end{equation} 

\noindent tying the generalization error to the training error, the number of training examples $M$ and the constants $C_{pde}$ and $C_{quad}$, for some $a > 0$ and $1 \leq p < \infty$. Similar-in-spirit bounds were provided in \cite{mishra2021physics,de2022error}. The common feature of the aforementioned estimates is that they examine the generalization error through the lens of the training error: \textit{as long as the training error is sufficiently small, and several other conditions are met, the generalization error will also be small} \cite{mishra2022estimates,mishra2021physics,de2022error}. 

Complementary to the above-mentioned theories, we highlight a new theoretical perspective on PINNs which explicitly accounts for the optimization process. This is the work by Wang et al. \cite{wang2022and} on the Neural Tangent Kernel (NTK) approach. The current implementations of the NTK focus on capturing the behavior of fully-connected neural networks in the infinite width limit during training via gradient descent. The authors of \cite{wang2022and} derived the NTK of PINNs and showed that the solution fields at the boundary and in the domain evolve as follows (Lemma 3.1 in \cite{wang2022and}, notation adapted for consistency):

\begin{equation}
\begin{split}
        \begin{bmatrix}
        \frac{d{\mathcal{B}} ({\Phi}(z))}{dt} \\
        \frac{d{\mathcal{F}} ({\Phi}(z))}{dt}
        \end{bmatrix} = -
        \begin{bmatrix}
        {\bf{K}}^{NTK}_{{\mathcal{B}}{\mathcal{B}}}(t) & {\bf{K}}^{NTK}_{{\mathcal{B}}{\mathcal{F}}}(t) \\
        {\bf{K}}^{NTK}_{{\mathcal{F}}{\mathcal{B}}}(t) & {\bf{K}}^{NTK}_{{\mathcal{F}}{\mathcal{F}}}(t)       
        \end{bmatrix} \cdot
        \begin{bmatrix}
        {\mathcal{B}} ({\Phi}(z)) - g(z) \\
        {\mathcal{F}} ({\Phi}(z)) - f(z) 
        \end{bmatrix} 
\end{split}
\label{NTK_eq1}
\end{equation} 

\noindent where the first matrix after the equal sign is the NTK and it denoted with ${\bf{K}}^{NTK}$. The authors of \cite{wang2022and} then introduced the relative change of the NTK and the parameter vector ${\Delta \theta}$ in order to monitor the PINN convergence during training:

\begin{equation}
\begin{split}
    \Delta K = \frac{|| {\bf{K}}^{NTK}_{i} - {\bf{K}}^{NTK}_{0} ||_{2}}{|| {\bf{K}}^{NTK}_{0} ||_{2}}
\end{split}
\label{DeltaKEqn_Wang}
\end{equation} 

\begin{equation}
\begin{split}
    \Delta \theta = \frac{|| \theta_{i} - \theta_{0} ||_{2}}{|| \theta_{0} ||_{2}}
\end{split}
\label{DeltaThetaEqn_Wang}
\end{equation} 

\noindent where the subscripts $0$ and $i$ indicate initialization and the $i$-th epoch  respectively. The study in \cite{wang2022and} showed that in the infinite-width limit of shallow networks, both metrics tend asymptotically to zero during training with gradient descent at an infinitesimally small learning rate. Thus, the PINN training error can be analyzed through its NTK at initialization, but we denote that this work has certain limitations due to its idealized setup (single-layer networks of infinite width, very low learning rate, gradient descent optimizer).

All these efforts have undoubtedly solidified the theoretical advancement of PINNs, however they all share an important deficit: they have been derived while overlooking the role of optimizers used in practice for PINNs, such as Adam and L-BFGS. The optimization process shapes a highly non-convex landscape, and even though a low training error can be an indicator of a small generalization error, there is currently no guarantee that the global minimization will be achieved. Consequently, if one wishes to get an insight on the actual performance of PINNs, then a numerical investigation is currently the only viable pathway given the available theoretical understanding of PINNs' performance. Even in this case, the conclusions of such an attempt can be considered robust only if all the following steps are taken: a) examine convergence against both the network complexity and dataset size, b) monitor both the training error and the global error of the solution, c) repeat several idealizations of the same configuration to explore various optimization paths and smooth out the effect of the parameters initialization, d) examine different shape configurations, training epochs and learning rate values to understand the impact of each factor, and e) monitor the optimizer path to convergence by assessing the network's behavior during training. These guidelines formulate the backbone of our methodology which is presented in detail in the next section, and it can be adapted to any PINN-based framework.

\section{Methodology}
\label{Sec:Methodology}

The overarching goal of this paper is to delve into the PINN component of I-FENN and establish a robust understanding of its error convergence trends and its hyperparameter-performance relationship. In this section we lay out our methodology, which relies on conducting a thorough and systematic numerical investigation to answer both objectives. We first present an overview of I-FENN and its application on the non-local gradient damage equation; the reader is referred to \ref{Appendix:nonlocalGradientTheory} and \ref{Appendix:IFENN} for more fundamental details. We then define the search space, metrics, and models of our investigation.

\subsection{I-FENN on non-local gradient damage mechanics}
\label{Sec:IFENN_on_non-localGradientTheory}


I-FENN is a newly established framework which relies on the synergistic action of two computational approaches to approximate the solution of partial differential equations: FEM and PINNs. The PINN targets one of the problem's governing PDEs and its corresponding state variable, and it is trained offline to compute this variable and its derivative. Then, in the online stage, the PINN is embedded directly in the finite element-level stiffness function, receives the deformation field as input information, and its output products are used together with the FEM shape functions to compute element-level quantities such as the Jacobian matrix and the element residual. The latter two quantities allow for the re-calculation of the displacement field using an iterative scheme such as Newton-Raphson or the arc-length method, and the entire online process is utilized repeatedly within an iterative non-linear solver until the convergence criterion is satisfied. Therefore, the conceptual novelty of the method is that only the displacement field is solved for as an FEM nodal variable, while the other variable of interest is computed at the element-level at a glance due to the swift predictive capability of the pre-trained PINN. 

\begin{figure}[H]
	\centering
	\includegraphics[width=\linewidth]{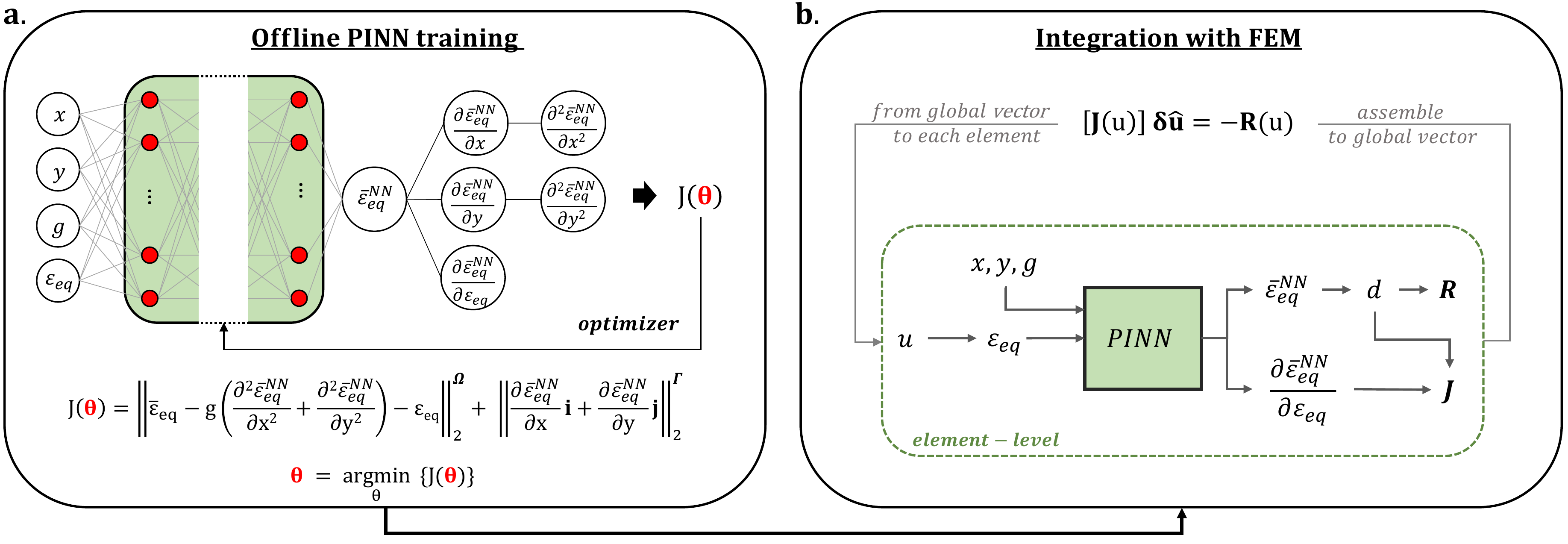}
	\caption{Overview of I-FENN setup. {\bf{a}}: The PINN is trained offline and learns the transformation mapping between the local strain and non-local strain fields. {\bf{b}}: The pre-trained PINN is embedded inside the stiffness function at the finite element-level, receives as input the current deformation state, and its output quantities are utilized in the calculation of the damage variable $d$, Jacobian matrix \textbf{J}(u) and residual vector \textbf{R}(u).}
	\label{Figure_Methodology_Flowchart}
\end{figure}

To this end, I-FENN has been applied in the context of non-local damage \cite{pantidis2022integrated} utilizing the gradient non-local strain transformation by Peerlings et al. \cite{peerlings1996gradient}:

\begin{equation}
\begin{split}
    \nabla \cdot \sigma = 0 \; \; \; in \; \; \Omega
\end{split}
\label{nonlocalGradientPDEBCs}
\end{equation}

\begin{equation}
\begin{split}
    {\bar{\varepsilon}}_{eq} - g \nabla^{2} {\bar{\varepsilon}}_{eq} - \varepsilon_{eq} = 0 \; \; \; in \; \; \Omega
\label{nonlocalGradientPDE}
\end{split}
\end{equation}

\begin{equation}
\begin{split}
    \nabla {\bar{\varepsilon}}_{eq} \cdot n_{i} = 0 \; \; \; in \; \; \Gamma
\end{split}
\label{nonlocalGradientPDEBCs}
\end{equation}

\noindent where ${\varepsilon}_{eq}$ and ${\bar{\varepsilon}}_{eq}$ are the local and non-local equivalent strain fields respectively, and $\nabla^2$ is the Laplacian operator. Figure \ref{Figure_Methodology_Flowchart} provides a schematic overview of the I-FENN setup for non-local gradient damage. As shown in Figure \ref{Figure_Methodology_Flowchart}a, the input variables of the fully-connected network are the coordinates of each material point ($x$ and $y$ for a 2D case), the length scale measure $g$ and the local equivalent strain $\varepsilon_{eq}$. The output variable is the non-local equivalent strain $\bar{\varepsilon}_{eq}^{NN}$, its first and second order derivatives with respect to the coordinates $\frac{\partial{\bar{\varepsilon}}_{eq}^{NN}}{\partial{x}}$, $\frac{\partial{\bar{\varepsilon}}_{eq}^{NN}}{\partial{y}}$, $\frac{\partial^{2}{{\bar{\varepsilon}^{NN}_{eq}}}}{\partial{x}^{2}}$, $\frac{\partial^{2}{{\bar{\varepsilon}^{NN}_{eq}}}}{\partial{y}^{2}}$ and its first order derivative with respect to the local equivalent strain $\frac{\partial \bar{\varepsilon}^{NN}_{eq}}{\partial \varepsilon_{eq}}$. The non-local equivalent strain is used to compute the damage variable $d$ and the residual vector \textbf{R}(u). The four partial derivatives with respect to the coordinates are used in the cost function definition, as shown in Figure \ref{Figure_Methodology_Flowchart}a, whereas $\frac{\partial \bar{\varepsilon}^{NN}_{eq}}{\partial \varepsilon_{eq}}$ is used in the definition of the Jacobian matrix \textbf{J}(u). The converged solution of the I-FENN setup is achieved by the iterative solution of the non-linear system of equations which aims at minimizing the residual \textbf{R}(u), see \ref{Appendix:IFENN} for more details. A comparison of the differences between the local damage, non-local gradient damage and its I-FENN implementation is shown in Table \ref{Table_Methodologies_Comparison}.

\begin{table}[H]
	\caption{Comparison between the three damage frameworks: local \cite{kachanovbook}, non-local gradient \cite{peerlings1996gradient} and I-FENN \cite{pantidis2022integrated}}
	\label{Table_Methodologies_Comparison}\centering
	\begin{adjustbox}{width=\textwidth}
	\begin{tabular}{c||c|c|c}
		\toprule
		& Local & Non-local gradient & I-FENN \\
		\midrule

        $\bf{\# DOFs}$ (2D) & $2 \times nodes$ & $3 \times nodes$ & $2 \times nodes$ \\
        
        \multirow{2}{*}{${\bf{R}}$ (Residual)} & \(\displaystyle
        R = \int_{\Omega} w^u \left[\left[C_{ijkl}(d)\varepsilon_{kl}\right]_{,j}\right] d\Omega \) & \(\displaystyle
        R^u = \int_{\Omega} w^u \left[\left[C_{ijkl}(d)\varepsilon_{kl}\right]_{,j}\right] d\Omega \) & \(\displaystyle
        R = \int_{\Omega} w^u \left[\left[C_{ijkl}(d)\varepsilon_{kl}\right]_{,j}\right] d\Omega \) \\
        &  & \(\displaystyle R^{\bar{\varepsilon}} = \int_{\Omega} w^{\bar{\varepsilon}} \left[\bar{\varepsilon}_{eq} - g \bar{\varepsilon}_{eq,ii}-\varepsilon_{eq}\right] d\Omega \)   \\
        
		${\bf{J}}$ (Jacobian) & $\begin{bmatrix} \frac{\partial{\bf{R}}}{\partial{\bf{\hat{u}}}} \end{bmatrix}$ & 
		$\begin{bmatrix}
        \frac{\partial{\bf{R}}^{u}}{\partial{\bf{\hat{u}}}} & \frac{\partial{\bf{R}}^{u}}{\partial{\bf{\hat{\bar{\varepsilon}}}}} \\
        \frac{\partial{\bf{R}}^{\bar{\varepsilon}}}{\partial{\bf{\hat{u}}}} & \frac{\partial{\bf{R}}^{\bar{\varepsilon}}}{\partial{\bf{\hat{\bar{\varepsilon}}}}}
        \end{bmatrix}$ & $\begin{bmatrix} \frac{\partial{\bf{R}}}{\partial{\bf{\hat{u}}}} \end{bmatrix}$  \\
		
		$\bf{Dependencies}$ & \(\displaystyle
        \frac{\partial{d}}{\partial \hat{u}_{k}} = 
        \frac{\partial{d}}{\partial{{\varepsilon}_{eq}}}       \; 
        \frac{\partial{\varepsilon_{eq}}}{\partial{\varepsilon_{ij}}}               \; 
        \frac{\partial{\varepsilon_{ij}}}{\partial{\hat{u}}_{k}}\) & 
        
        \(\displaystyle
        \frac{\partial{d}}{\partial \hat{\bar{\varepsilon}}_{eq}}, \frac{\partial{\varepsilon_{eq}}}{\partial{\hat{u}}_{k}} =  \frac{\partial{\varepsilon_{eq}}}{\partial{\varepsilon_{ij}}}               \; 
        \frac{\partial{\varepsilon_{ij}}}{\partial{\hat{u}}_{k}} \)
        
        & \(\displaystyle
        \frac{\partial{d}}{\partial \hat{u}_{k}} = 
        \frac{\partial{d}}{\partial{{\bar{\varepsilon}}^{NN}_{eq}}}                \; 
        \frac{\partial{\bar{\varepsilon}^{NN}_{eq}}}{\partial{{\varepsilon}_{eq}}}       \; 
        \frac{\partial{\varepsilon_{eq}}}{\partial{\varepsilon_{ij}}}               \; 
        \frac{\partial{\varepsilon_{ij}}}{\partial{\hat{u}}_{k}}\) \\
        \bottomrule
	\end{tabular}
	\end{adjustbox}
\end{table}

\subsection{Search space} 
\label{Sec:SearchSpace}

Here we define the search space which is common for both objectives stated in Section \ref{Scope_and_Outline}, and we then proceed to the specific constraints of each one. Henceforth, the network shape is denoted as $L \times N$, and we use the term \textit{network aspect ratio}, $AR = N/L$, to describe the relationship between the network width $N$ and depth $L$.

The PINN training is conducted using the Adam optimizer \cite{kingma2014adam} followed by L-BFGS \cite{liu1989limited}, which is a common practice in the PINN literature \cite{markidis2021old,cuomo2022scientific}. We consider two values for the Adam learning rate, $lr = 10^{-3}$ and $lr = 10^{-4}$. For each learning rate we consider two cases of Adam training epochs,  $ep = 5000$ and $ep = 10000$. L-BFGS is operated in all cases until a pre-defined numerical tolerance criterion is satisfied. Henceforth we use the notation $c = [L,N,ep,lr]$ to refer to a case combination $c$. The network parameters are initialized using the Xavier initialization method \cite{glorot2010understanding}, and the hyperbolic tangent $tanh()$ is used as the activation function. For all analyzed cases, we perform 10 independent idealizations of that configuration, and we report the average values. This is to ensure independence from the randomness induced by the weights initialization, while still maintaining a reasonable computational effort. The above setup is common to both research objectives, that of the PINN error convergence and the PINN engineering-guided HPS. 

\subsubsection{PINN error convergence}
Regarding the PINN error convergence analysis, we fix the aspect ratio at $AR = 1$ and we increase equally $L$ and $N$ to obtain the following configurations: $8 \times 8$, $12 \times 12$, $20 \times 20$ and $32 \times 32$. Each network configuration is trained on all the available mesh resolutions of the investigation and validation models, which are shown later in this section. This allows us to deduce conclusions on the error convergence against both the network complexity and the training sample size. The results of this investigation are reported in Section \ref{Sec:Section_PINN_Error_Convergence}, and we focus on both the training error $\mathcal{E}_{T}$ and the global error $\mathcal{E}$ as discussed in the previous section.

Here we remark that defining a robust complexity measure is not straightforward. In the study by \cite{safran2017depth}, the authors reported a significantly better performance on both the training and the validation dataset using a network with 3 layers and 100 neurons versus a network with 2 layers and 800 neurons, even though the second network has a drastically larger number of trainable weights and neurons. This example is mentioned here to illustrate that defining complexity is a challenge on its own, and to emphasize the impact of additional layers in the network accuracy. To enhance the robustness of our approach, we increase monotonically both the width and depth, and our four configurations are increasingly more complex whether one measures complexity by the number of adjustable parameters, non-linear operational units (neurons), or network dimensions (width or depth). We note that the $AR$ constraint is the essential key to navigate the complexity space more steadily and deduce robust conclusions. 

\subsubsection{PINN hyperparameter search}

\begin{figure}[H]
	\centering
	\includegraphics[width=\linewidth]{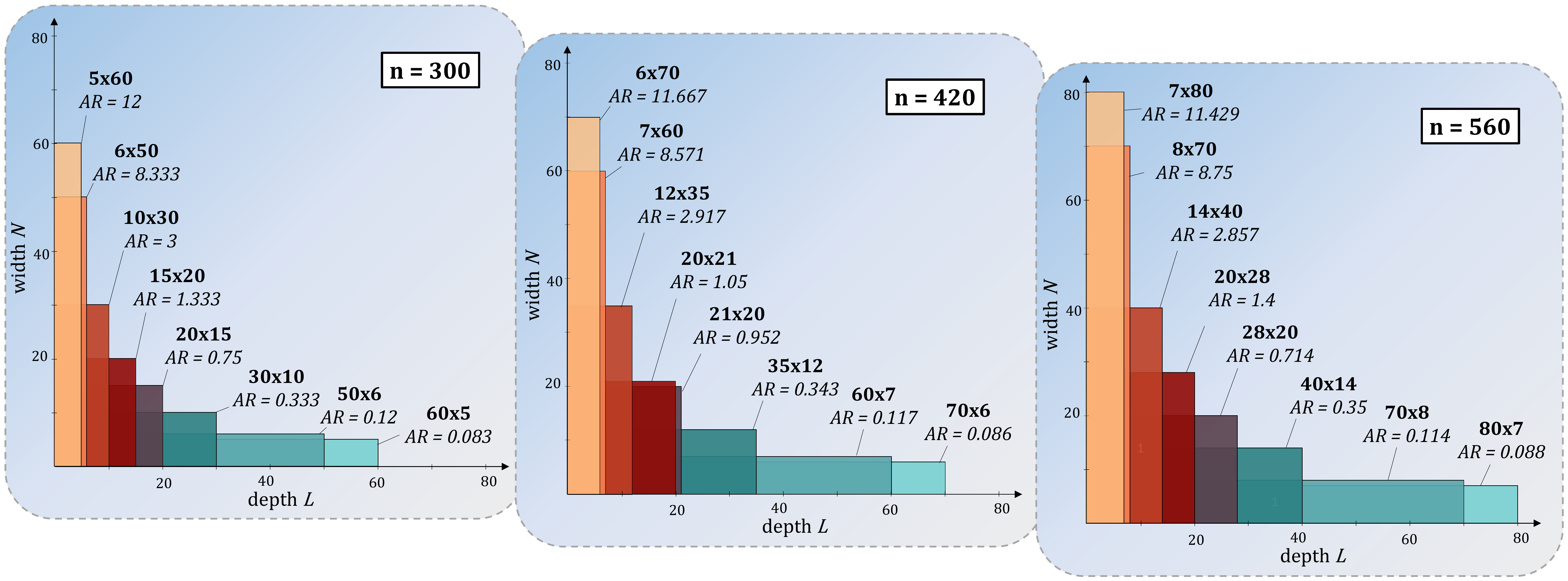}
	\caption{Shape and aspect ratio ($AR$) values of the networks investigated in Section \ref{Sec:PINN_Engineering_Guided_HPS}. The shape is reported as the depth $L$ (no. layers) $\times$ the width $N$ (no. neurons per hidden layer) of the network. The aspect ratio is the ratio of the width over the depth. For example, a network with 15 layers and 20 neurons is denoted as 15x20 and has an aspect ratio $AR = 1.333$.}
	\label{Figure_Search_Space}
\end{figure}

For the PINN HPS investigation, the following constraints are applied. First, we choose to characterize complexity by the total number of neurons $n$ in the hidden layers. This choice is motivated by all the previous discussions and the practical fact that one would design a network more conveniently based on this variable compared to the number of trainable parameters. We consider three cases of network complexity, $n = 300, 420, 560$. For each fixed value of $n$, we vary the network width and depth such as only its aspect ratio changes. For each $n$, we consider 8 cases of aspect ratio values. The shapes and associated aspect ratio values are reported in Figure \ref{Figure_Search_Space}. The networks in all cases retain their rectangular shape. This investigation is conducted against one mesh discretization from each model and our results are reported in Section \ref{Sec:PINN_Engineering_Guided_HPS}.

%
%
%
%
%
\subsection{Performance metrics} 
\label{Performance_metrics} 

We use the following metrics to determine the PINN accuracy:

\begin{itemize}

\item The value of the PINN training error $J$ which is minimized during the optimization process. This is equivalent to the training error $\mathcal{E}_{T}$ as mentioned in the previous section. 
    
\item The L2-norm of the non-local strain relative prediction error $\bar{\varepsilon}_{L2RSE}$, which is defined as follows:

\begin{equation}
\begin{split}
{\bar{\varepsilon}}_{L2RSE} = \sqrt{\sum_{i = 1}^{m_{b}} {\frac{(  {\bar{\varepsilon}}_{i,pred} - {\bar{\varepsilon}}_{i,true})^2}{({\bar{\varepsilon}}_{i,true})^2}}}
\label{non-localStrain_L2RSE}
\end{split}
\end{equation} 

This measure is used to estimate the PINN global error $\mathcal{E}$ since it takes into consideration the predicted and target solution fields, ${\bar{\varepsilon}}_{pred}$ and ${\bar{\varepsilon}}_{true}$ respectively. We emphasize that this value takes inherently into account the contribution of the optimizer, since it is a numerical byproduct of the optimization process. 

\end{itemize}

Additionally, we inspect the performance of the Adam optimizer by observing:

\begin{itemize}

\item The relative change of the parameter vector $\theta$, denoted as $\Delta\theta$

\begin{equation}
\begin{split}
    \Delta \theta = \frac{|| \theta_{i} - \theta_{i-1} ||_{2}}{|| \theta_{i-1} ||_{2}}
\end{split}
\label{DeltaThetaEqn}
\end{equation} 

\noindent where $\theta$ is recorded every 100 Adam epochs. This metric is similar in spirit with the one used in \cite{wang2022and} and reported in Eqn. \ref{DeltaThetaEqn_Wang}. The only difference is that it is calculated across sequential measurements and not with respect to the initialized vector. Its goal is to provides an insight on the convergence of the PINN adjustable weights during training. 

\item The slopes of the best fit lines of $\Delta \theta$, which are denoted as $s_{\Delta \theta}$. A close-to-zero value for this metric indicates that a) either the network has reached a local minimum and it is incapable of additional learning, with the weights oscillating in the optimization landscape, or b) a very slow learning rate is used. Overall, since this metric directly targets the learnt parameters, it is a very useful diagnostic tool to monitor the robustness of the learning process. We note that in the computation of $s_{\Delta \theta}$ we always omit the first entry of the $\Delta \theta$ vector which corresponds to initialization. This outlier value heavily skews the computation of the slope, without adding significance to it.

\end{itemize}

In this study, we attempted to calculate the NTK of our PINN configurations as another diagnostic tool of the network convergence \textit{during} training; however, the computational demands exceeded the available resources of computational memory, even when this was attempted on the High-Performance Cluster facility of NYUAD. We therefore remark this current limitation for the NTK analysis which is even more pronounced in the case of deeper networks, and we highlight the need for additional work by the community along this direction.

Finally, we note that estimating the computational efficiency of a PINN configuration is of paramount importance for any practical application. Below, we provide the means by which we assess the computational effort, and we note that all time-related values have been measured on a Dell G5 5500 machine with an Intel i7-10750H CPU and Nvidia GeForce GTX 1650Ti GPU acceleration:

\begin{itemize}

\item The running time of 100 Adam epochs, denoted as $Adam_{RT100}$. The total number of Adam epochs is typically user-defined and it is in the order of thousands \cite{kingma2014adam,markidis2021old,cuomo2022scientific}. Therefore this measure provides a building-block level speed estimate for the Adam stage.

\item The number of L-BFGS iterations, denoted as $Iter_{LBFGS}$. Typically L-BFGS is operated until the convergence criterion of the algorithm is satisfied \cite{liu1989limited}, and therefore the total number of L-BFGS iterations is simulation-dependent.

\item The number of I-FENN convergence iterations once the trained PINN is embedded in the finite element numerical solver, denoted as $Iter_{IFENN}$. This is a very important additional layer of time efficiency check, since we are interested in the total computational effort of I-FENN and not just the offline PINN training stage.

\end{itemize}
Table \ref{Table_Error_Metrics} presents the set of metrics which is used to evaluate the PINN performance. In total, we report the results from 640 trained PINNs for the error convergence objective and 1120 trained PINNs for the HPS objective.

\begin{table}[H]
    \caption{Performance metrics: symbols, definitions and purpose}
    \label{Table_Error_Metrics}\centering
    \begin{tabular}{c c c}   
	\toprule

        $\bf{Symbol}$                   &  \bf{Definition}          &  \bf{Purpose}     \\  
        
        \midrule
        
        $\mathrm{J}$                            &  PINN training error      &  \multirow{2}{*}{Predictive accuracy}     \\  
        ${\bar{\varepsilon}}_{L2RSE}$  &  PINN global error        &       \\ 
        \hline
        $\Delta \theta$                &  Relative change of parameters during training      &  \multirow{2}{*}{PINN training convergence}     \\  
        $s_{\Delta \theta}$            &  Slopes of best fit lines of $\Delta \theta$ curves &   \\  
        \hline
        $\mathrm{Adam_{RT100}}$        &  Running time of 100 Adam epochs          &  \multirow{3}{*}{Computational cost}     \\ 
        $\mathrm{Iter_{LBFGS}}$        &  No. of L-BFGS iterations until convergence  &       \\  
        $\mathrm{Iter_{IFENN}}$        &  No. of Newton-Raphson iterations in I-FENN  &       \\

    \bottomrule
	\end{tabular}
\end{table}

\subsection{Investigation and Validation Models}
\label{Sec:Investigation_and_Validation_Models}

\begin{figure}[H]
	\centering
	\includegraphics[width=\linewidth]{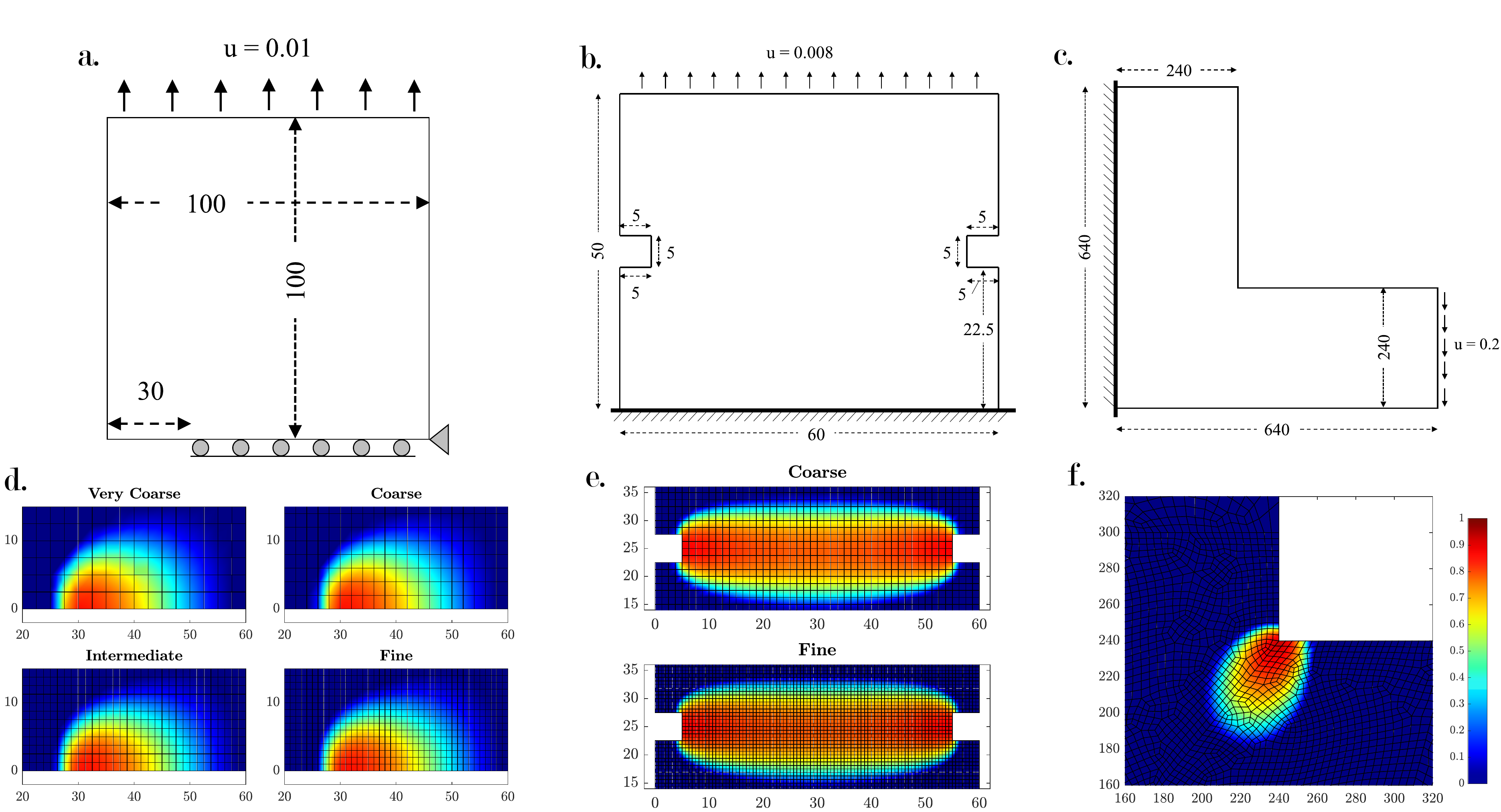}
	\caption{{\bf{a-c}}: Geometry, boundary and loading conditions of the investigation model (single-notch) and the validation models (double-notch and L-shaped specimens). {\bf{d-f}}: Non-local damage profiles for each  model at the loadfactor of interest. Images are adapted from \cite{pantidis2022integrated}.}
	\label{Figure_Models_details}
\end{figure}

We use the following models for our study: an investigation model, which is used for the complete exploration of the search space, and two additional validation models. Below we provide a brief description, and the interested reader is referred to \cite{pantidis2022integrated} for more details. The investigation model is a a single-notch specimen under a mode-I loading setup, as shown in Figure \ref{Figure_Models_details}a. The bottom rollers constrain only the y-displacement, and the pin at the bottom right node constrains both the x- and y-displacements. Plane strain conditions are assumed, and we apply four different mesh discretizations. The resulting models are termed as \textit{Very Coarse}, \textit{Coarse}, \textit{Intermediate} and \textit{Fine}, and they each have 1600, 2500, 6400 and 10000 square finite elements respectively. Figure \ref{Figure_Models_details}d shows the non-local damage profile in a zoomed-in region around the crack tip at the loadfactor of interest for all models. Evidently, the damage field is consistent across all idealizations which verifies the mesh-independent nature of the non-local solver. 

Complementary to the complete investigation of the single-notch case, we perform sample checks in two validation models. Plane strain conditions are considered for both. The first validation model is the double-notch specimen shown in Figure \ref{Figure_Models_details}b. The bottom edge is constrained in both directions and uniaxial tension is applied across the top edge. Two mesh discretizations are utilized, and the resulting models are termed \textit{Coarse} and \textit{Fine}, with 1888 and 7552 finite elements respectively. Figure \ref{Figure_Models_details}e shows the non-local damage profile at the investigated loadfactor for both models, which is evidently independent of the mesh resolution. The second validation model is the L-shaped domain shown in Figure \ref{Figure_Models_details}c. The left edge is constrained in both directions and the outer right edge is pulled downwards. An unstructured mesh with refined resolution around the anticipated location of the damage zone is used. Figure \ref{Figure_Models_details}f shows the non-local damage profile at the loadfactor of interest, where for the sake of clarity only a close-up view of the damage zone is shown.

\section{PINN error convergence}
\label{Sec:Section_PINN_Error_Convergence}

The first goal of this study is to establish that our PINN setup exhibits convergence to a more accurate solution as either the network size or the training sample size increase, and here we present the results of this investigation. 

\subsection{Error convergence analysis}
\label{Sec:Error_convergence_analysis}

\begin{figure}[H]
	\centering
	\includegraphics[scale=0.55]{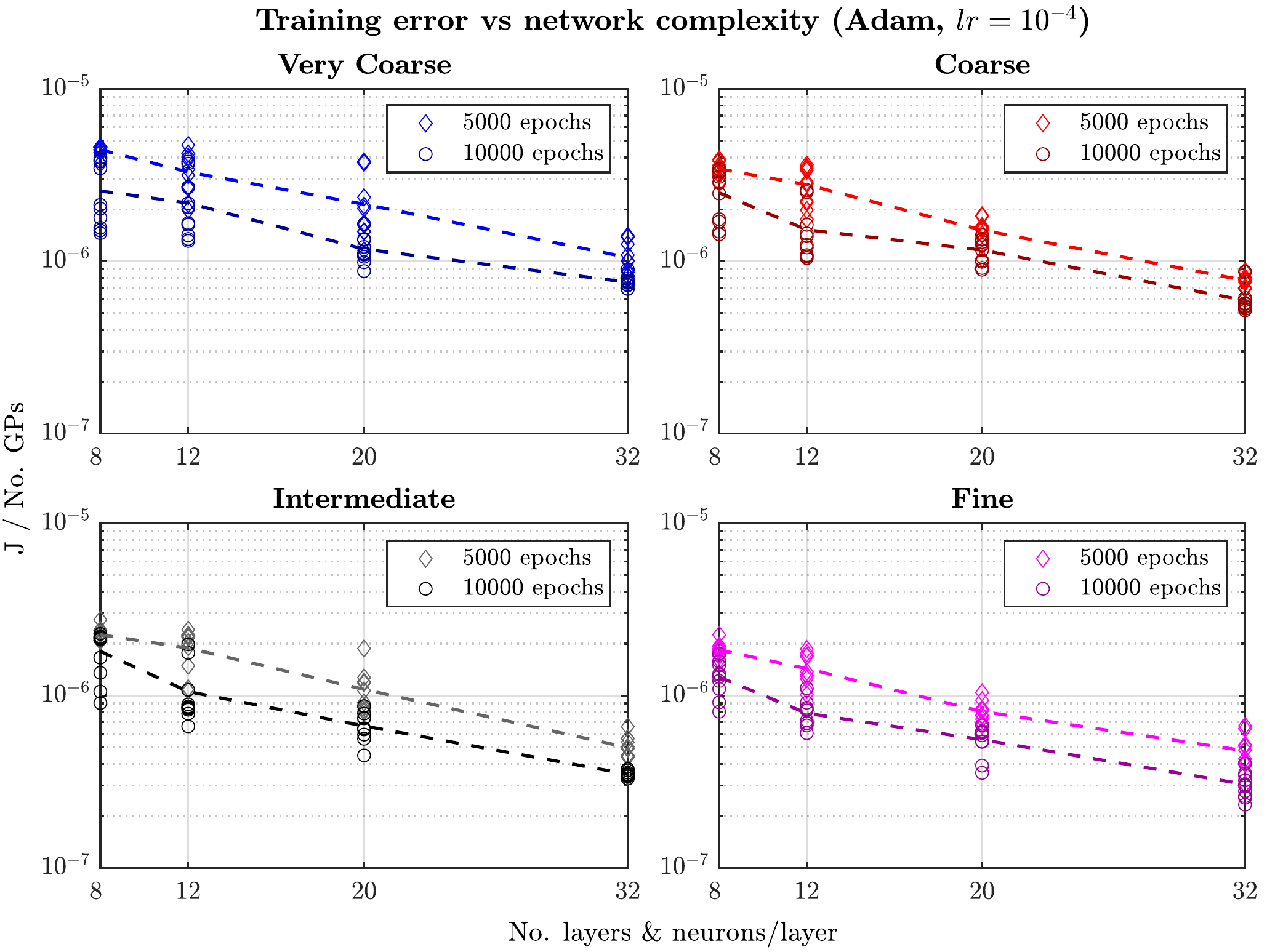}
	\caption{Training error $J$ convergence with increasing network complexity. The values are taken at the end of Adam and for $lr = 10^{-4}$ across all mesh idealizations, and they are normalized over the number of Gauss points of each mesh. The markers correspond to 10 independent simulations, and the dashed lines connect their average. All networks are square.}
	\label{Figure_Convergence_JvsLayers_Adam_lowlr}
\end{figure}

\begin{figure}[H]
	\centering
	\includegraphics[scale=0.55]{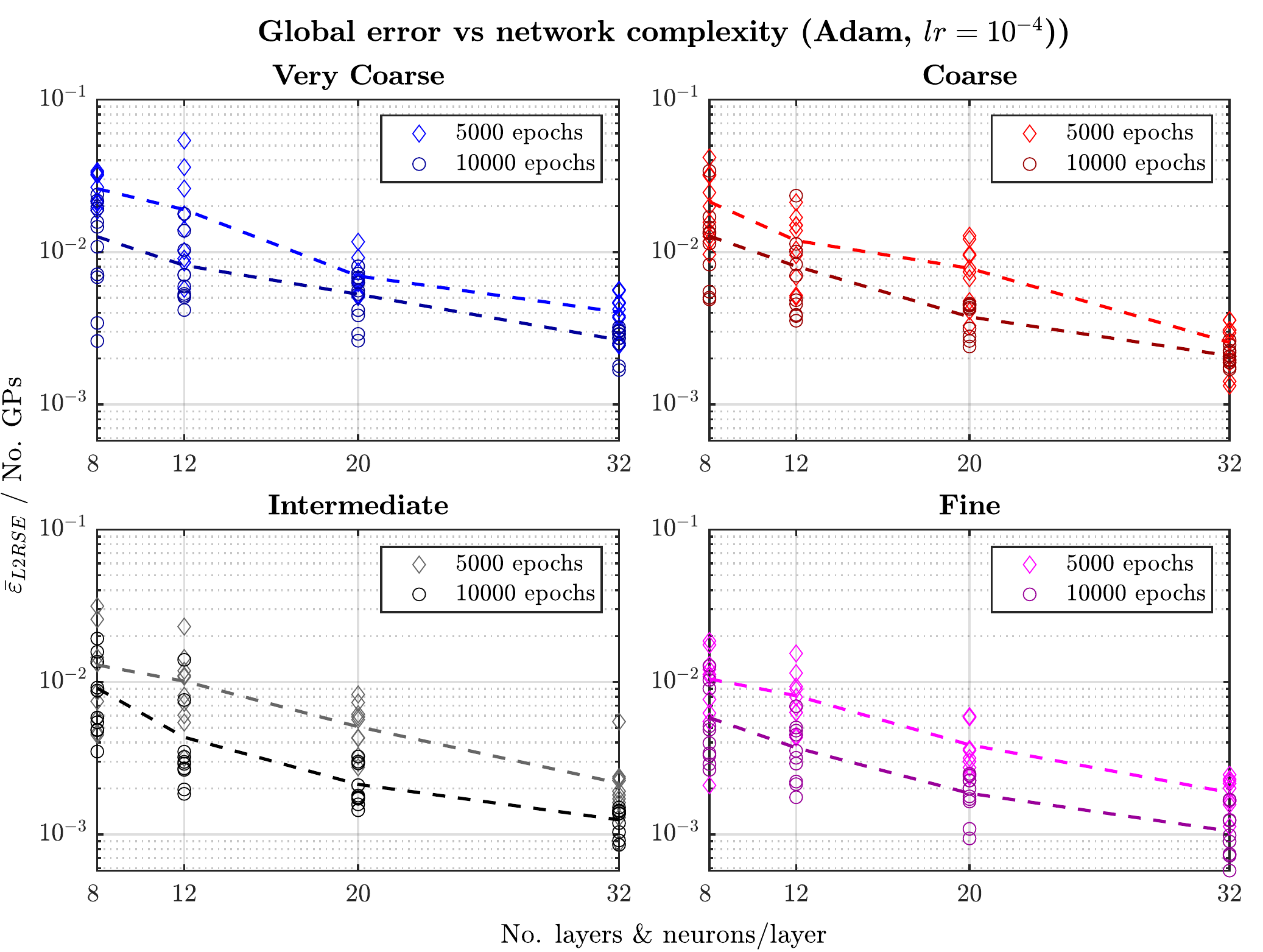}
	\caption{Global error ${\bar{\varepsilon}}_{L2RSE}$ convergence with increasing network complexity. The values are measured at the end of Adam and for $lr = 10^{-4}$ across all mesh idealizations, and they are normalized over the number of Gauss points of each mesh. The markers correspond to 10 independent simulations, and the dashed lines connect their average. All networks are square.}
	\label{Figure_Convergence_L2RSEvsLayers_Adam_lowlr}
\end{figure}

\begin{figure}[H]
	\centering
	\includegraphics[scale=0.55]{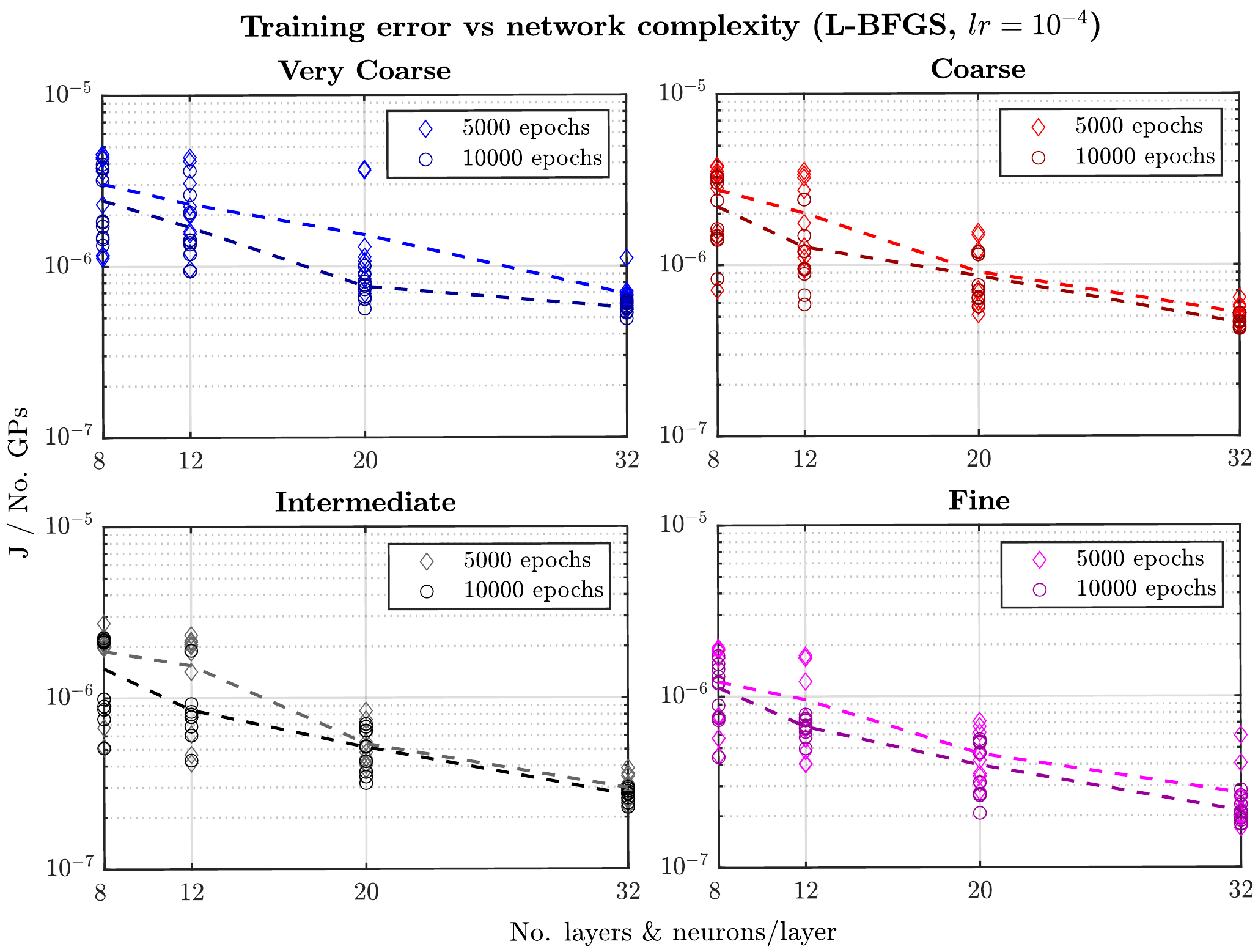}
	\caption{Training error $J$ convergence with increasing network complexity. The values are taken at the end of L-BFGS and for $lr = 10^{-4}$ across all mesh idealizations, and they are normalized over the number of Gauss points of each mesh. The markers correspond to 10 independent simulations, and the dashed lines connect their average. All networks are square.}
	\label{Figure_Convergence_JvsLayers_LBFGS_lowlr}
\end{figure}

\begin{figure}[H]
	\centering
	\includegraphics[scale=0.55]{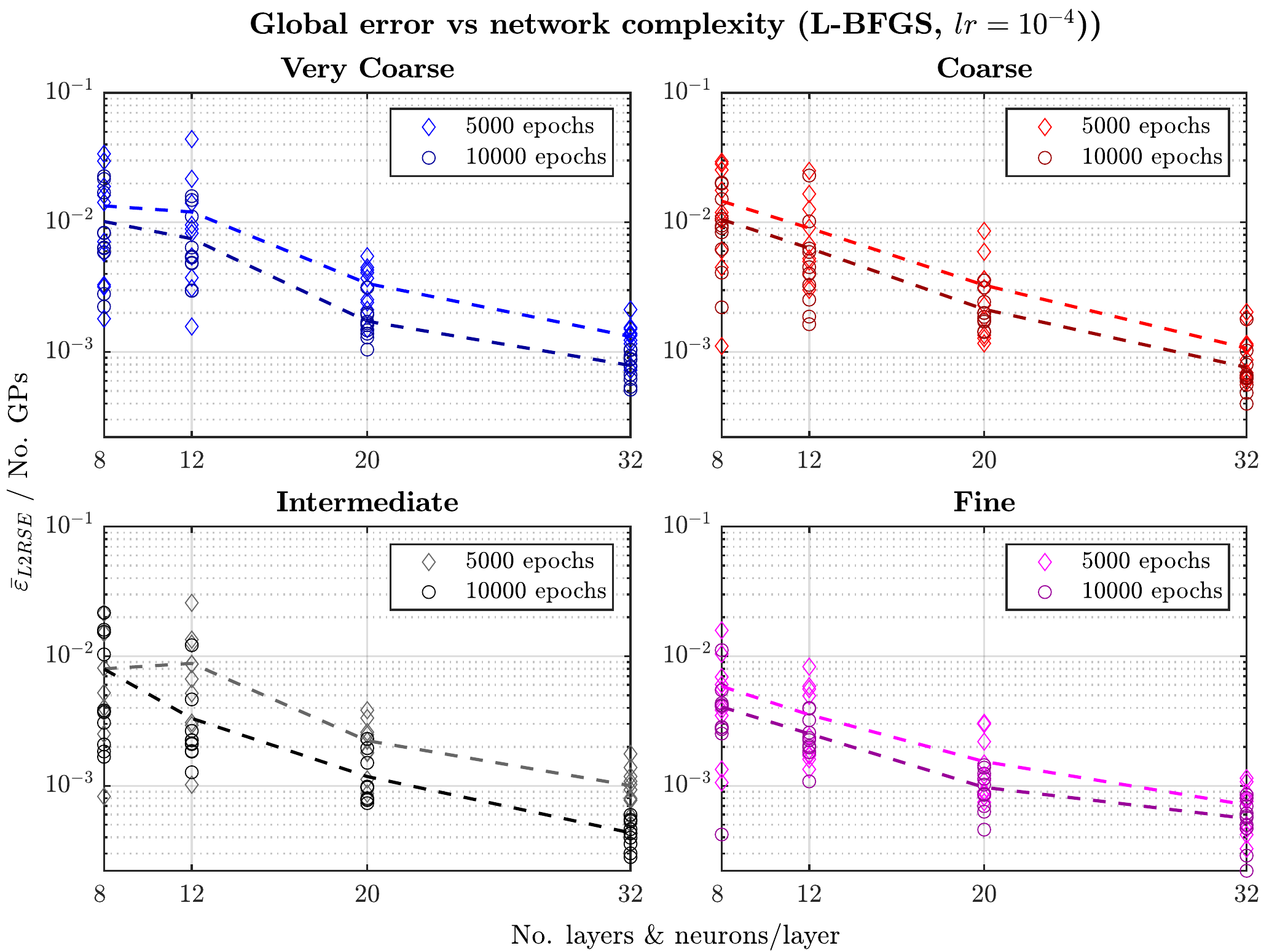}
	\caption{Global error ${\bar{\varepsilon}}_{L2RSE}$ convergence with increasing network complexity. The values are measured at the end of L-BFGS and for $lr = 10^{-4}$ across all mesh idealizations, and they are normalized over the number of Gauss points of each mesh. The markers correspond to 10 independent simulations, and the dashed lines connect their average. All networks are square.}
	\label{Figure_Convergence_L2RSEvsLayers_LBFGS_lowlr}
\end{figure}

\noindent {\bf{Compliance with Universal Approximation Theory (UAT):}} We begin with the error convergence analysis against the network complexity, using the single-notch specimen and exploring the search space described in Section \ref{Sec:SearchSpace}. We report the response of the training error $J$ and global error $\bar{\varepsilon}_{L2RSE}$ at the end of Adam in Figures \ref{Figure_Convergence_JvsLayers_Adam_lowlr} and \ref{Figure_Convergence_L2RSEvsLayers_Adam_lowlr} and at the end of L-BFGS in Figures \ref{Figure_Convergence_JvsLayers_LBFGS_lowlr} and \ref{Figure_Convergence_L2RSEvsLayers_LBFGS_lowlr} respectively. These graphs refer to the case of $lr = 10^{-4}$, and the values are normalized over the number of Gauss points. The markers correspond to the data from each independent simulation, and the dashed lines connect their average values in order to indicate their mean-value performance. Each of the four subplots in each figure corresponds to a different finite element mesh resolution, as described in Section \ref{Sec:Investigation_and_Validation_Models}. In all cases, we observe a clear sign of convergence with respect to the network size for both $J$ and $\bar{\varepsilon}_{L2RSE}$, which in turn indicates the conformance to convergence of the training error $\mathcal{E}_{T}$ and the global error $\mathcal{E}$ against the PINN complexity as laid out in the theoretical derivations in Section \ref{Sec:Convergence_analysis}. These converging trends remain consistent for any combination of epochs, network size and training dataset size. We then repeat the entire study for the case of $lr = 10^{-3}$. The results can be found in \ref{Appendix:Appendix_Dataset_Size}, and they further verify the observed conclusions. In view of the previous discussion, where we established the importance of demonstrating the sanity of any new method in terms of numerical convergence, this is a valuable finding that contributes to the rigor of the PINN reliability and the consequent I-FENN setup.

\begin{figure}[H]
	\centering
	\includegraphics[scale=0.55]{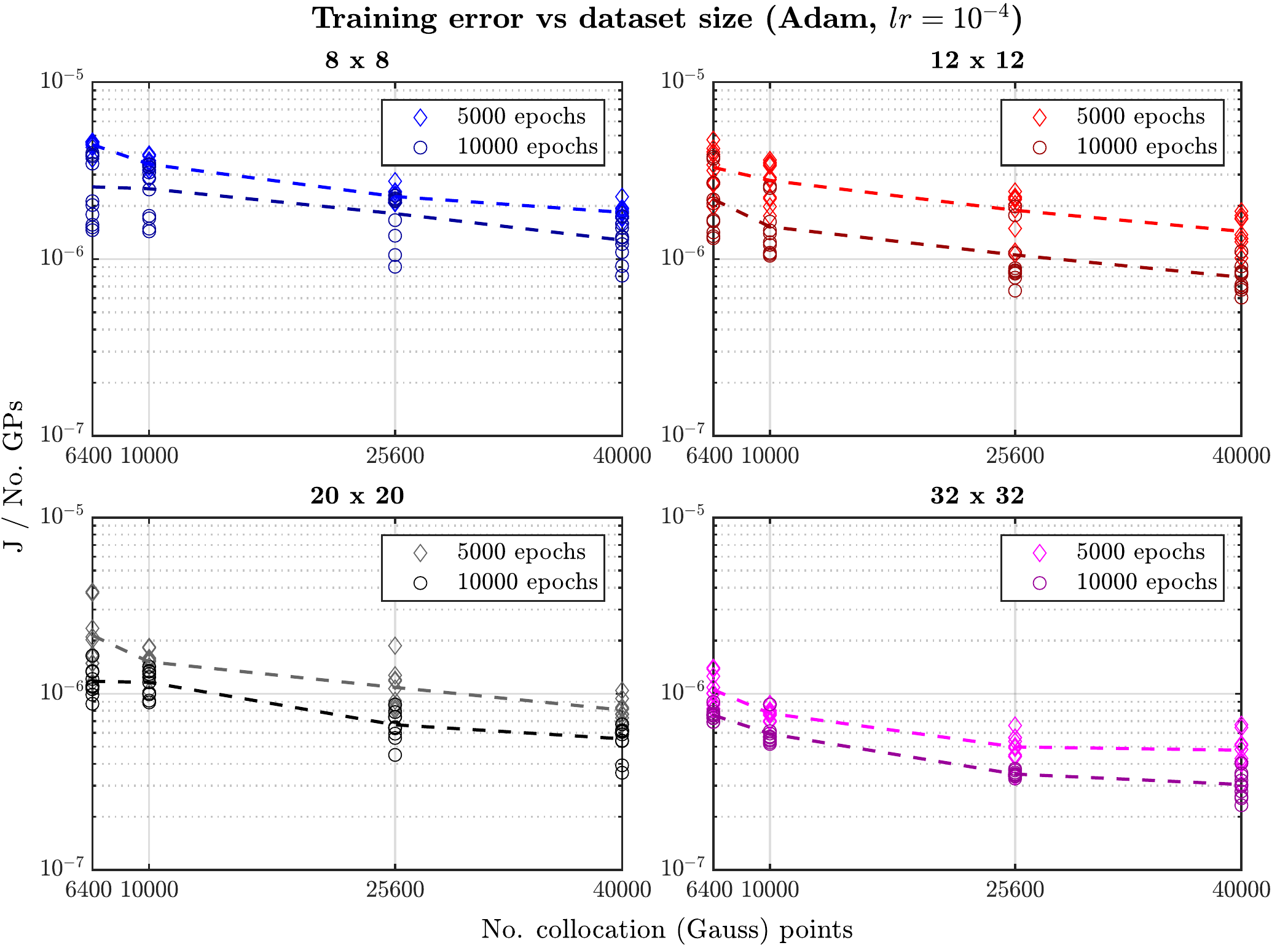}
	\caption{Training error $J$ convergence with increasing number of training samples. Each subplot corresponds to a square network of size [$L \times N$] = [$8 \times 8$, $12 \times 12$, $20 \times 20$, $32 \times 32$]. The values are measured at the end of Adam and for $lr = 10^{-4}$ across all mesh idealizations, normalized over the number of Gauss points of each mesh. The markers correspond to 10 independent simulations, and the dashed lines connect their average. }
	\label{Figure_Convergence_JvsGPs_Adam_lowlr}
\end{figure}

\begin{figure}[H]
	\centering
	\includegraphics[scale=0.55]{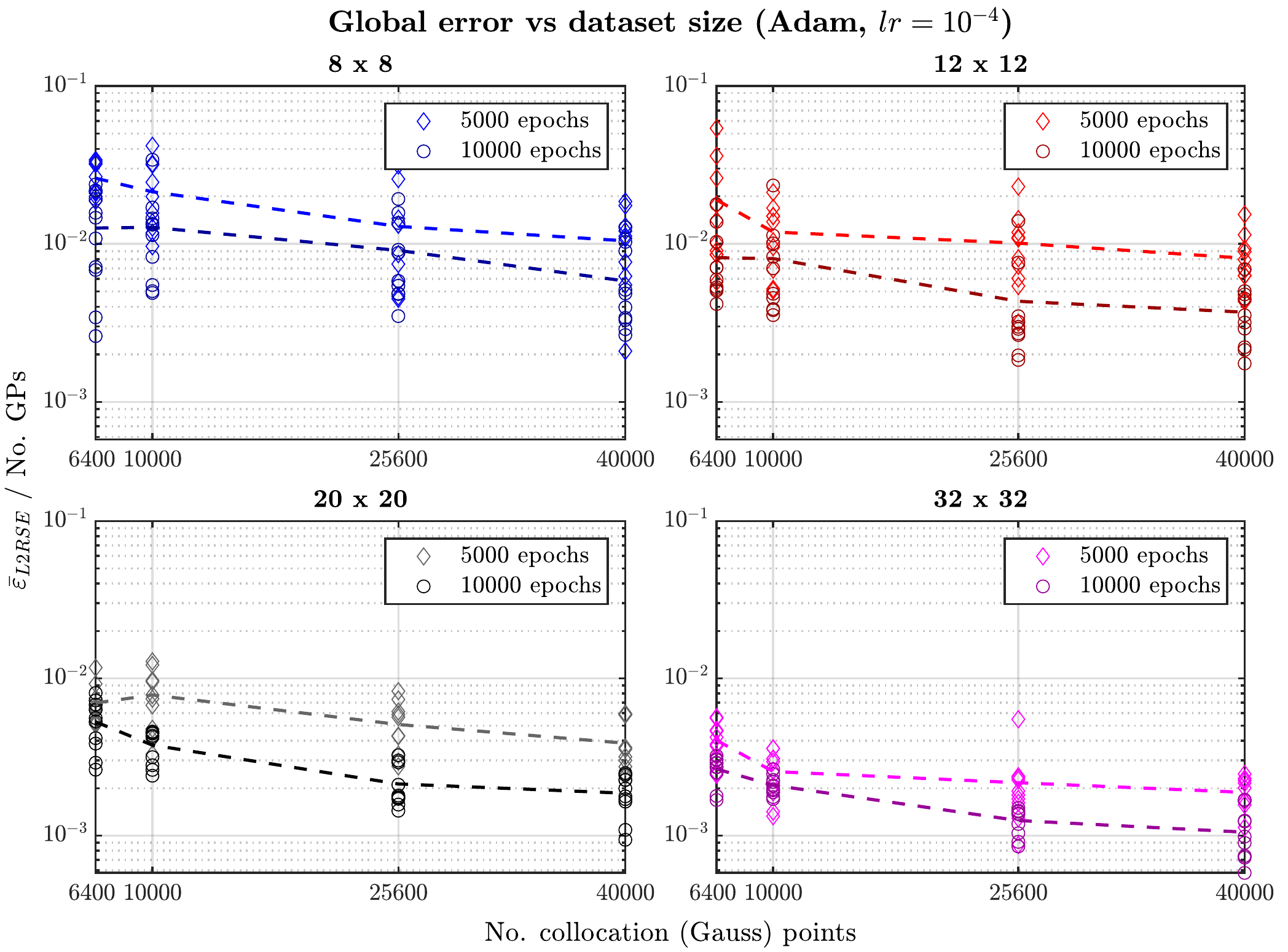}
	\caption{Global error ${\bar{\varepsilon}}_{L2RSE}$ convergence with increasing number of training samples. Each subplot corresponds to a square network of size [$L \times N$] = [$8 \times 8$, $12 \times 12$, $20 \times 20$, $32 \times 32$]. The values are measured at the end of Adam and for $lr = 10^{-4}$ across all mesh idealizations, normalized over the number of Gauss points of each mesh. The markers correspond to 10 independent simulations, and the dashed lines connect their average.}
	\label{Figure_Convergence_L2RSEvsGPs_Adam_lowlr}
\end{figure}

\begin{figure}[H]
	\centering
	\includegraphics[scale=0.55]{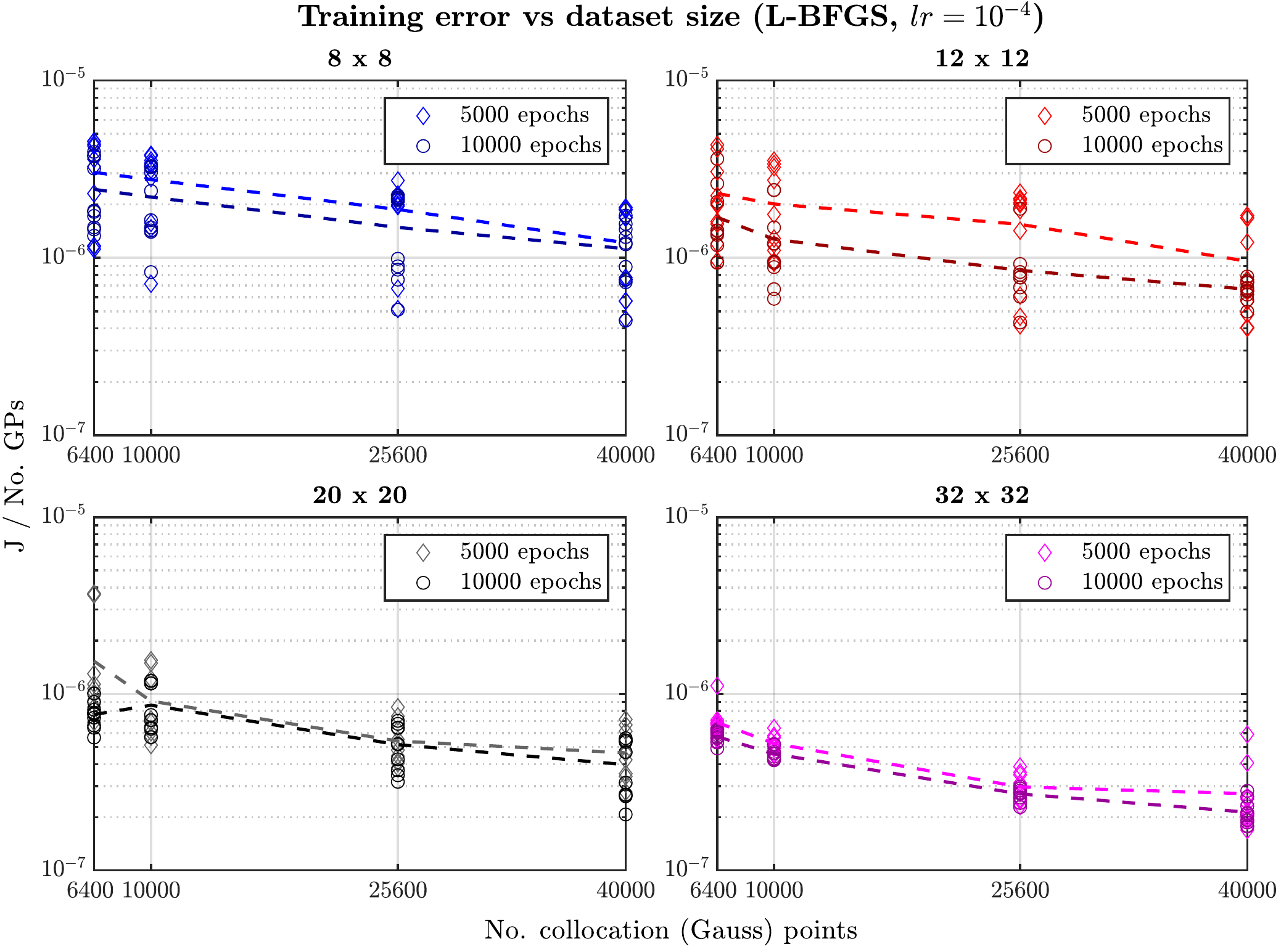}
	\caption{Training error $J$ convergence with increasing number of training samples. Each subplot corresponds to a square network of size [$L \times N$] = [$8 \times 8$, $12 \times 12$, $20 \times 20$, $32 \times 32$]. The values are measured at the end of L-BFGS and for $lr = 10^{-4}$ across all mesh idealizations, normalized over the number of Gauss points of each mesh. The markers correspond to 10 independent simulations, and the dashed lines connect their average.}
 \label{Figure_Convergence_JvsGPs_LBFGS_lowlr}
\end{figure}

\begin{figure}[H]
	\centering
	\includegraphics[scale=0.55]{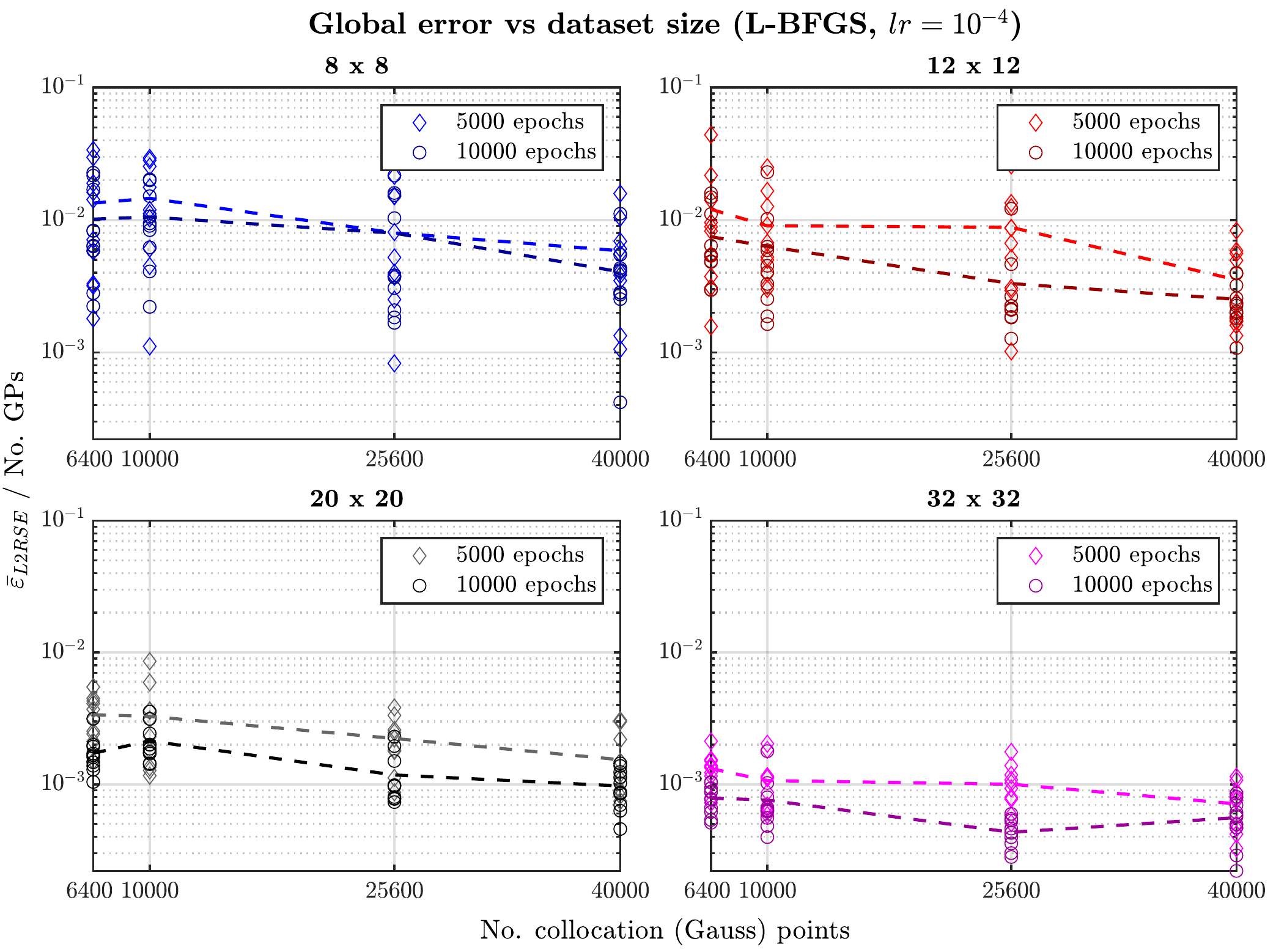}
	\caption{Global error ${\bar{\varepsilon}}_{L2RSE}$ convergence with increasing number of training samples. Each subplot corresponds to a square network of size [$L \times N$] = [$8 \times 8$, $12 \times 12$, $20 \times 20$, $32 \times 32$]. The values are measured at the end of L-BFGS and for $lr = 10^{-4}$ across all mesh idealizations, normalized over the number of Gauss points of each mesh. The markers correspond to 10 independent simulations, and the dashed lines connect their average.}
	\label{Figure_Convergence_L2RSEvsGPs_LBFGS_lowlr}
\end{figure}

\noindent {\bf{Compliance with training sample size:}} We now re-arrange the results and plot the training error $J$ and global error $\bar{\varepsilon}_{L2RSE}$ against the number of Gauss points. This serves the purpose of observing convergence signs in the training sample limit, and the results for $J$ and $\bar{\varepsilon}_{L2RSE}$ for $lr = 10^{-4}$ are shown in Figures \ref{Figure_Convergence_JvsGPs_LBFGS_lowlr} and  \ref{Figure_Convergence_L2RSEvsGPs_LBFGS_lowlr} respectively at the end of L-BFGS, and \ref{Figure_Convergence_JvsGPs_Adam_lowlr} and  \ref{Figure_Convergence_L2RSEvsGPs_Adam_lowlr} at the end of Adam. Each subplot in these figures corresponds to a different network size, and similar to the previous plots, the values of $J$ and $\bar{\varepsilon}_{L2RSE}$ are normalized over the number of collocation points. We observe a consistent drop of both metrics of accuracy at the training dataset limit. This behavior remains consistent for any network size and both number of epochs. This monotonically decreasing nature is another sign of convergence of our PINN predictions. 

Overall, Figures \ref{Figure_Convergence_JvsLayers_LBFGS_lowlr}-\ref{Figure_Convergence_L2RSEvsGPs_Adam_lowlr} showcase the sanity of the PINN scheme in the non-local gradient strain equation, demonstrating a consistent convergence response of both the training and global error against both the network complexity and the training dataset. The results of this extensive investigation answer affirmatively to the first research objective as it was defined in the introduction: the series of PINN minimizers converge in terms of the training and the global error against both the PINN size and the training dataset size. The answer is the same regardless of the training optimization mechanism, number of epochs and learning rate we considered. They also show that a progressively lower training error will lead to better approximation capabilities, which is a similar-in-spirit observation with the interpretations of the theoretical results in \cite{mishra2022estimates}.

\subsection{Optimal strategy: Larger network vs Larger dataset}
\begin{figure}[H]
	\centering
	\includegraphics[width=\linewidth]{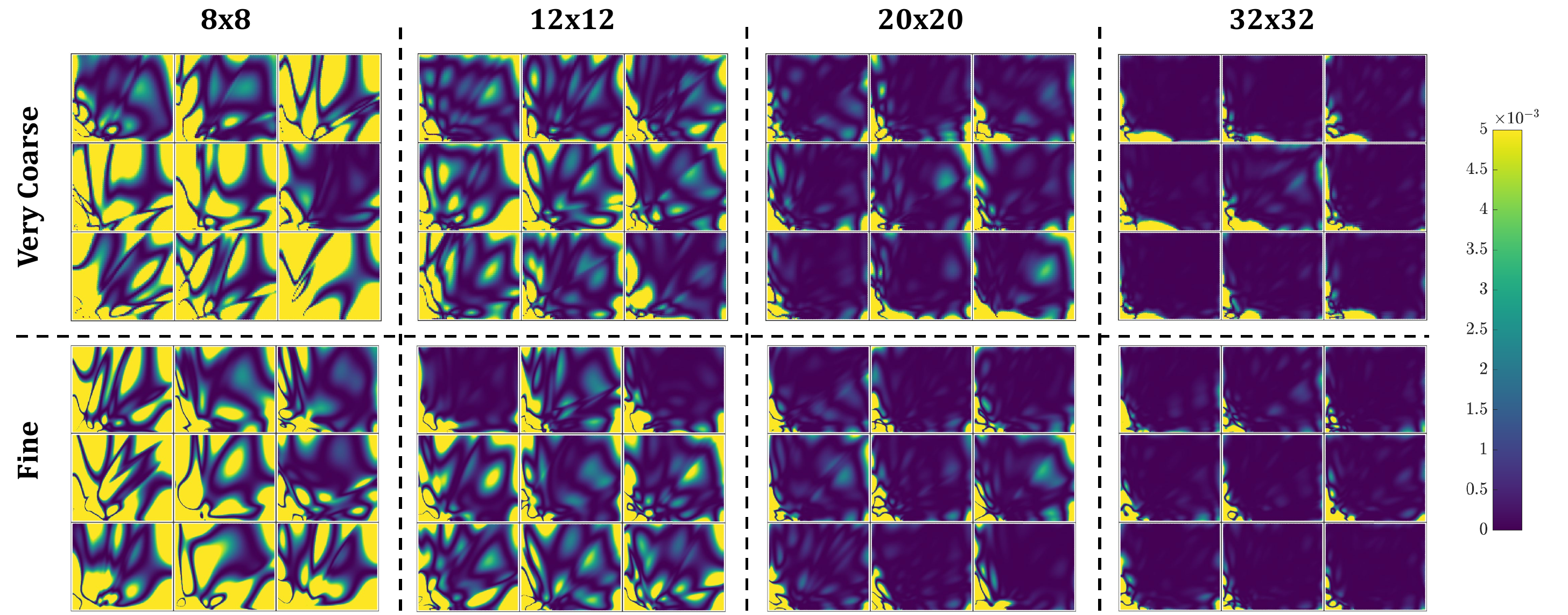}
	\caption{Relative squared error maps of the non-local strain predictions for the Very Coarse (top row) and Fine (bottom row) models across all complexity values. For each case we report the results from nine independent simulations.}
	\label{Figure_comparison_preds}
\end{figure}

The results in Section \ref{Sec:Error_convergence_analysis} clearly demonstrate that increasing the size of either the network or the dataset improves the overall PINN performance. The natural next question is: \textit{which one of the two yields the greatest improvement in the algorithm performance}? In other words, if the number of training epochs is the same, is it preferable to train higher complexity networks in smaller data sizes or lower complexity networks on bigger datasets? Figure \ref{Figure_comparison_preds} shows the relative squared error of the non-local strain profile for all shapes at $lr = 10^{-3}$ and $ep = 10000$, for the Very Coarse (top row) and Fine (bottom row) models. It is apparent that the algorithm performance is mostly dominated by the PINN size. The increased complexity allows for much better approximation compared to using smaller networks for a refined the mesh grid.  The same conclusion can be drawn by analysing the trends of $J$ and ${\bar{\varepsilon}}_{L2RSE}$ in Figures \ref{Figure_Convergence_JvsLayers_LBFGS_lowlr}, \ref{Figure_Convergence_L2RSEvsLayers_LBFGS_lowlr}, \ref{Figure_Convergence_JvsGPs_LBFGS_lowlr} and \ref{Figure_Convergence_L2RSEvsGPs_LBFGS_lowlr}.

\begin{figure}[H]
	\centering
	\includegraphics[scale=0.75]{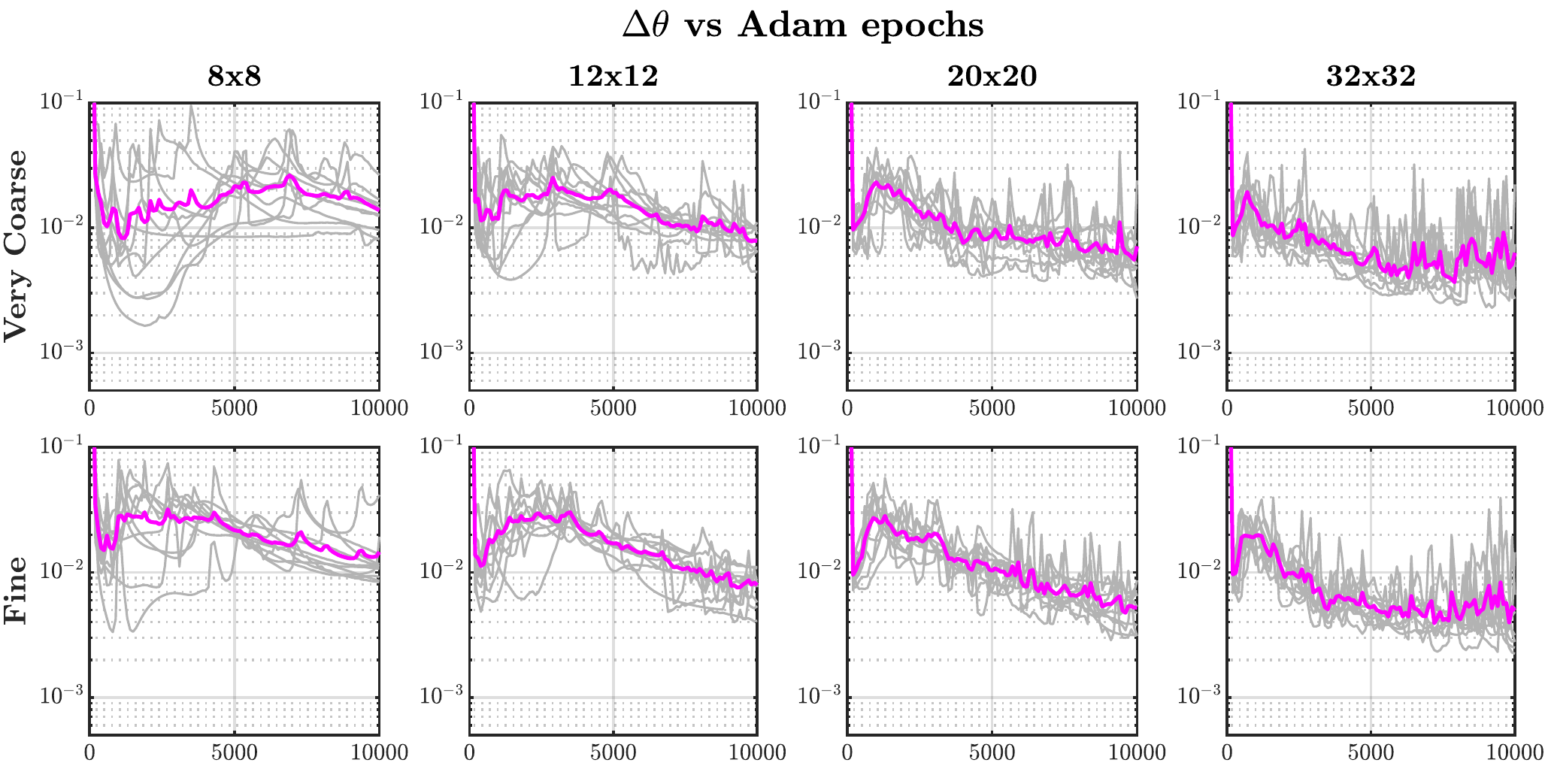}
	\caption{Plots of the parameters relative change $\Delta \theta$ against the Adam epochs, for the Very Coarse model (top row) and Fine model (bottom row) and all network sizes. We observe a tighter grouping of the curves $\Delta \theta$ as the network's size increases at both mesh levels. This demonstrates that by avoiding areas of prematurely low $\Delta \theta$, a larger network eliminates the impact of initialization and helps the optimizer performance. Its impact is more significant than increasing the dataset size.}
	\label{Figure_Comp3_Dtheta}
\end{figure}

We now examine, in more detail, the training dynamics of the optimization process through the lens of the relative parameter change $\Delta \theta$ as presented in Figure \ref{Figure_Comp3_Dtheta}, where we report the $\Delta \theta$ curves for the Very Coarse and Fine model for all network sizes. The curves for all independent simulations are shown in gray color, and their average value - which is utilized to characterize more compactly the overall trends - is shown in magenta. For both mesh models we observe similar patterns, which gives further evidence to our previous conclusion. First, in the low complexity zone $8 \times 8$, we observe a stark difference between the $\Delta \theta$ paths of each independent simulation. This signifies a strong dependency of the low complexity networks on the parameter initialization scheme, and as a result the optimizer path is drastically different from case to case. This observation holds true for both mesh models, and particularly for the initial 5000 iterations. Also, we note that prematurely low values of $\Delta \theta$ - in the order of $10^{-3}$ - indicate signs of slow and inefficient training. This is in contrast to signs of healthier training for larger networks, where the $\Delta \theta$ curves follow a steadier downward path until they start to exhibit oscillations, a sign that the optimizer approaches a local minimum. These observations are consistently more apparent as the complexity increases, and they are supported more evidently by inspecting the average-value magenta curves. 

\begin{figure}[H]
	\centering
	\includegraphics[scale=0.65]{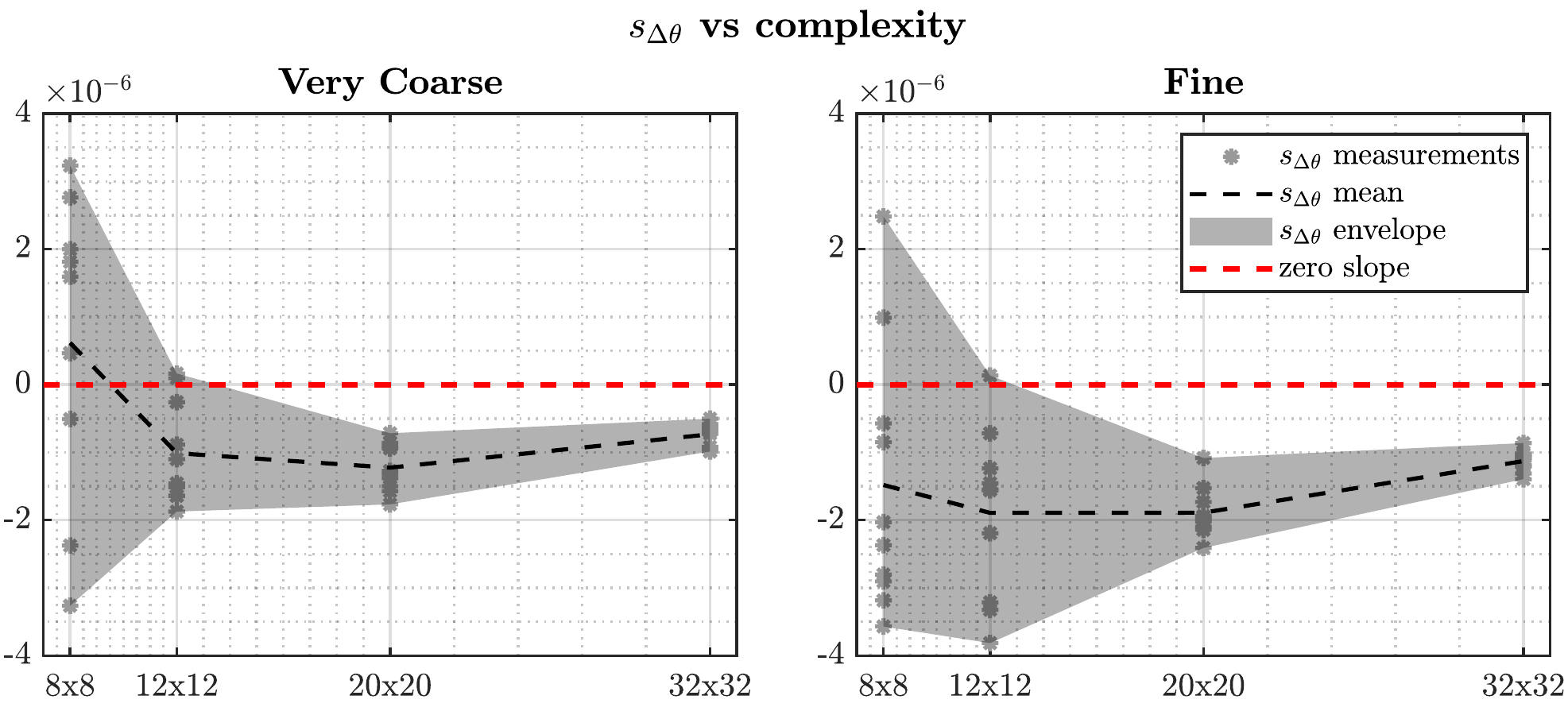}
	\caption{We obtain the best fit lines of the $\Delta \theta$ curves and extract their slopes, denoted as $s_{\Delta \theta}$. Here we plot $s_{\Delta \theta}$ against the  network complexity for the Very Coarse model (left) and Fine model (right). Each marker corresponds to an independent measurement. The black dashed line is the average of the 10 measurements, the gray zone indicates their envelope, and the red dashed line signifies the zero slope of an ideal horizontal line. At both mesh levels we observe that a significant deviation in the $\Delta \theta$ slopes at low complexities, which becomes more bounded with increasing network size. At the largest size, $s_{\Delta \theta}$ tends towards the zero slope line, indicating that at that stage the training reaches convergence.}
	\label{Figure_Comp4_SDtheta}
\end{figure}

An alternative way to monitor the $\Delta \theta$ paths is by extracting the slopes of their best fit lines, $s_{\Delta \theta}$, as described in Section \ref{Performance_metrics}. We plot the $s_{\Delta \theta}$ in Figure \ref{Figure_Comp4_SDtheta} for the Very Coarse (left) and Fine models (right). Observation of these subplots verifies our previous conclusions. For low complexity PINNs, $s_{\Delta \theta}$ obtains both positive and negative values, with the positive ones being mainly affected by the prematurely low $\Delta \theta$. As the complexity increases and $\Delta \theta$ follows a more stable descending path, the corresponding $s_{\Delta \theta}$ are consistently grouped more tightly together and obtain negative values. Finally, at the highest complexity of 32x32, the $\Delta \theta$ oscillation signs tend to drive the $s_{\Delta \theta}$ values closer to the zero threshold, another sign that the PINN has reached a local minimum and the training process has converged. These observations are the same regardless of the chosen mesh refinement. Consequently, we can conclude that training a higher-complexity network with a smaller data size yields higher learning efficiency than selecting a smaller network with more data points.

\subsection{Improving I-FENN performance}

Having established these trends, one can reasonably ask: Can we train a high-complexity PINN on a small dataset, and use it for predictions in larger samples? Is a PINN which is well-trained on a coarser, yet representative, grid capable of approximating the field of interest when the latter is evaluated on a denser grid? The computational advantage in this case is twofold: a) faster PINN training due to the reduced data-size input, and more importantly b) faster execution of the FEM component of I-FENN due to the much smaller Jacobian matrix, thus enabling I-FENN to be used across finer mesh discretizations. The inverse question though is also instructive to examine: can we train a PINN on a fine mesh and use it to coarser approximations? 

\begin{figure}[H]
	\centering
	\includegraphics[width=\linewidth]{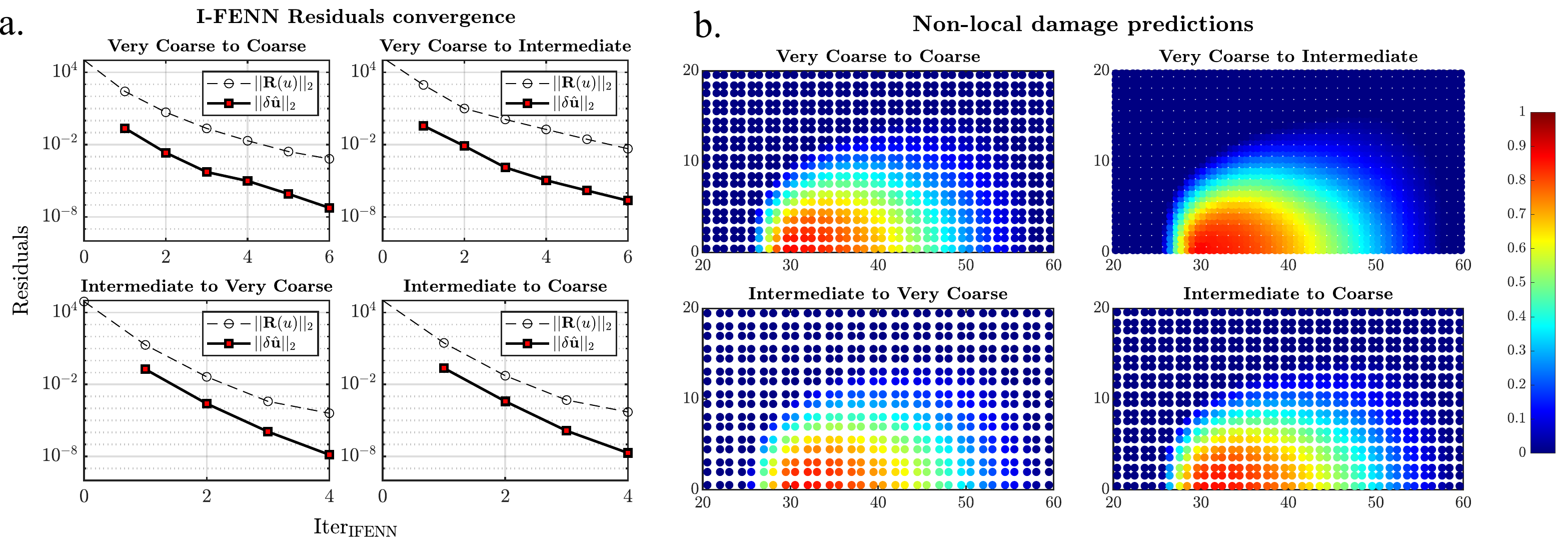}
	\caption{I-FENN implementation with PINNs trained and tested at different mesh densities. Four cases are analyzed: a PINN trained on the Very Coarse mesh is tested against the Coarse and the Intermediate, and an Intermediate-mesh trained PINN is tested against the Very Coarse and Coarse. {\bf{a}}: Convergence of the internal stresses residual (dashed) and  displacement vector residual (solid). {\bf{b}}: Predicted non-local damage profiles.}
	\label{Figure_IFENN_ConvergenceComparison}
\end{figure}

To answer both, we perform two numerical experiments. First, we select a $[20,20,10000,10^{-3}]$ PINN which is trained on the Very Coarse mesh and use it within I-FENN to make predictions of the non-local strain and damage profile in the Coarse and the Intermediate mesh. Then, a $[32,32,10000,10^{-3}]$ PINN which is trained on the Intermediate mesh is examined against the Very Coarse and Coarse idealizations. The results of this study are shown in Figure \ref{Figure_IFENN_ConvergenceComparison}. In that case, Figure \ref{Figure_IFENN_ConvergenceComparison}a demonstrates the residuals convergence and \ref{Figure_IFENN_ConvergenceComparison}b shows the non-local damage profile for all cases on a Gauss-point basis. These graphs clearly indicate that the answer to both questions is affirmative. The computational benefit is particularly evident in the top-right case, where the PINN is trained on the Very Coarse (6400 GPs) and tested against the Intermediate (25600 GPs) is noteworthy. For reference, the average training of a $[20,20,10000,10^{-3}]$ PINN is $t_{train} \approx 10min$ on the Very Coarse and $t_{train} \approx 20min$ on the Intermediate mesh. Additionally, the execution time of the load increment of interest for the Intermediate mesh with I-FENN and the reduced Jacobian is $t \approx 2min$, versus $t \approx 6min$ for the previous load increment using conventional FEM and the original Jacobian. Therefore, these results clearly reveal the role of understanding convergence and error minimization trends towards enhancing the computational performance of the I-FENN setup.

\subsection{Concluding the PINN error convergence investigation}
\label{Sec:Concluding_PINN_errorconvergence}

Here we provide a concise summary of our observations and conclusions from our investigation on the PINN error convergence:

\begin{itemize}

    \item We show that both of the training error $\mathcal{E}_{T}$ (represented by $J$) and the global error $\mathcal{E}$ (represented by ${\bar{\varepsilon}}_{L2RSE}$) are minimized as the network complexity and training dataset increase. This proves empirically the conformance of our PINN setup with the convergence theories summarized in Section \ref{Sec:Convergence_analysis}.
    
    \item We show that using a higher complexity PINN yields greater improvement on the network prediction accuracy than using a larger training dataset.
    
    \item We show that the conclusions above can lead to a more enhanced I-FENN computational performance by training on coarser (smaller) datasets and making predictions on finer (larger) models.
    
\end{itemize}

\section{PINN engineering-guided hyperparameter search}
\label{Sec:PINN_Engineering_Guided_HPS}


\subsection{Optimal aspect ratio: accuracy vs. cost}
\begin{figure}[H]
	\centering
	\includegraphics[width=\linewidth]{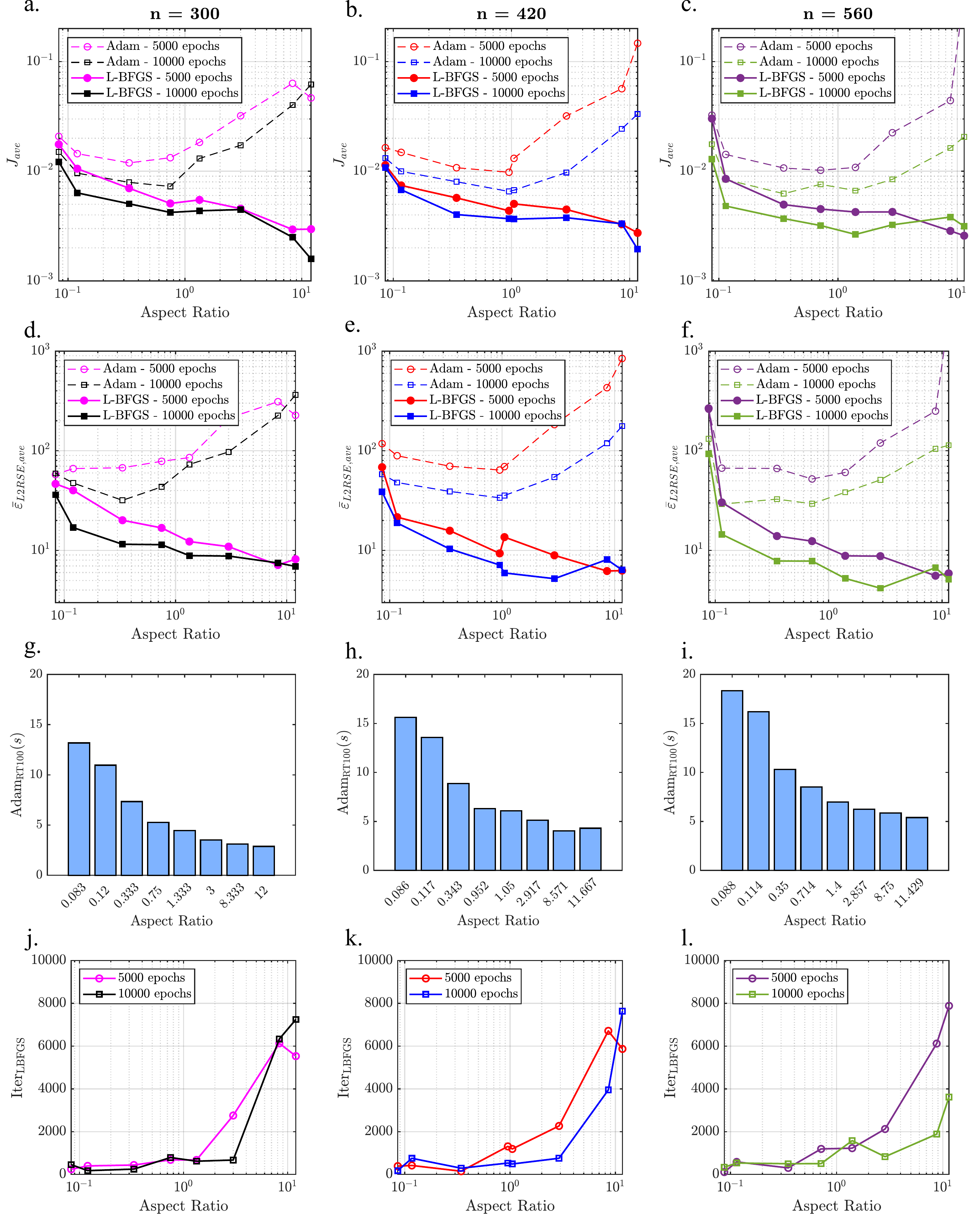}
	\caption{Impact of network shape on the PINN predictive accuracy and computational effort for the high learning rate ($lr = 10^{-3}$). Three cases with a fixed total number of neurons are analyzed, and the depth and width are varied accordingly. The cases correspond to 300 (left column), 420 (middle column) and 560 (right column) neurons in total. {\bf{a-c}}: Training error $J$ and {\bf{d-f}}: L2RSE of the predicted non-local strain, at the end of Adam (dashed) and L-BFGS (solid) algorithm for 5000 and 10000 epochs. {\bf{g-i}}: Average simulation time (sec) of 100 Adam epochs and {\bf{j-l}}: number of L-BFGS iterations for each case. Each data point in plots {\bf{a-f,j-l}} corresponds to the average value of 10 independent training simulations.}
	\label{Figure_StrC100_ARstudy_highlr}
\end{figure}

\begin{figure}[H]
	\centering
	\includegraphics[width=\linewidth]{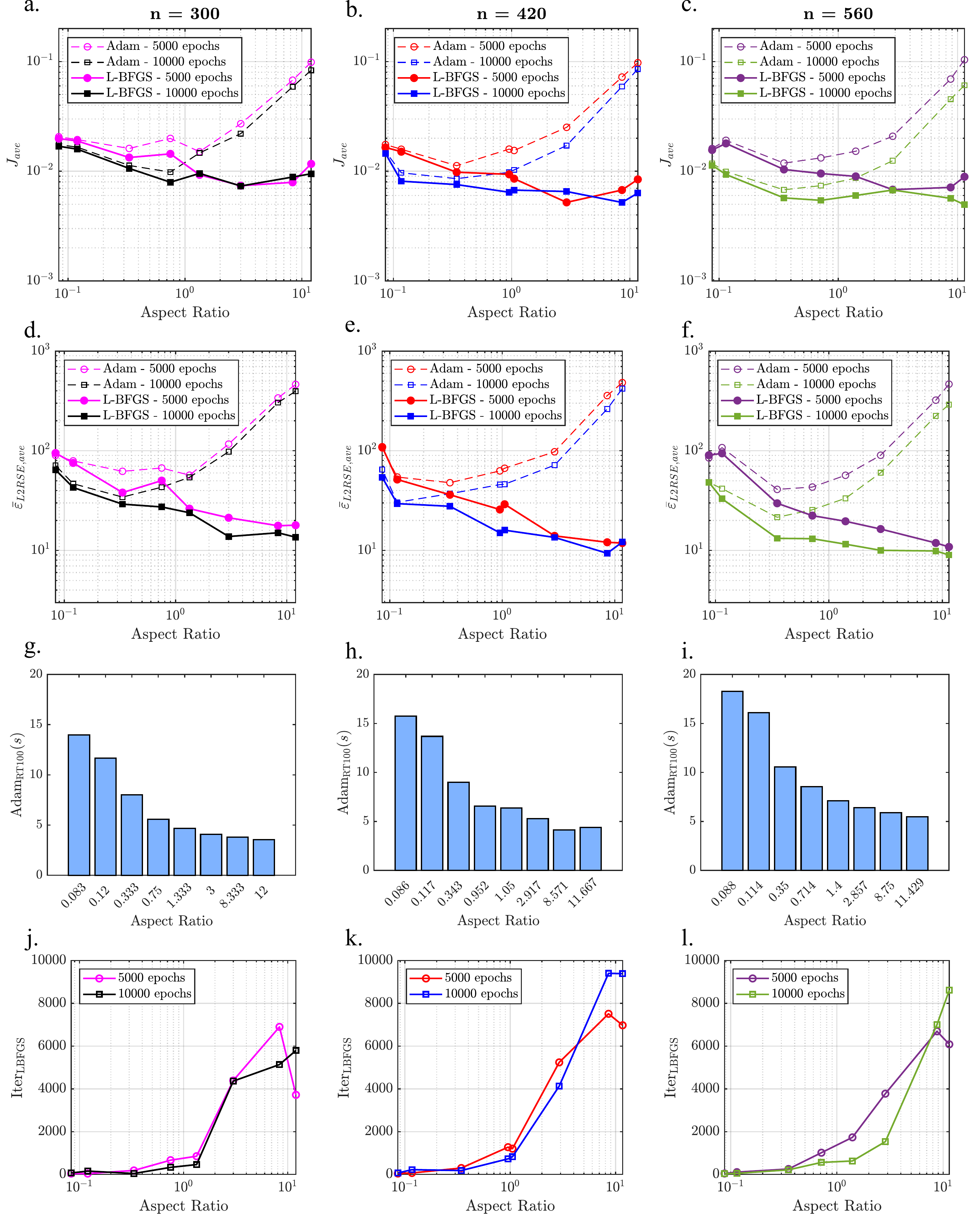}
	\caption{Impact of network shape on the PINN predictive accuracy and computational effort for the low learning rate ($lr = 10^{-4}$). {\bf{a-c}}: Training error $J$ and {\bf{d-f}}: L2RSE of the non-local strain, at the end of Adam (dashed) and L-BFGS (solid) algorithm for 5000 and 10000 epochs. {\bf{g-i}}: Average simulation time (sec) of 100 Adam epochs and {\bf{j-l}}: number of L-BFGS iterations for each case. Each data point in plots {\bf{a-f,j-l}} corresponds to the average value of 10 independent training simulations.}
	\label{Figure_StrC100_ARstudy_lowlr}
\end{figure}

In this section we present the results of our numerical investigation on the impact of the PINN architecture and other hyperparameters on the network predictive accuracy, computational effort, and chances of arriving at trivial solutions. Figures \ref{Figure_StrC100_ARstudy_highlr} and \ref{Figure_StrC100_ARstudy_lowlr} show the results for $lr = 10^{-3}$ and $lr = 10^{-4}$, respectively. Both figures follow an identical layout. Each column of graphs corresponds to a different $n$ case: 300, 420 and 560 neurons. The first row of graphs shows the average training cost values, and the second row depicts the average values of the $\bar{\varepsilon}_{L2RSE}$. The third row reports $\mathrm{Adam_{RT100}}$ which is the average simulation time of 100 Adam epochs, and the fourth row shows $\mathrm{Iter_{LBFGS}}$ which is the average number of the L-BFGS iterations until convergence. The observation of Figures \ref{Figure_StrC100_ARstudy_highlr} and \ref{Figure_StrC100_ARstudy_lowlr} leads to the empirical establishment of several patterns with regards to the predictive accuracy and computational efficiency of the training process, which are discussed below:

\begin{enumerate}

\item The average training error $J$ follows a loosely described $U-shape$ trend at the end of the Adam training stage. The minimum values are obtained when the network aspect ratio is in the range $0.1 \leq AR \leq 1$, and it follows a sharp increase beyond these limits. This behavior is observed for any combination of neurons, epochs and learning rate. This shows a strong and consistent correlation between the $AR$ of a fully-connected PINN and the training error trend at the end of Adam stage, in view of the number of neurons constraint. 

\item The L-BFGS stage has a minimal influence on the average training error $J$ and the global error $\bar{\varepsilon}_{L2RSE}$ when the network is too deep-and-narrow ($AR \leq 0.1$), but its impact becomes more pronounced as the network becomes shallower-and-wider. This is a clear indication that these networks are not the best candidates for the non-local gradient PDE, as they perform distinctively worse than their shallow-and-wide counterparts.

\item As the network aspect ratio decreases below unity, the Adam training time increases significantly. For example, for the case of 300 neurons and $lr = 10^{-3}$, training a deep-and-narrow network with $AR = 0.083$ is 4.5 times slower than training a wide network with $AR = 12$, and the observed trend holds true for any total number of neurons. This is clearly indicating the computational expense of training very deep-and-narrow networks, and suggests avoiding aspect ratios which are significantly lower than 1. 

\item As the aspect ratio becomes larger than unity, the number of L-BFGS iterations $Iter_{LBFGS}$ increases drastically compared to the cases where $AR \leq 1$. This holds true for any number of total neurons and regardless the number of epochs or learning rate value. In view of the first two comments, it is reasonable to expect a greater effort from L-BFGS in the high aspect ratio regime. The magnitude in the difference though is quite remarkable; see for example the case of $[300, 5000, 10^{-3}]$, where it takes more than 6000 L-BFGS iterations for $AR = 8.33$, versus less than 700 for $AR = 0.75$ which is an order of magnitude difference. This is a clear sign that as the network width increases with respect to its depth, utilization of this algorithm becomes significantly more expensive from a computational standpoint.

\item We finally note that as the number of total neurons increases, the $\bar{\varepsilon}_{L2RSE}$ at the end of the entire training process gets smaller. This trend holds true for all combinations of epochs and learning rate, and generally holds across the entire $AR$ range. This further confirms that the global error is decreased as the PINN complexity increases, which is consistent with the results of the previous section.


\end{enumerate}

\subsection{Trivial solutions: Role of aspect ratio}
\label{Sec:Trivial_solutions}

\begin{figure}[H]
	\centering
	\includegraphics[width=\linewidth]{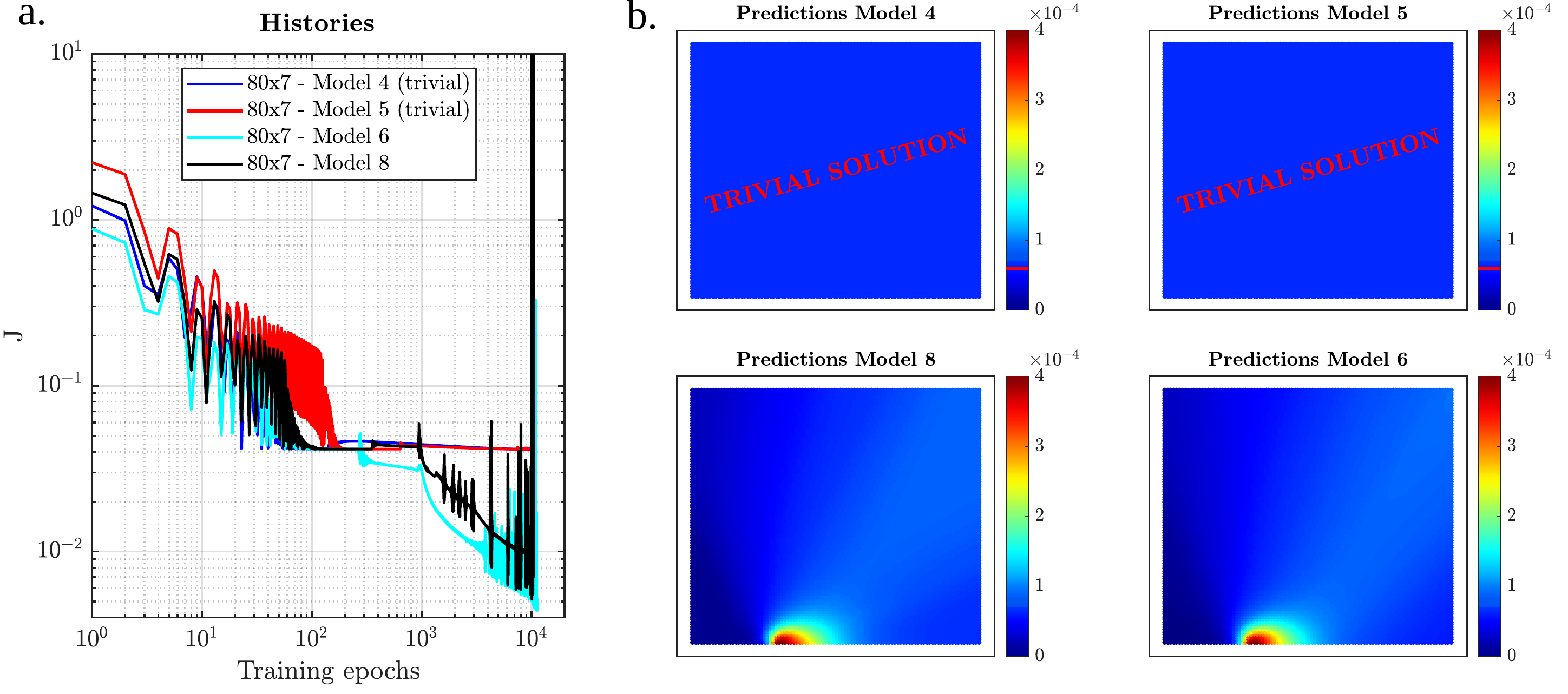}
	\caption{\textbf{a}: Training error evolution and \textbf{b}: non-local strain predicted profiles for four idealizations of the $[80,7,10000, 10^{-3}]$ case. Models 6 and 8 are trained successfully, but training of Models 4 and 5 gets stuck during the Adam stage and the learning process is stalled. Consequently, the partial derivatives of the non-local strain with respect to the input variables vanish and the network predicts the same value for the non-local strain at all Gauss points. This value corresponds to the average value of the input local strain and it is marked in the colorbars with horizontal red lines. This figure indicates a major disadvantage of training very deep-and-narrow PINNs.}
	\label{Figure_Trivial_Solutions}
\end{figure}

Training very deep-and-narrow networks presents one additional challenge: these networks are prone to the well-known issue of vanishing gradients \cite{hochreiter2001gradient, hanin2018neural}. This phenomenon may arise during the gradient-descent training stage, and it is more pronounced when activation functions such as the sigmoid or the hyperbolic tangent are used, as is the case herein. In some cases, the depth of network leads to an adverse effect on the learning which is observed when the derivatives of the cost function with respect to its weights can become vanishingly small. Accordingly, the back-propagation algorithm fails to update the weights and the network training is stalled. In these cases, the partial derivatives of the output variable with respect to the inputs vanish to zero, and this results in the Laplacian term in Equation \ref{nonlocalGradientPDE} vanishing to zero as well. Consequently, and in order to minimize the cost function in a full-batch mode, the network maps the non-local strain at all collocation points to the average value of the input local strain. This is clearly shown in Figure \ref{Figure_Trivial_Solutions}, which investigates the $n = 560$ case with $ep = 10000$ and $lr = 10^{-3}$. Figure \ref{Figure_Trivial_Solutions}a shows the training error history for 4 out of the 10 idealizations, Models 4, 5, 6 and 8. The latter two are trained successfully, and the bottom row of graphs in Figure \ref{Figure_Trivial_Solutions}b shows the predicted non-local strain profile. Training of Models 4 and 5 however gets stuck at a trivial solution during the Adam stage, and the network is unable to be trained beyond that point. As a result, the PINN predicts a constant value at all points, which corresponds to the average value of the input local strain ($\varepsilon_{eq} = 6.6978 \times 10^{-4}$), see the top row of graphs in Figure \ref{Figure_Trivial_Solutions}b.

\subsection{Analysis of network parameters relative change \texorpdfstring{$\Delta \theta$}{}} 

\begin{figure}[H]
	\centering
	\includegraphics[width=\linewidth]{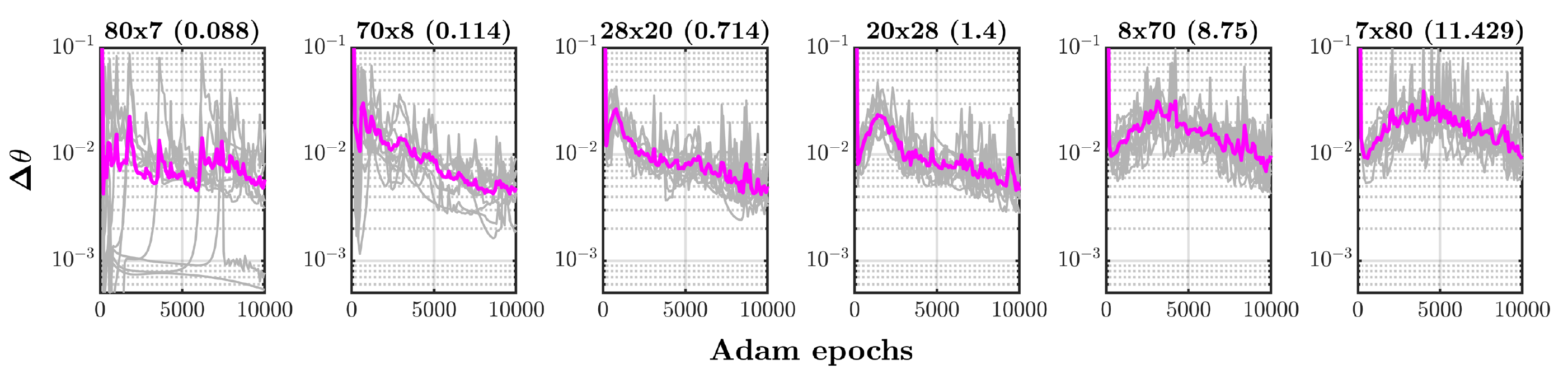}
	\caption{Relative change of the parameter vector during the Adam training for all shapes of $n = 560$, $lr = 10^{-3}$, $ep = 10000$. The titles reflect the network dimensions and the aspect ratio value (inside the parentheses). The gray curves correspond to the different idealizations of each shape, and the magenta curve is their average.}
	\label{Figure_ARstudy_comp5}
\end{figure}

Next, we monitor how the trainable weights change throughout the optimization process with the goal to uncover trends throughout the $AR$ range. We plot in Figure \ref{Figure_ARstudy_comp5} the $\Delta \theta$ curves for all shapes of $n = 560$, $lr = 10^{-3}$, $ep = 10000$. The gray curves correspond to the different idealizations of each shape, and the magenta curve is their average. At the low value range of $AR$, and particular for $70 \times 8$ ($AR = 0.11$) and $80 \times 7$ ($AR = 0.088$), we observe a distinct scatter between the curves. Particularly for the $80 \times 7$ shape we observe prematurely low values of $\Delta \theta$, which stay almost unchanged for several hundreds of epochs. This phenomenon is apparent in  several idealizations, and it is a clear sign of networks that get stuck during training. This further verifies our previous conclusions,in Section \ref{Sec:Trivial_solutions}, on the inability of very deep-and-narrow networks to train successfully. As the aspect ratio is increasing we observe signs of healthier training. In particular, the optimizer takes faster steps at the beginning of the training, and this is more clearly shown by the initial upward path in the magenta curves. The duration of this upward direction (progressively higher $\Delta \theta$) increases as the network becomes shallower and wider, and it is more evident for $8 \times 70$ ($AR = 8.75$) and $7 \times 80$ ($AR = 11.429$)

\subsection{I-FENN performance: maximum strain and number of iterations}

It is also crucial to touch upon another unique characteristic of the investigated problem: the importance of estimating correctly the $maximum$ value of the non-local strain: $\bar{\varepsilon}_{max}$. This is a peculiarity of the specific problem's nature and it is directly linked to the convergence of the iterative non-linear solution within the I-FENN setup. The value of $\bar{\varepsilon}_{max}$ dictates the corresponding maximum value of damage, which in turn impacts the smoothness of the FEM solution. We note that the nature of damage is irreversible and highly localized, and therefore crack propagation is largely dictated by the value of damage just in the fracture process zone \cite{hu1992fracture}. Therefore, there is an imperative need for accurately assessing the high frequencies in the non-local strain field. This is not to undermine the significance of approximating the non-local strain profile across the entire domain, since new cracks can be initiated and need to be captured in other areas of the domain, but to emphasize the importance of correctly capturing  $\bar{\varepsilon}_{max}$. 

\begin{figure}[H]
	\centering
	\includegraphics[scale=0.6]{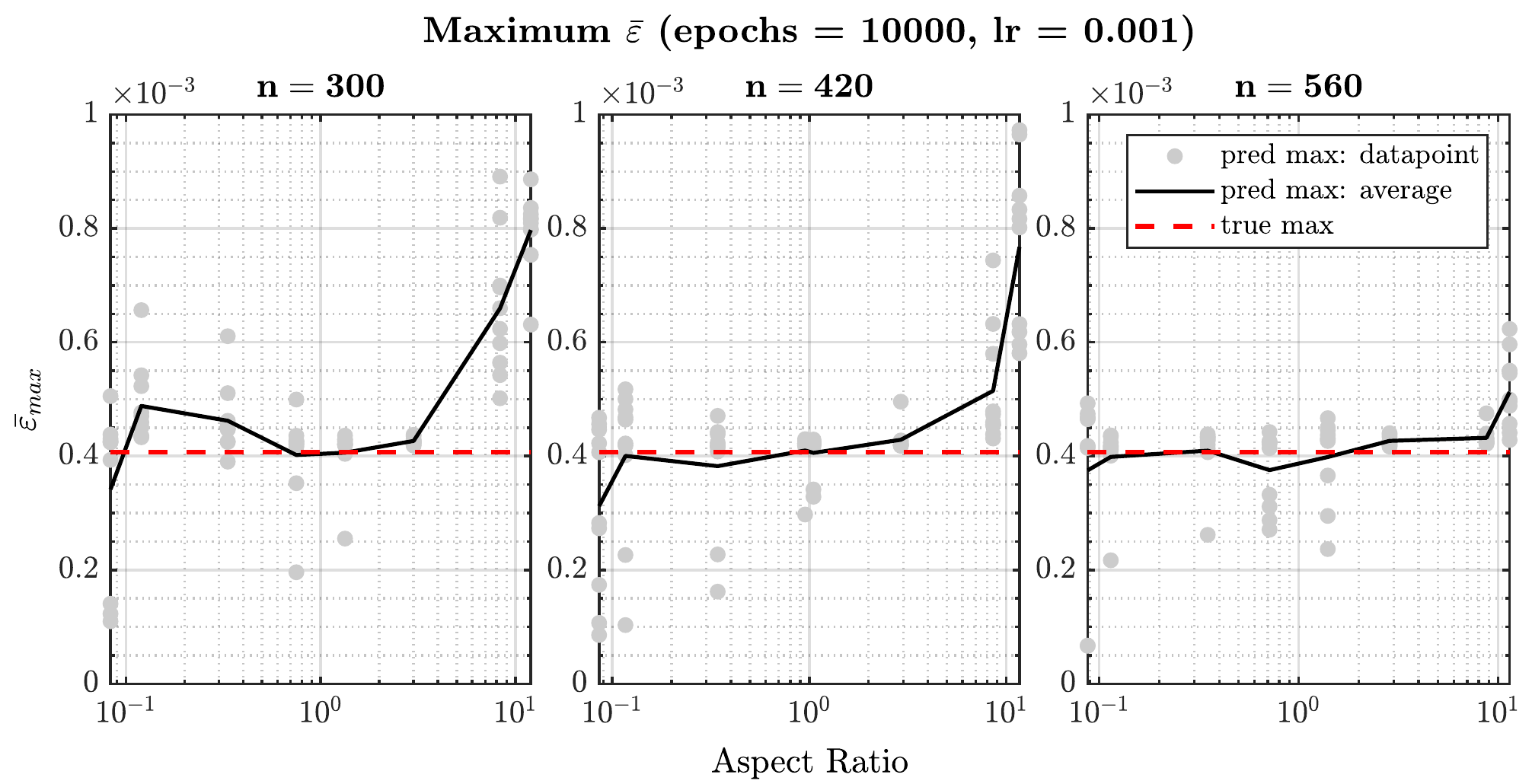}
	\caption{Maximum values of the $\bar{\varepsilon}$, for all neuron cases at $ep = 10000$ and $lr = 10^{-3}$. The grey circled markers are the predictions of $\bar{\varepsilon}_{max}$ for each independent simulation, and the solid black line connects their average values. The dashed red line indicates the true value of $\bar{\varepsilon}_{max}$. Two trends are apparent: a) as the network size decreases, shallow-and-wide networks tend to overestimate $\bar{\varepsilon}$, and b) as the network size increases the average prediction from the entire AR range approaches the true value.}
	\label{Figure_MaxStrains_10000_highlr}
\end{figure}

With this in mind, we plot in Figure \ref{Figure_MaxStrains_10000_highlr} the predicted values of $\bar{\varepsilon}_{max}$ for all $n$ cases with 10000 epochs and $lr = 10^{-3}$. The black solid line corresponds to their average predictions, and the dashed red line denotes the true value of $\bar{\varepsilon}_{max}$. There are two clear patterns: a) for a given number of neurons $n$, shallow-and-wide networks ($AR > 1$) consistently overestimate $\bar{\varepsilon}_{max}$ whereas the other shapes are much closer to the true value, and b) as the network size increases, the entire range of PINNs tend to estimate correctly $\bar{\varepsilon}_{max}$. This is a very important observation, which indicates that when shallow-and-wide PINNs of smaller size are used in I-FENN, the numerical solution is likely to require more Newton-Raphson iterations to converge due to the inaccurate $\bar{\varepsilon}_{max}$ prediction.

\begin{figure}[H]
	\centering
	\includegraphics[scale=0.55]{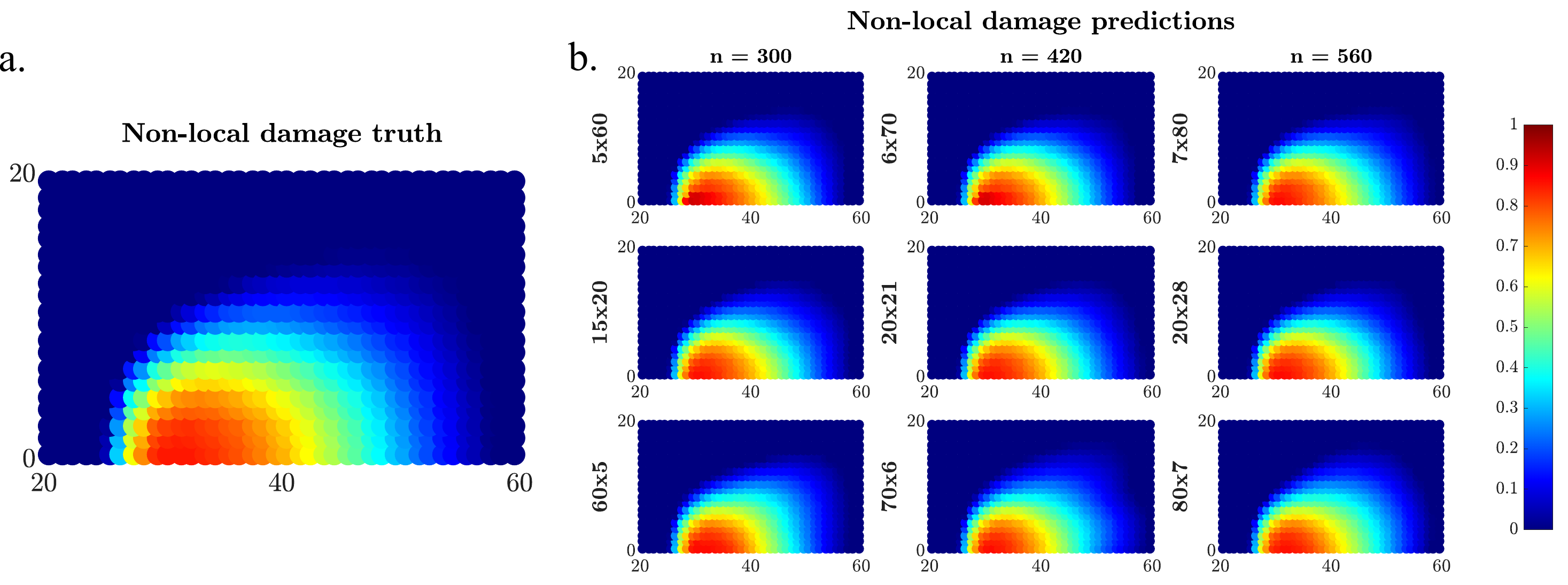}
	\caption{{\bf{a}}: True values of the non-local damage for the Coarse mesh at $lf = 0.82$. Each circle in the graph depicts the damage value at the corresponding Gauss point. {\bf{b}}: Predictions of the non-local damage profile, using the PINNs that are closest to the $\bar{\varepsilon}_{max}$ for the shallowest-and-widest ($AR >> 1$), close-to-square ($AR \approx 1$) and deepest-and-narrowest ($AR << 1$).}
	\label{Figure_IFENN_Damage_ARstudy}
\end{figure}

\begin{figure}[H]
	\centering
	\includegraphics[scale=0.6]{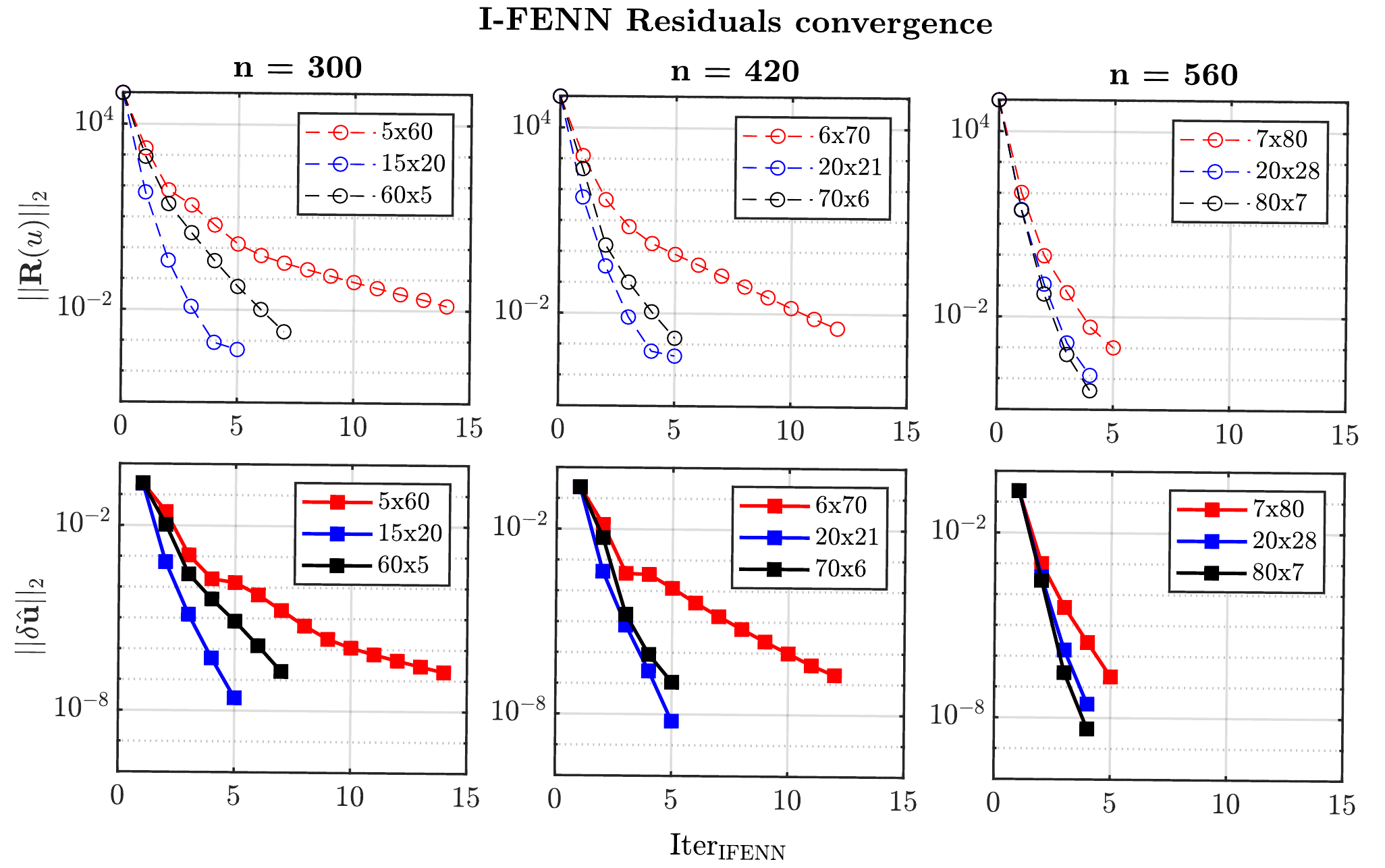}
	\caption{Convergence of I-FENN once the trained PINNs are integrated in the numerical solver. The top row shows the residuals of internal stresses ${\bf{R}}(u)$, and the bottom row the residuals of $\delta {\bf{\hat{u}}}$. The selected networks are those with their predicted $\bar{\varepsilon}_{max}$ closest to the true value, for the cases of the shallowest-and-widest ($AR >> 1$), close-to-square ($AR \approx 1$) and deepest-and-narrowest ($AR << 1$). The legends in the plots indicate the network shapes.}
	\label{Figure_IFENN_Residuals_ARstudy}
\end{figure}

To verify this assumption, we perform the following experiment. We select the trained PINNs which are closest to the true $\bar{\varepsilon}_{max}$ from the shallowest-and-widest ($AR >> 1$), close-to-square ($AR \approx 1$) and deepest-and-narrowest ($AR << 1$) for all $n$ cases. We execute I-FENN with each one of these 9 PINNs, and we report the numerical results in Figures \ref{Figure_IFENN_Damage_ARstudy} and \ref{Figure_IFENN_Residuals_ARstudy}. Figure \ref{Figure_IFENN_Damage_ARstudy} shows the predicted damage profiles for all 9 models, and we observe a sufficient qualitative and quantitative resemblance with the true damage ($d$) field. This is a finding which clearly shows that PINNs across the entire aspect ratio range are viable candidates for the particular problem. We now plot all the residual minimization trials from the non-linear I-FENN solution in Figure \ref{Figure_IFENN_Residuals_ARstudy}, where the top row shows the internal stresses residual ${\bf{R}}(u)$, the bottom row shows the the $\delta {\bf{\hat{u}}}$ residual, and each column corresponds to another $n$ case. Inspection of the residual minimization is very instructive and verifies our earlier observations: even though all cases converge, the shallowest-and-widest cases of the smaller sizes (5x60 and 6x70) require a substantially high number of iterations. On the contrary, all the other cases converge within 5-7 iterations. Since the main goal of I-FENN is the acceleration of the numerical solution, the importance of this result is evident. This finding reveals a rather hidden feature of shallow-and-wide networks, which could not be captured by the conventional metrics used at the beginning of this section. Combined with the trends observed in Figure \ref{Figure_MaxStrains_10000_highlr}, these results assist in understanding the impact of the PINN aspect ratio in the numerical solution and shed more light in selecting a suitable network shape.

\subsection{Concluding the PINN engineered HPS}
\label{Sec:Concluding_PINN_HPS}

Here we provide a summary of the main observations from the engineering-guided HPS:

\begin{itemize}

    \item Shallow-and-wide networks require a significant number of L-BFGS iterations $\mathrm{Iter_{LBFGS}}$ to yield low training and global errors, and they are fast to train with Adam.  

    \item As PINNs become more deep-and-narrow, they become computationally slower during Adam. L-BFGS tends to converge faster, but it has only a minor impact on their predictive accuracy. 

    \item Shallow-and-wide networks of low complexity tend to overestimate the maximum strain value $\bar{\varepsilon}_{max}$, which in turn increases the number of Newton-Raphson iterations within I-FENN until convergence.
    
    \item Very deep-and-narrow networks are prone to getting stuck at trivial solutions due to vanishing gradients. We report several cases of complete failure of training, and provide a detailed explanation of its mechanics.

    \item While accounting for all criteria (accuracy, cost, trivial solutions), the optimum network architecture for the non-local gradient PDE has a close-to-square shape ($AR \approx 1$).     
    
\end{itemize}

\section{Validation studies}
\label{Sec:Section_Validation}

The goal of this section is to provide further validation on the conclusions we have established so far, which are summarized in Sections \ref{Sec:Concluding_PINN_errorconvergence} and \ref{Sec:Concluding_PINN_HPS}. We therefore conduct a few sample checks on the PINN error convergence and engineered HPS and below we present the results for the double-notch and the L-shaped models which are described in Section \ref{Sec:Investigation_and_Validation_Models}.

\subsection{Validation of PINN error convergence}
\label{Sec:Val_PINN_error}

\begin{figure}[H]
	\centering
	\includegraphics[scale = 0.45]{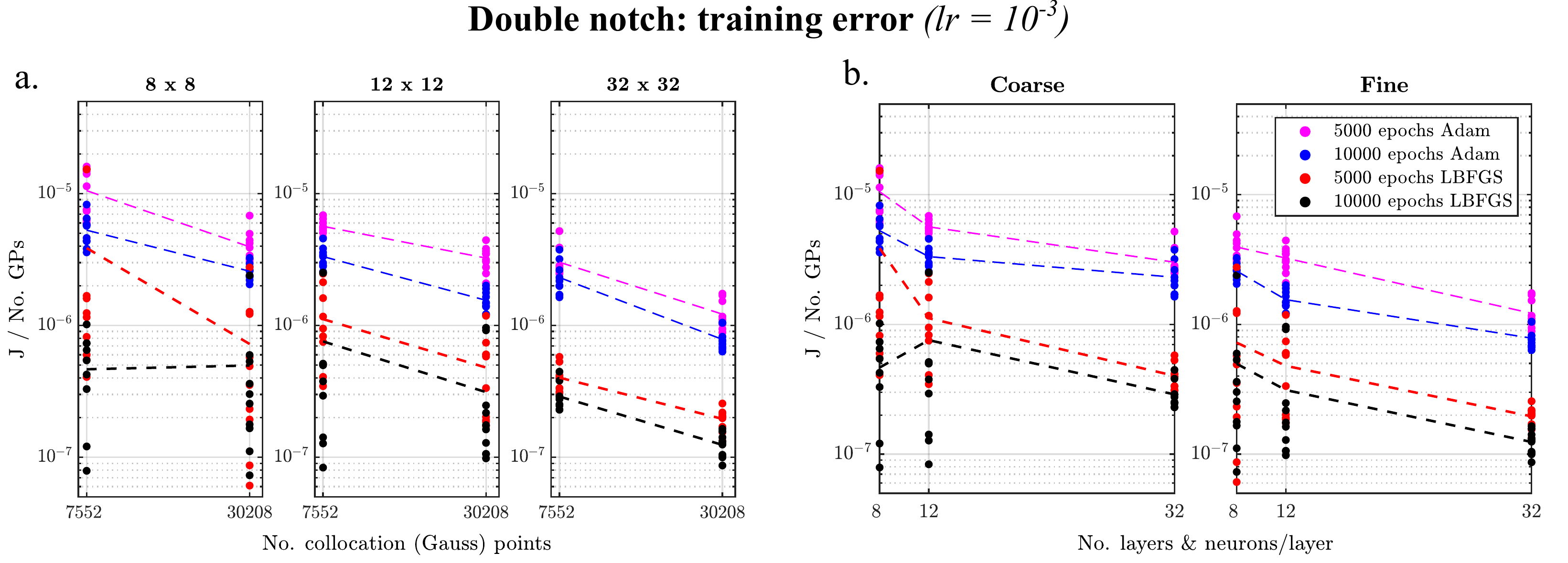}
	\caption{Training error convergence for the double-notch specimen against {\bf{a}}: increasing training sample size and {\bf{b}}: increasing network complexity. The values of $J$ are obtained both at the end of Adam and L-BFGS for $lr = 10^{-3}$ across the Coarse and Fine mesh idealizations, and they are normalized over the number of Gauss points of each mesh. The markers correspond to each of the 10 independent simulations, and the dashed lines connect their average. All networks have a square shape.}
	\label{Figure_Convergence_DoubleNotch_J_highlr}
\end{figure}

\begin{figure}[H]
	\centering
	\includegraphics[scale = 0.45]{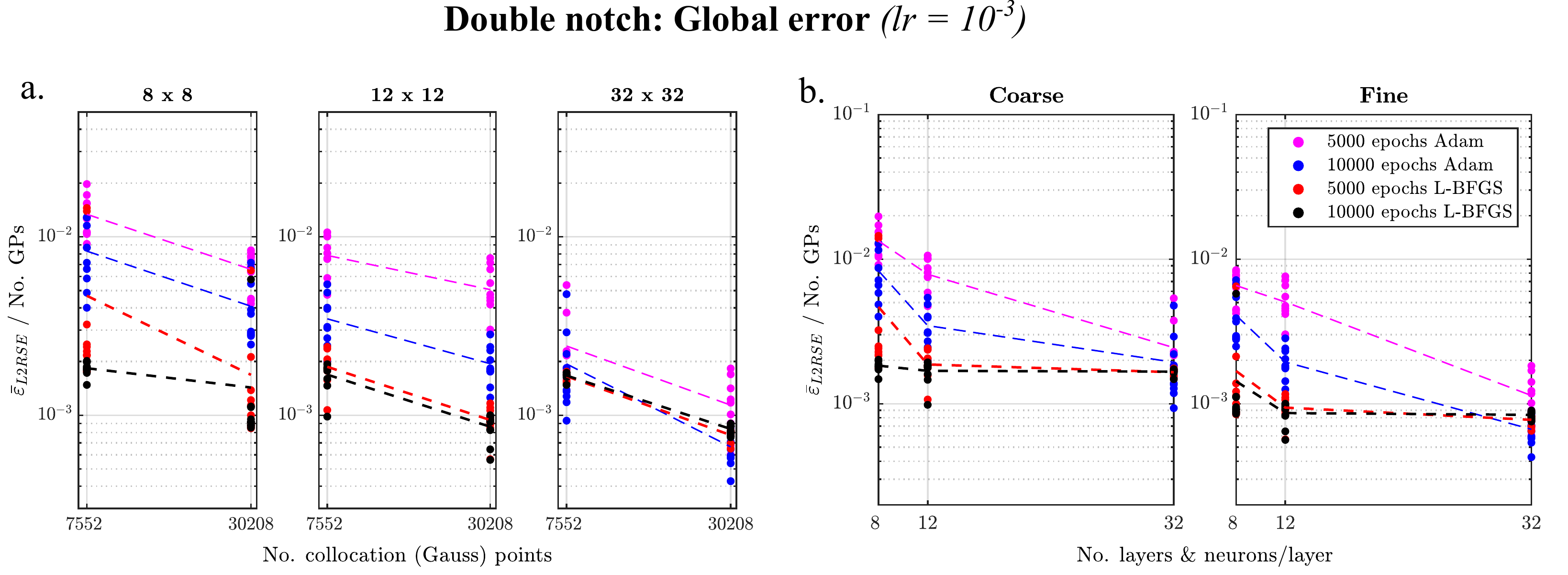}
	\caption{Convergence of the global error ${\bar{\varepsilon}}_{L2RSE}$ for the double-notch specimen with {\bf{a}}: increasing  training sample size and {\bf{b}}: increasing network complexity. The values of ${\bar{\varepsilon}}_{L2RSE}$ are obtained both at the end of Adam and L-BFGS for $lr = 10^{-3}$ across the Coarse and Fine mesh idealizations, and they are normalized over the number of Gauss points of each mesh. The markers correspond to each of the 10 independent simulations, and the dashed lines connect their average. All networks are square.}
	\label{Figure_Convergence_DoubleNotch_L2RSE_highlr}
\end{figure}

We begin by examining the PINN error convergence trends for the double-notch model. We perform sample investigations for the $8x8$, $12x12$ and $32x32$ configurations at $lr = 10^{-3}$ for $ep = 5000$ and $ep = 10000$. We show the normalized training error $J$ values against the network size and number of collocation points in Figure \ref{Figure_Convergence_DoubleNotch_J_highlr}a and \ref{Figure_Convergence_DoubleNotch_J_highlr}b, both at the end of Adam and L-BFGS. A similar layout is followed for the global error $\bar{\varepsilon}_{L2RSE}$ in Figure \ref{Figure_Convergence_DoubleNotch_L2RSE_highlr}a and \ref{Figure_Convergence_DoubleNotch_L2RSE_highlr}b respectively. In the network complexity limit, regardless the number of training epochs or selected optimizer we see an evident decrease for both error metrics. A similar consistent trend is observed in the training sample limit. As the networks become increasingly more complex or are trained on larger datasets, their accuracy is improved. Overall, these results further solidify the generalization of our previous conclusions in Section \ref{Sec:Section_PINN_Error_Convergence}. 

\subsection{Validation of PINN engineering-oriented HPS}
\label{Sec:Val_PINN_HPS}

\begin{figure}[H]
	\centering
	\includegraphics[width=\linewidth]{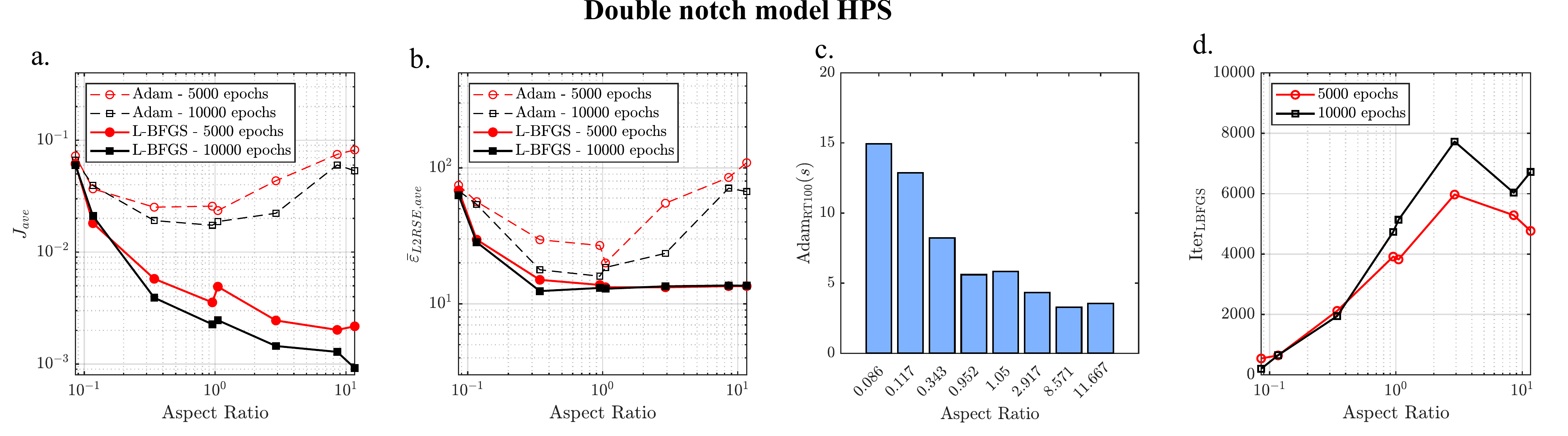}
	\caption{Impact of network shape on the PINN prediction accuracy and training time for the double-notch model, using $n = 420, lr = 10^{-3}$. {\bf{a}}: Training error $J$ value at the end of Adam (solid) and L-BFGS (dashed) algorithm, for 5000 and 10000 epochs. Each data point is the average of 10 independent training simulations. {\bf{b}}: L2RSE of the non-local strain. {\bf{c}}: Average simulation time (sec) of 100 Adam epochs. {\bf{d}}: Number of L-BFGS iterations.}
	\label{Figure_DoubleNotch_HPS}
\end{figure}

Here we perform sample checks on the engineering-oriented HPS. We fix the total number of neurons at $n = 420$ and learning rate at $lr = 10^{-3}$, and we use the same configurations as reported in Figure \ref{Figure_Search_Space}, training for both $ep = 5000$ and $ep = 10000$. Figure \ref{Figure_DoubleNotch_HPS} shows the impact of the network shape on the predictive accuracy and computational effort. We observe the \textit{U} shape of the average training error $J$ value at the end of Adam, and a decreasing trend for $J$ at the end of L-BFGS as the aspect ratio increases. The error metric of the non-local strain obtains its highest values at the small AR regime, and it plateaus after $AR \geq 0.3$. In terms of computational effort, the number of L-BFGS iterations tends to increase with the aspect ratio, whereas execution of Adam is computationally more expensive at the lower aspect ratio zone. Overall, these observations indicate the generalizability of the conclusions that we established for the engineering-oriented HPS using the single-notch model, which were reported in Section \ref{Sec:PINN_Engineering_Guided_HPS}.

\begin{figure}[H]
	\centering
	\includegraphics[width=\linewidth]{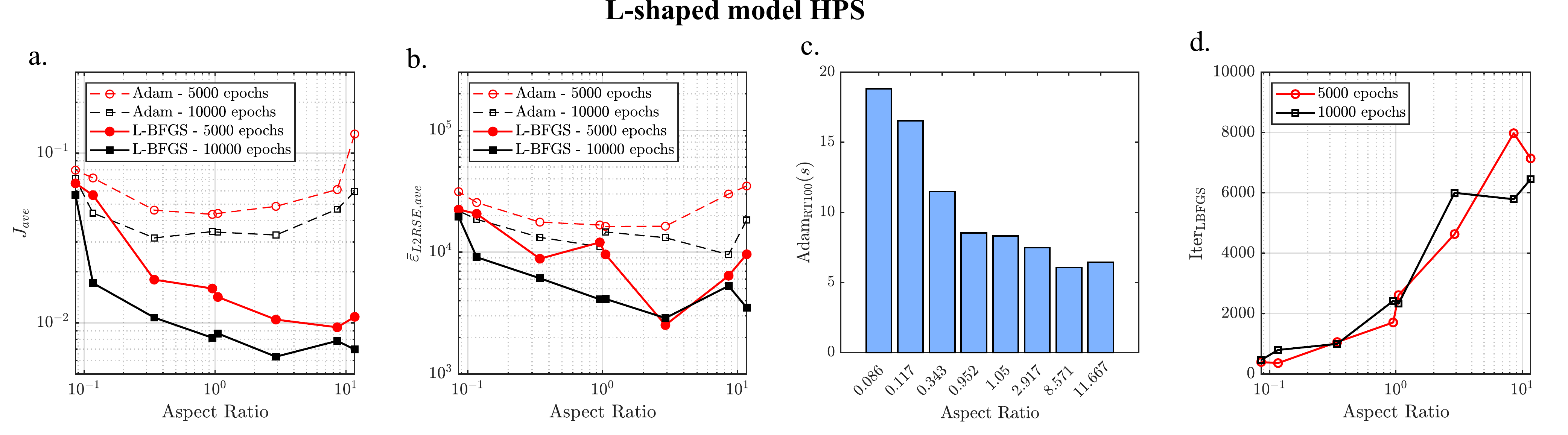}
	\caption{Impact of network shape on the PINN prediction accuracy and training time for the L-shaped model, using $n = 420, lr = 10^{-3}$. {\bf{a}}: Training error $J$ at the end of Adam (solid) and L-BFGS (dashed) algorithm, for 5000 and 10000 epochs. Each data point is the average of 10 independent training simulations. {\bf{b}}: L2RSE of the non-local strain. {\bf{c}}: Average simulation time (sec) of 100 Adam epochs. {\bf{d}}: Number of L-BFGS iterations.}
	\label{Figure_Lshaped_HPS}
\end{figure}

\section{Summary and conclusions}
\label{Sec:Discussion_Summary}

Harnessing PINNs for complex engineering-oriented applications persists as a challenging task that requires substantial efforts to a) establish their robustness as a numerical/computational tool, and b) develop a road-map for reliable and optimal PINN design strategies. In view of our recently proposed I-FENN framework, this paper attempts a detailed numerical investigation on both aspects.  

Regarding the first objective, our convergence analysis is informed by the available mathematical theories in the field, and thus it moves past the realm of uneducated search for optimal hyperparameters. We underline that such a detailed convergence investigation against the network complexity, dataset size, choice of optimizer and other hyperparameters has not been reported in the literature. We demonstrate that the training error and the global error of our PINN setup exhibit numerical convergence. We also show that a higher complexity network is more impactful on the PINN accuracy than a larger dataset, which leads to more enhanced I-FENN performance by allowing to train bigger networks on coarse datasets and predicting on finer ones.

Regarding the second objective, we establish a holistic set of metrics that characterize the PINN accuracy, computational cost, chances of trivial solutions, robustness of optimizer path, and network impact on I-FENN performance. This aligns with our goal to establish a start-to-finish, comprehensive computational tool that is designed for time-sensitive industrial applications. While accounting for all criteria, we show that the optimum PINN architecture for the non-local gradient PDE has an aspect ratio close to unity. This shape outperforms more shallow-and-wide networks in terms of computational cost in the L-BFGS and I-FENN stages, while deeper-and-narrower networks are overall less accurate, more expensive during Adam and more prone to trivial solutions. We refer to Section \ref{Sec:Concluding_PINN_HPS} for a more detailed list of our findings.

Overall, the outcomes of this paper include: a) higher confidence in the PINN setup for the non-local gradient I-FENN, b) more optimized I-FENN for enhanced performance, and c) a systematic approach that can be extended to the analysis of the performance of other PINN and Scientific Machine Learning applications. Although this paper has taken an essential step to fulfill long-standing gaps in the field, there are still several open questions regarding I-FENN and PINNs. First, generalizing I-FENN in the time and space domain is anticipated to significantly improve its competitiveness against existing numerical methods. Additionally, tackling more than one PDEs with its embedded neural network, as well as elucidating the impact of the curse of dimensionality within the I-FENN setup, are two closely-related trajectories that are expected to enhance drastically the I-FENN performance and fully harness its potential against multi-scale problems in mechanics. These trajectories are subject to current work by the authors.

\section*{CRediT authorship statement}
\label{Sec:Credit_Statement}

\textbf{Panos Pantidis:} Conceptualization, Methodology, Software, Formal Analysis, Writing - Original draft, Data Curation, Visualization, Supervision. \textbf{Habiba Eldababy:} Formal Analysis, Writing - Review and Editing. \textbf{Christopher Miguel Tangle:} Formal Analysis, Writing - Review and Editing. \textbf{Mostafa Mobasher:} Conceptualization, Methodology, Writing - Review and Editing, Supervision, Project administration, Funding acquisition.

\section*{Acknowledgements}
\label{Sec:Acknowledgements}

The authors would like to acknowledge the support of the NYUAD Center for Research Computing for providing resources, services, and staff expertise, and in particular Fatema Salem Alhajeri for her assistance and her support during this line of work. The authors would also like to thank the anonymous reviewers for their constructive and diligent feedback during the review process, which significantly improved the quality and clarity of the manuscript.

\section*{Data availability}
\label{Sec:Data_Availability}

The code and data used in this study are publicly available at: https://github.com/ppantidis/I-FENN-part2-Error-convergence-and-HPS-of-PINNs.git

\newpage
\appendix
\section{Gradient non-local damage theory}
\label{Appendix:nonlocalGradientTheory}

\begin{figure}[H]
	\centering
	\includegraphics[scale=0.7]{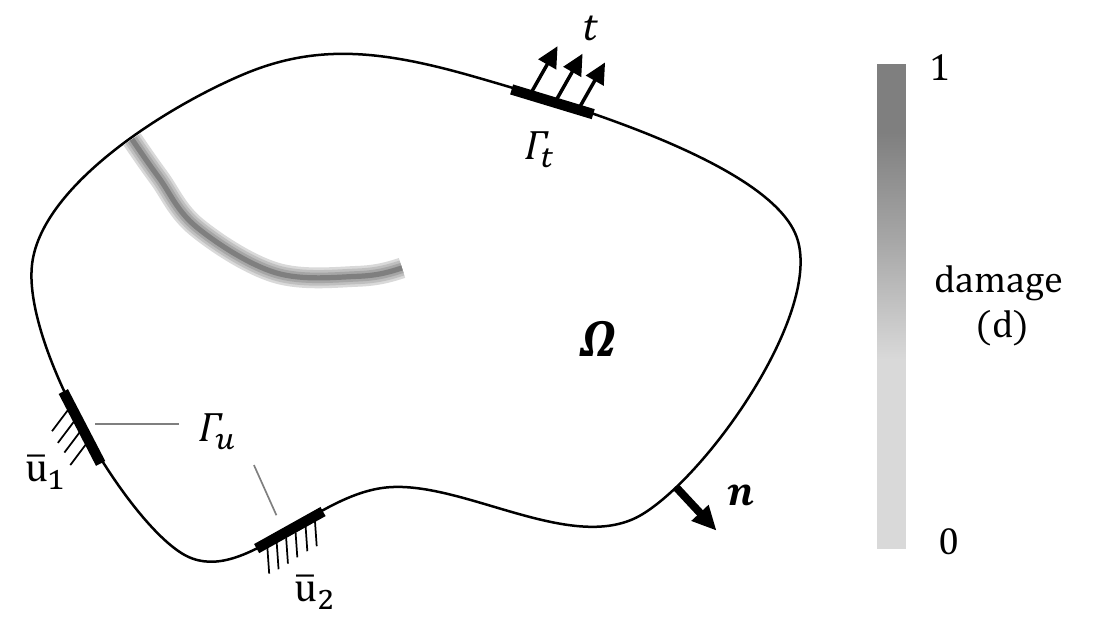}
	\caption{Schematic illustration of an elastic body with prescribed displacement and traction boundary conditions. The crack is represented through the smeared damage field.}
	\label{Figure_Elastic_domain}
\end{figure}

Consider the elastic domain $\Omega$ shown in Figure \ref{Figure_Elastic_domain}, where $\Gamma$ denotes its total boundary and $\bf{q}$ is the outward normal unit vector. Traction forces ${\bar{\bf{t}}}$ are applied in the Neumann boundary part $\Gamma_{t}$ while displacements $\bar{\bf{u}}$ are prescribed in the Dirichlet boundary part $\Gamma_{u}$. The governing equations in the presence of damage read as follows \cite{kachanovbook}:

\begin{equation}
\begin{split}
    \sigma_{ij,j} = 0 \; \; \; in \; \; \Omega
\end{split}
\label{Equilibrium_Condition}
\end{equation}

\begin{equation}
\begin{split}
    u_{i} = {\bar{u}_{i}} \; \; \; in \; \; \Gamma_{u}
\end{split}
\label{Dirichlet_BCs}
\end{equation}

\begin{equation}
\begin{split}
    \sigma_{ij} \cdot q_{i} = {\bar{t}_{j}} \; \; \; in \; \; \Gamma_{t}
\end{split}
\label{Neumann_BCs}
\end{equation}

\begin{equation}
\begin{split}
    \sigma_{ij} = (1 - d) \: C_{ijkl} \varepsilon_{kl} \; \; \; in \; \; \Omega
\end{split}
\label{Cauchy_stress}
\end{equation}

\noindent where $\sigma_{ij}$ is the Cauchy stress tensor, $C_{ijkl}$ is the fourth-order elasticity tensor, $\varepsilon_{kl}$ is the strain tensor. In the context of continuum damage mechanics, the discrete crack is represented through a continuous stiffness degradation field. The magnitude of the stiffness loss at each material point is described by a scalar variable $d$, which represents the damage at this point and ranges between 0 (intact) and 1 (completely failed). Damage is a function of the $local \ equivalent \ strain$, which is a scalar invariant measure of the macroscopic deformation at each material point. However, the dependence of damage to a locally defined strain results to the well-known loss of uniqueness of the numerical solution upon mesh refinement \cite{geersstrainbased}. To avoid the mesh dependent nature of the solution, several non-local formulations have been proposed \cite{pijaudier1987nonlocal, peerlings1996gradient, moes2021lip, deborstcomparison}. In the gradient-enhanced damage formulation by Peerlings et al. \cite{peerlings1996gradient}, $d$ is dispersed over a material zone that is defined by a characteristic length scale measure $l_{c}$, and it is computed as a function of a non-local equivalent strain $\bar{\varepsilon}_{eq}$:

\begin{equation}
\begin{split}
    {\bar{\varepsilon}}_{eq} - g \nabla^{2} {\bar{\varepsilon}}_{eq} - \varepsilon_{eq} = 0 \; \; \; in \; \; \Omega
\label{nonlocalGradientPDE_app}
\end{split}
\end{equation}

\begin{equation}
\begin{split}
    \nabla {\bar{\varepsilon}}_{eq} \cdot n_{i} = 0 \; \; \; in \; \; \Gamma
\end{split}
\label{nonlocalGradientPDEBCs_app}
\end{equation}

\begin{equation}
\begin{split}
    d = d(\bar{\varepsilon}_{eq})
\end{split}
\label{nonlocalGradientDamage}
\end{equation}

\noindent where $\nabla^{2}$ is the Laplacian operator and $g = l_{c}^{2}/2$. For this study we select the Mazars damage evolution relationship \cite{mazars1986description} to compute the non-local damage variable.

\section{I-FENN on non-local gradient damage}
\label{Appendix:IFENN}

Below we provide the mathematical backbone of the I-FENN implementation for the non-local gradient damage equation (see \cite{pantidis2022integrated} for a detailed version). The residual \textbf{R}(u) and Jacobian \textbf{J}(u) can be calculated as:

\begin{equation}
    {\bf{R}}(u) = \int_{\Omega} {\bf{B}}^{T} (1 - d) C_{ijkl}\varepsilon_{kl} \; d\Omega
\label{FEM_WeakR_prop}
\end{equation}

\begin{equation}
    {\bf{J}}(u) = {\bf{K}}(u) + \frac{\partial{\bf{K}}(u)}{\partial{\bf{\hat{u}}}}{\bf{\hat{u}}}
\label{FEM_Jexpr1}
\end{equation}

\begin{equation}
    {\bf{K}}(u) = \int_\Omega \! {\bf{B^{T}}} (1-d) {C_{ijkl}} {\bf{B}} \, \mathrm{d}\Omega
\label{FEM_K_prop}
\end{equation}

\begin{equation}
    \frac{\partial{\bf{K}}(u)}{\partial{\bf{\hat{u}}}} = \int_\Omega \! {\bf{B^{T}}} {C_{ijkl}} \, \left(-\frac{\partial{d}}{\partial {\bf{\hat{u}}}}\right) \, {\bf{B}} \, \mathrm{d}\Omega
\label{FEM_DKdu_prop}
\end{equation}

\begin{equation}
    \frac{\partial{d}}{\partial \hat{u}_{k}} = 
    \frac{\partial{d}}{\partial{{\bar{\varepsilon}}^{NN}_{eq}}}                \; 
    \frac{\partial{\bar{\varepsilon}^{NN}_{eq}}}{\partial{{\varepsilon}_{eq}}}       \; 
    \frac{\partial{\varepsilon_{eq}}}{\partial{\varepsilon_{ij}}}               \; 
    \frac{\partial{\varepsilon_{ij}}}{\partial{\hat{u}}_{k}}
\label{FEM_Dddu_prop}
\end{equation}

In equations \ref{FEM_WeakR_prop} - \ref{FEM_Dddu_prop}, we denote that $\hat{\textbf{u}}$ is the vector of nodal displacements, \textbf{B} is the matrix of shape function derivatives, $C_{ijkl}$ is the material constitutive matrix, $\frac{\partial{d}}{\partial{{\bar{\varepsilon}}^{NN}_{eq}}}$ is given by the damage governing law, $\frac{\partial{\varepsilon_{eq}}}{\partial{\varepsilon_{ij}}}$ describes the dependence of the local equivalent strain to its tensorial version, and $\frac{\partial{\varepsilon_{ij}}}{\partial{\hat{u}}_{k}}$ reflects the displacement-strain transformation. Once \textbf{J}(u) and \textbf{R}(u) have been calculated, once can compute $\delta {\bf{\hat{u}}}$ as:

\begin{equation}
    {\bf{J}}(u) \delta {\bf{\hat{u}}} = - {\bf{R}}(u)
\label{FEM_JduR_1}
\end{equation}

\begin{algorithm}
	\caption{Algorithm of the I-FENN framework in the case of non-local gradient damage}
	\label{algorithm_hybrid_FEM_PINN}
    \While {$\frac{||\delta{\bf{\hat{u}}}_{i = last}||_{2}}{||\delta{\bf{\hat{u}}}_{i = 1}||_{2}} \leq tol$}
    {\For{$\mathrm{each}$ $\mathrm{integration}$ $\mathrm{point}$ $and$ $\mathrm{each}$ $\mathrm{boundary}$ $\mathrm{node}$}
            {
            Compute the shape functions $\bf{N}$ and their derivatives $\bf{B}$ \\
            Compute and extract coordinates, $g$ and $\varepsilon_{eq}$
            }
        Use the pre-trained PINN to predict $\bar{\varepsilon}^{NN}_{eq}$ and $\frac{\partial \bar{\varepsilon}^{NN}_{eq}}{{\partial \varepsilon}_{eq}}$ for all IPs \\
        \For{$\mathrm{each}$ $\mathrm{finite}$ $\mathrm{element}$}
        {
            \For{$\mathrm{each}$ $\mathrm{integration}$ $\mathrm{point}$}
            {
            Compute the shape functions $\bf{N}$ and their derivatives $\bf{B}$ \\
            Calculate $\frac{\partial \varepsilon_{eq}}{\partial \varepsilon_{ij}}$, based on the local equivalent strain definition \\
            Calculate $d$ and $\frac{\partial d}{\partial \bar{\varepsilon}^{NN}_{eq}}$, based on the governing damage law \\
            Compute the IP contribution to the element jacobian matrix and residual vector
            }
        }
        Assemble global $\bf{J}$, solve for $\bf{u}$, check convergence of $\bf{R}$}
\end{algorithm}

\newpage
\section{Convergence of PINNs: $lr = 10^{-3}$}
\label{Appendix:Appendix_Dataset_Size}

Here we report the convergence graphs versus the network size and sample limit size of $J$ and ${\bar{\varepsilon}}_{L2RSE}$, for the high learning rate $lr = 10^{-3}$.

\begin{figure}[H]
	\centering
	\includegraphics[scale=0.50]{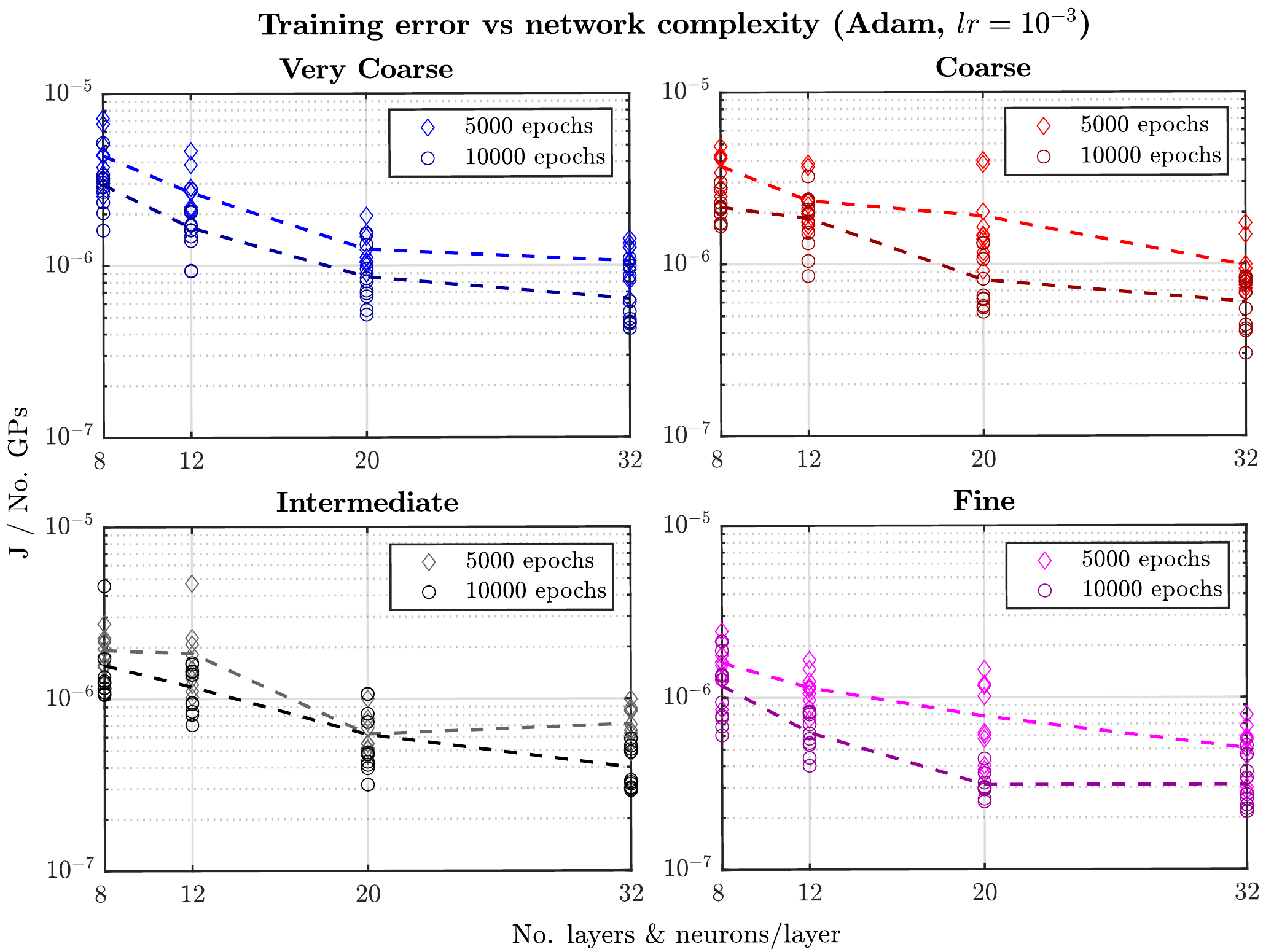}
	\caption{Identical layout with Figure \ref{Figure_Convergence_JvsLayers_Adam_lowlr} with $lr = 10^{-3}$.}
	\label{Figure_Convergence_JvsLayers_Adam_highlr}
\end{figure}

\begin{figure}[H]
	\centering
	\includegraphics[scale=0.50]{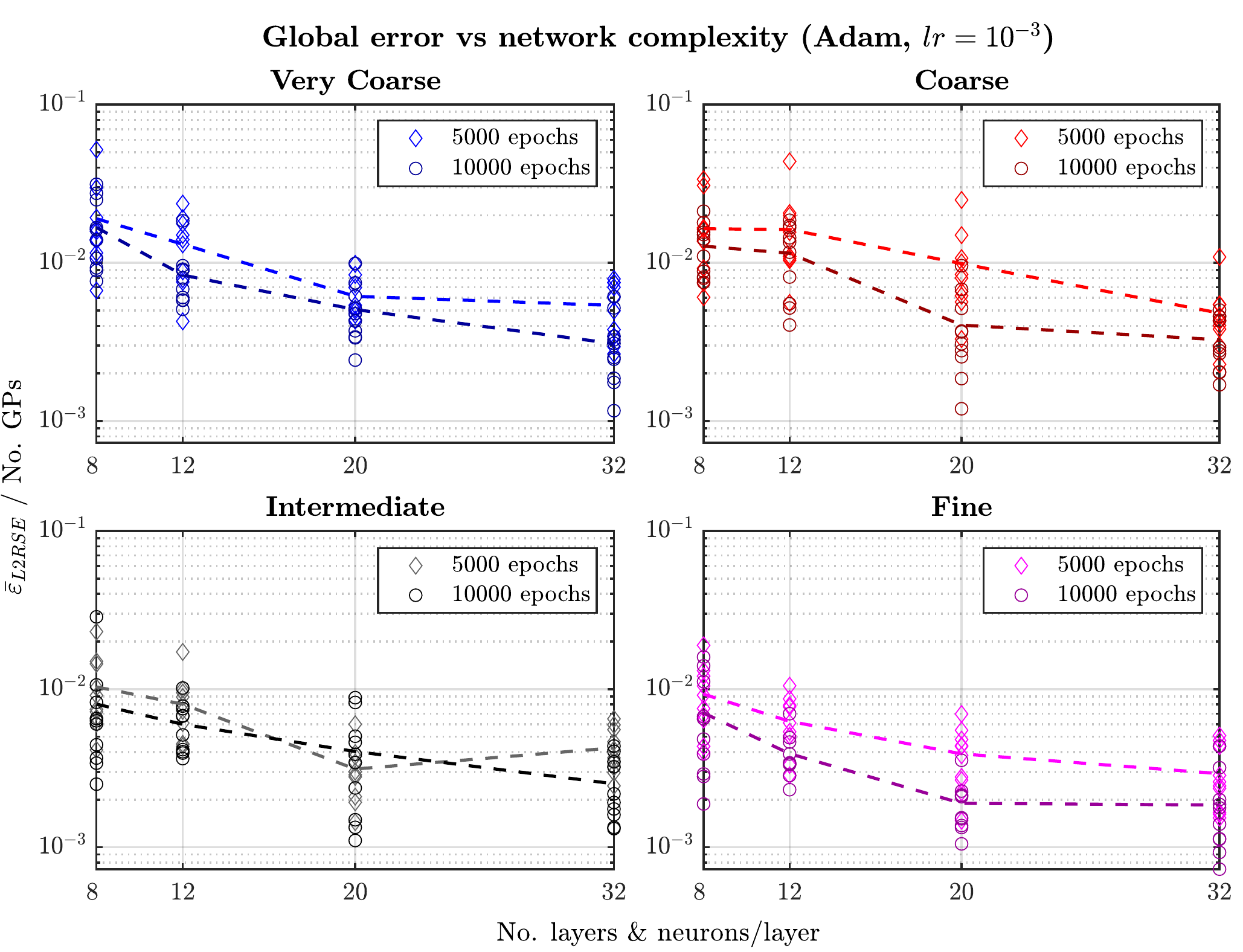}
	\caption{Identical layout with Figure \ref{Figure_Convergence_L2RSEvsLayers_Adam_lowlr} with $lr = 10^{-3}$.}
	\label{Figure_Convergence_L2RSEvsLayers_Adam_highlr}
\end{figure}

\begin{figure}[H]
	\centering
	\includegraphics[scale=0.52]{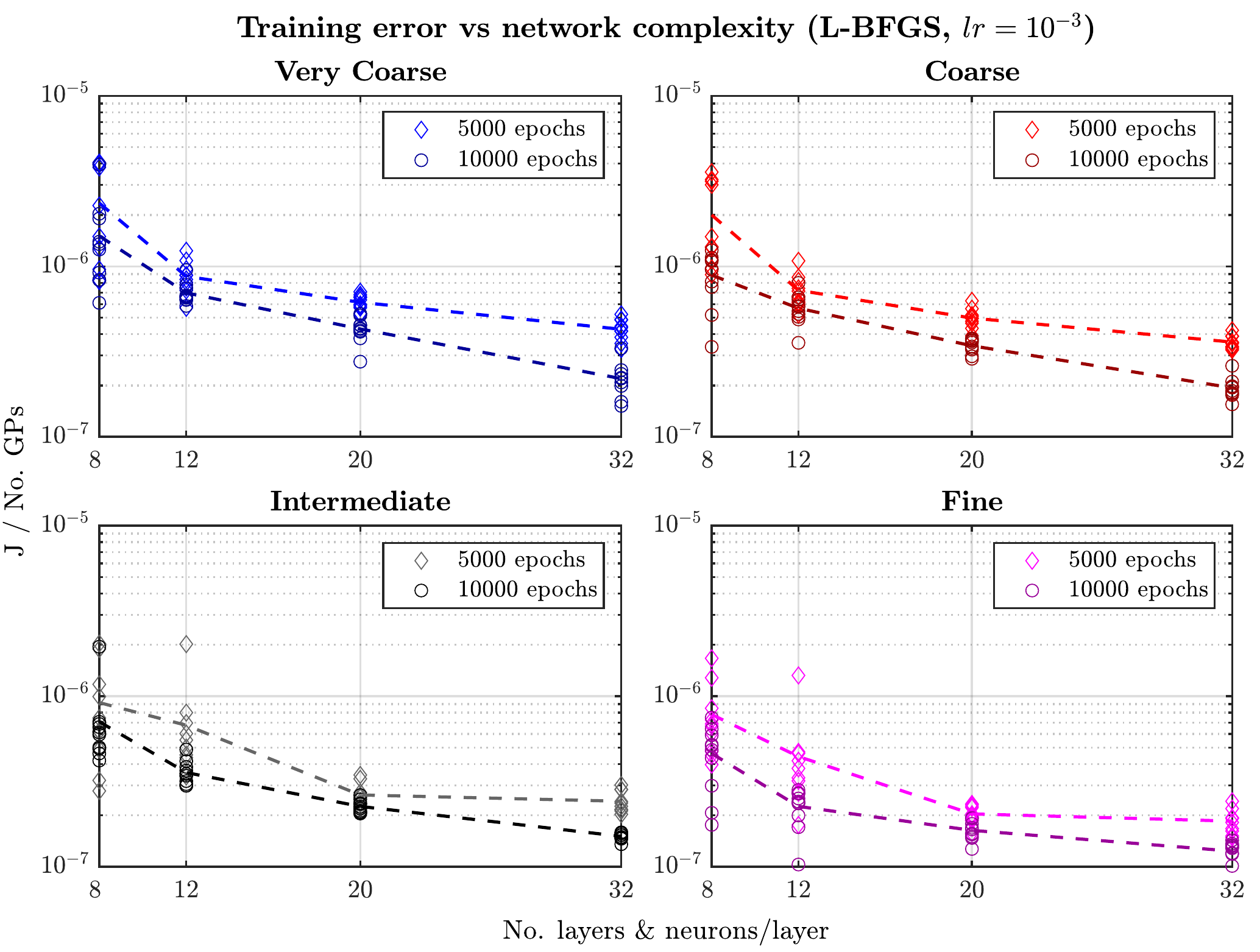}
	\caption{Identical layout with Figure \ref{Figure_Convergence_JvsLayers_LBFGS_lowlr} with $lr = 10^{-3}$.}
	\label{Figure_Convergence_JvsLayers_LBFGS_highlr}
\end{figure}

\begin{figure}[H]
	\centering
	\includegraphics[scale=0.52]{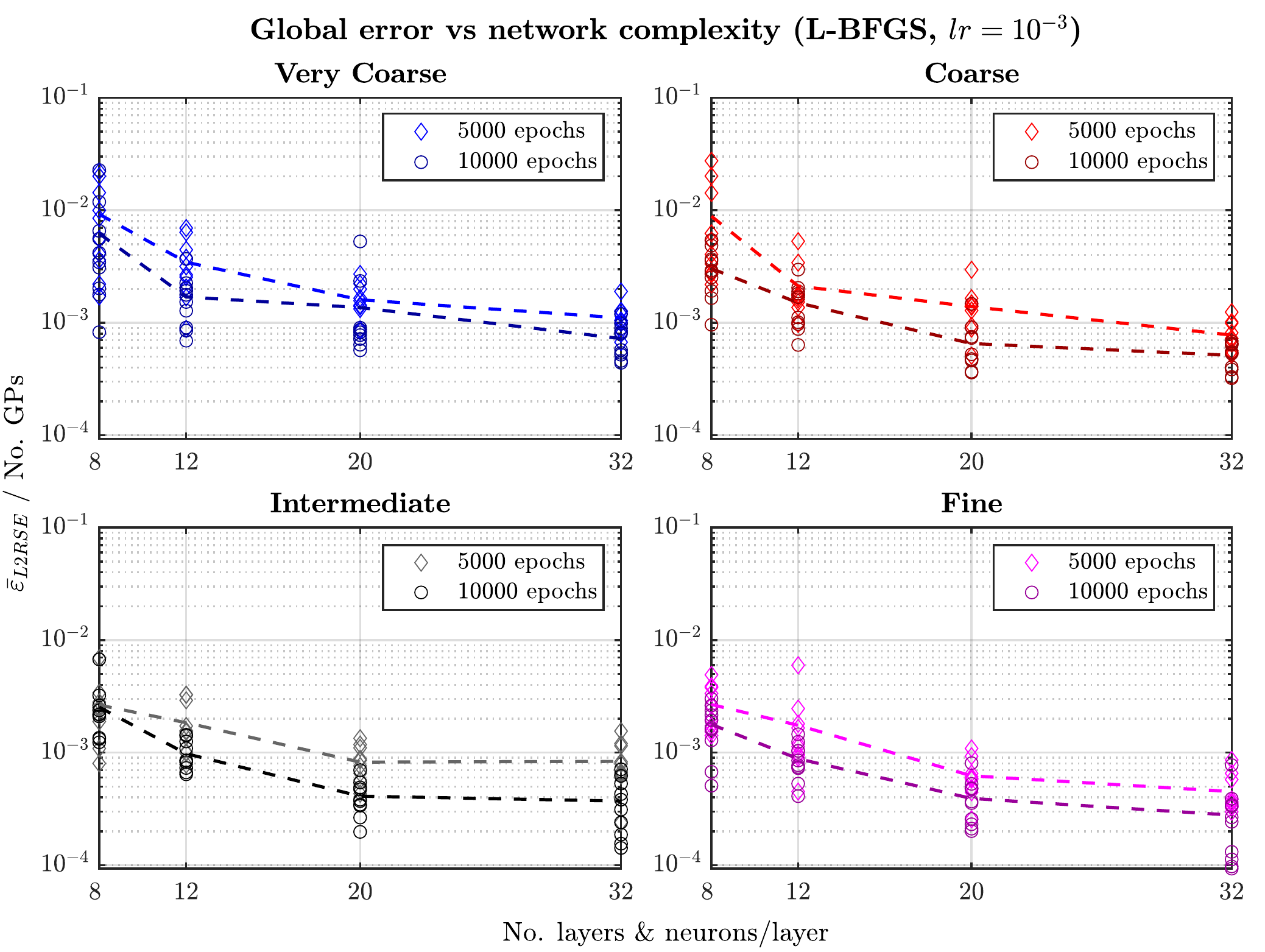}
	\caption{Identical layout with Figure \ref{Figure_Convergence_L2RSEvsLayers_LBFGS_lowlr} with $lr = 10^{-3}$.}
	\label{Figure_Convergence_L2RSEvsLayers_LBFGS_highlr}
\end{figure}

\begin{figure}[H]
	\centering
	\includegraphics[scale=0.52]{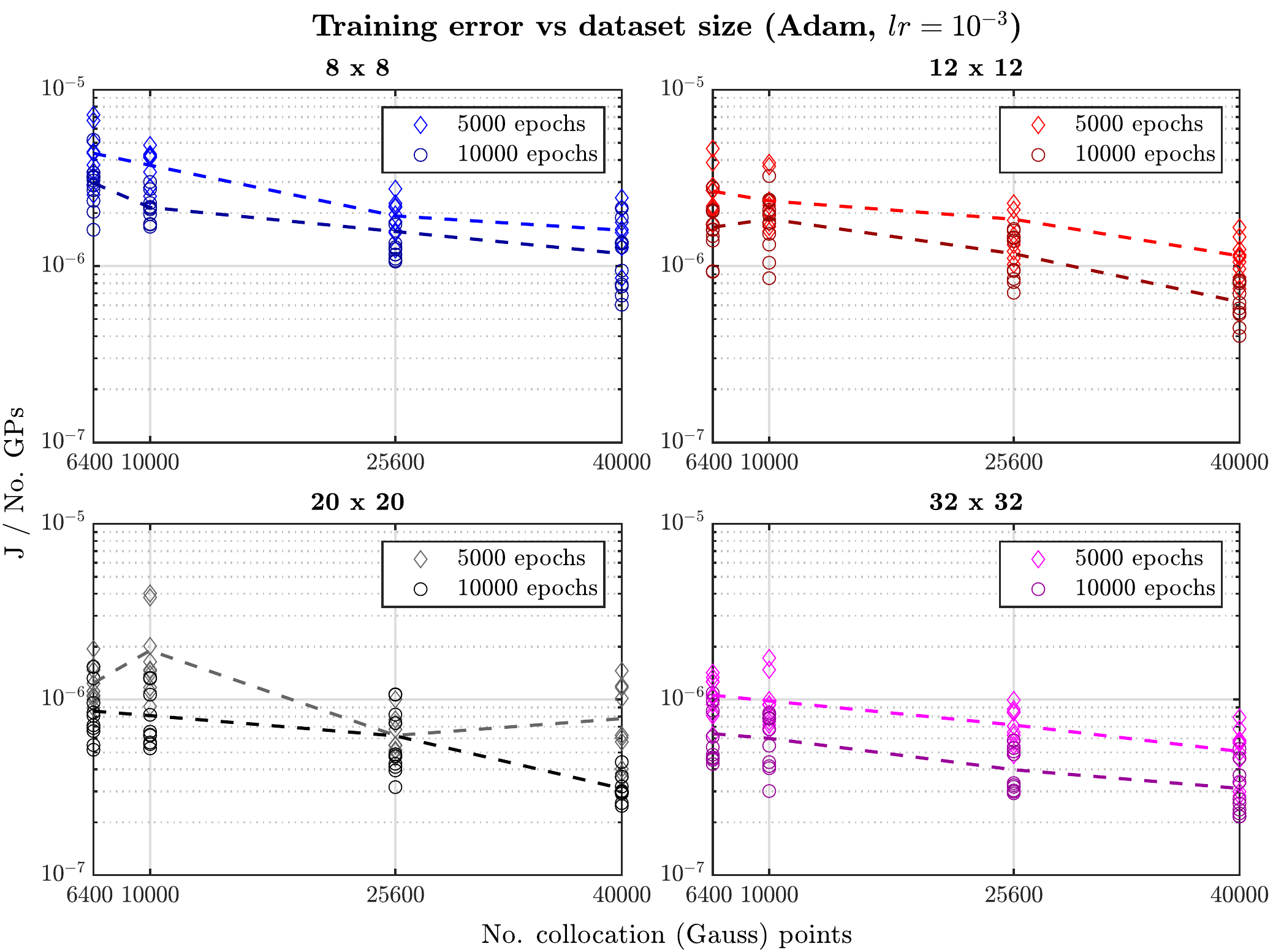}
	\caption{Identical layout with Figure \ref{Figure_Convergence_JvsGPs_Adam_lowlr} with $lr = 10^{-3}$.}
	\label{Figure_Convergence_JvsGPs_Adam_highlr}
\end{figure}

\begin{figure}[H]
	\centering
	\includegraphics[scale=0.52]{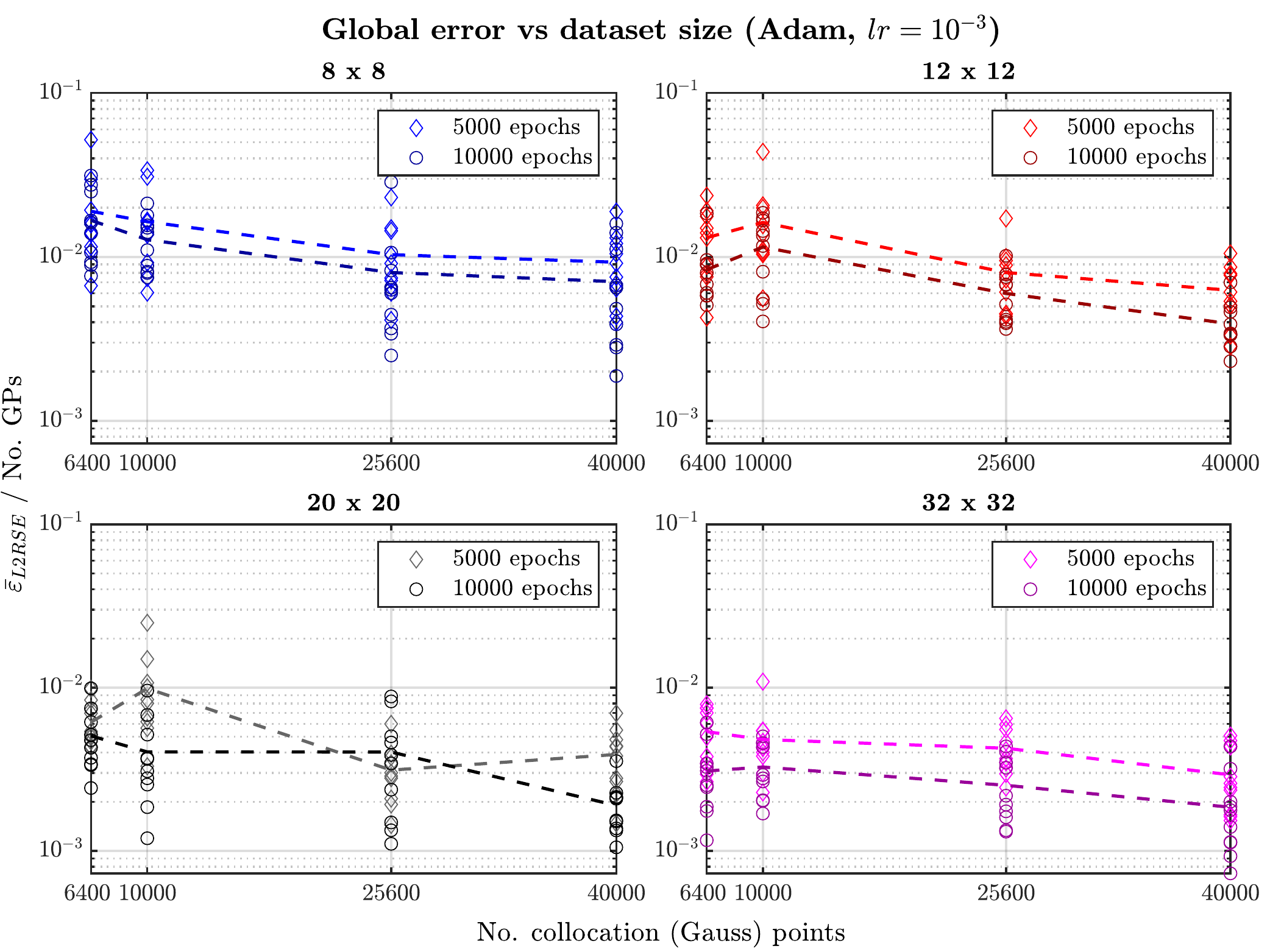}
	\caption{Identical layout with Figure \ref{Figure_Convergence_L2RSEvsGPs_Adam_lowlr} with $lr = 10^{-3}$.}
	\label{Figure_Convergence_L2RSEvsGPs_Adam_highlr}
\end{figure}

\begin{figure}[H]
	\centering
	\includegraphics[scale=0.52]{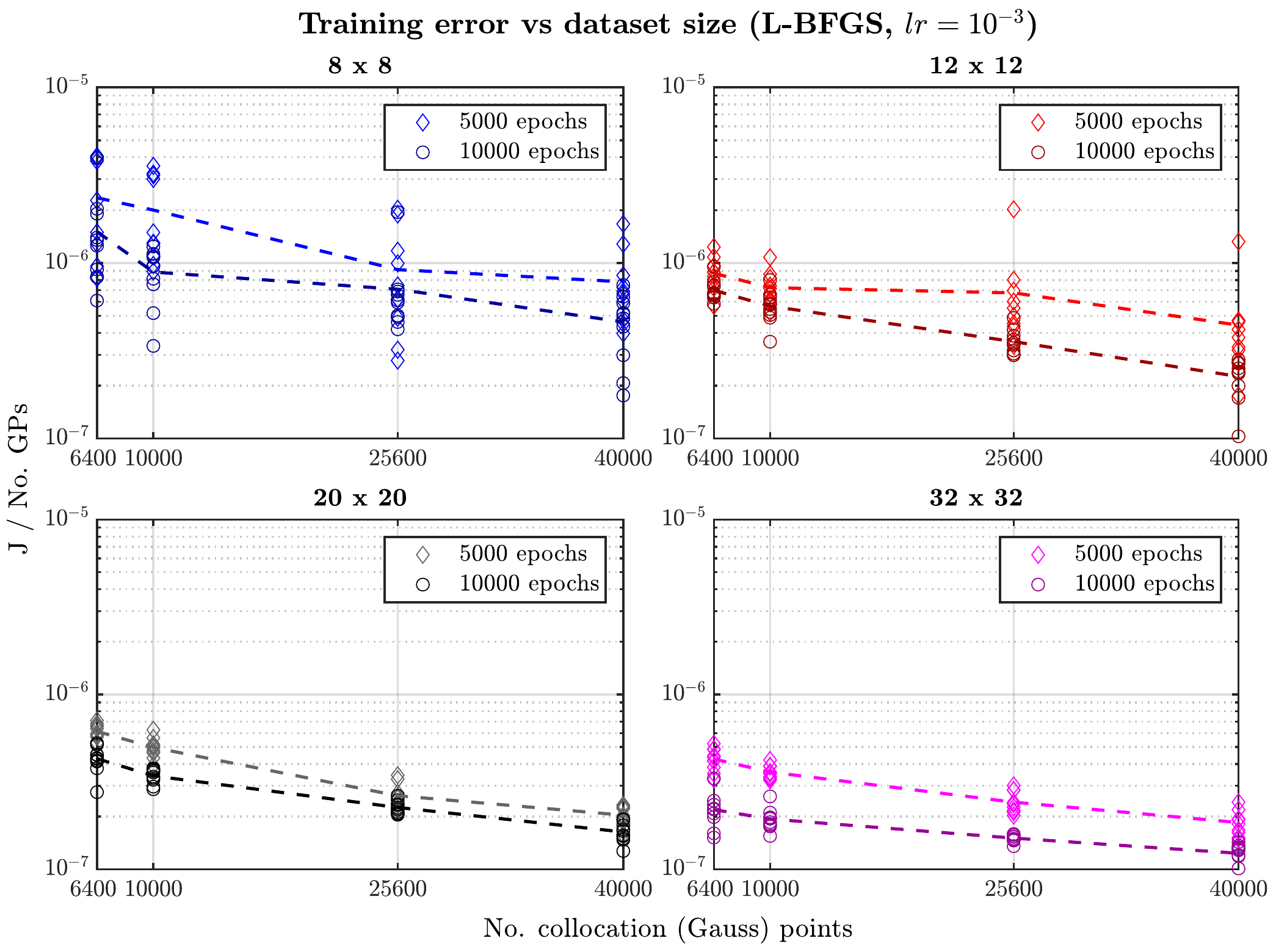}
	\caption{Identical layout with Figure \ref{Figure_Convergence_JvsGPs_LBFGS_lowlr} with $lr = 10^{-3}$.}
	\label{Figure_Convergence_JvsGPs_LBFGS_highlr}
\end{figure}

\begin{figure}[H]
	\centering
	\includegraphics[scale=0.52]{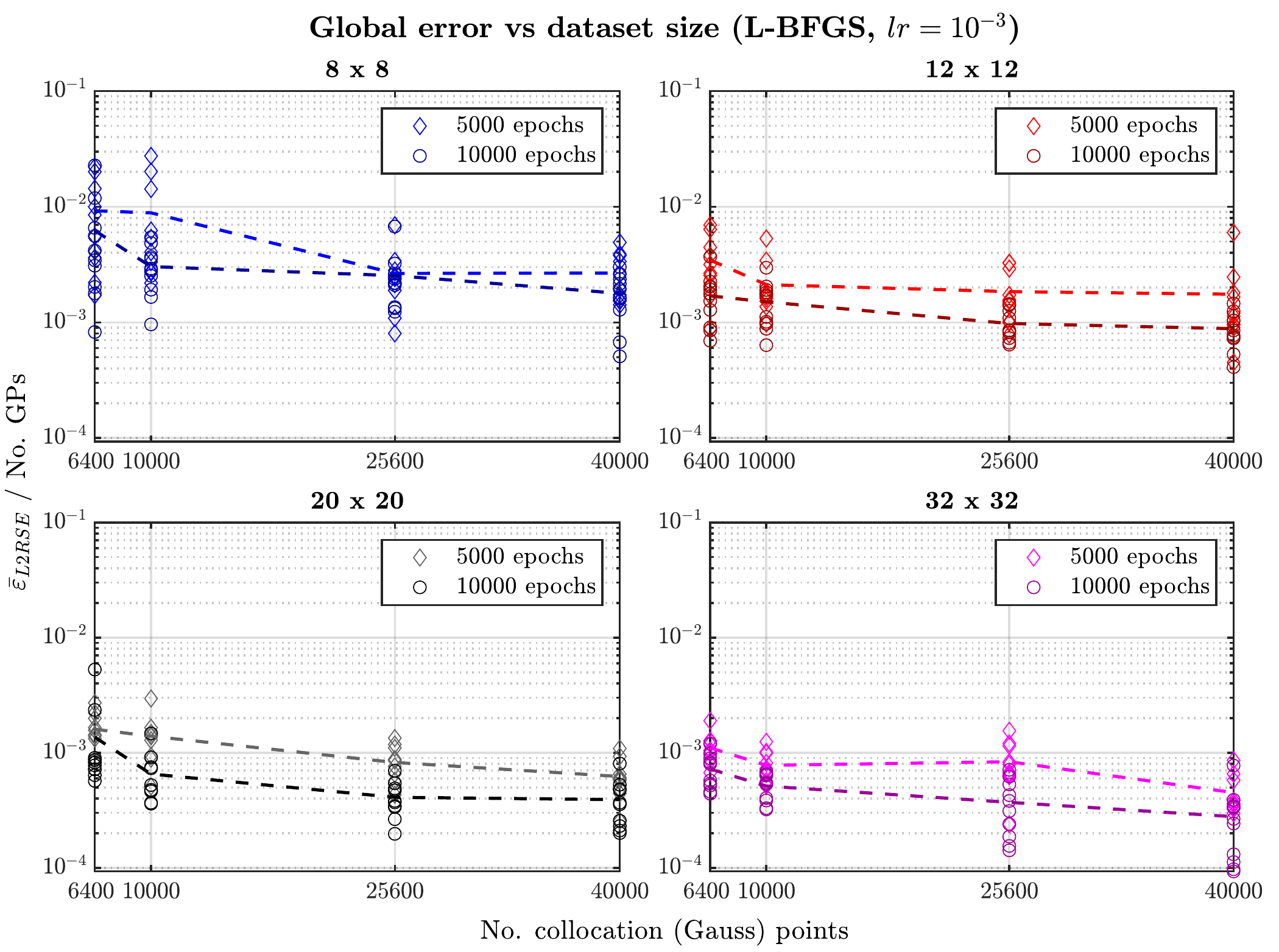}
	\caption{Identical layout with Figure \ref{Figure_Convergence_L2RSEvsGPs_LBFGS_lowlr} with $lr = 10^{-3}$.}
	\label{Figure_Convergence_L2RSEvsGPs_LBFGS_highlr}
\end{figure}

\newpage
\bibliography{bibliography}

\end{document}